\documentclass[10pt,twocolumn,letterpaper]{article}

\usepackage[pagenumbers]{cvpr} %

\def\paperID{1570} %
\def\confName{CVPR}
\def\confYear{2024}

\usepackage[utf8]{inputenc} %
\usepackage[T1]{fontenc}    %
\usepackage[pagebackref=true]{hyperref}       %
\usepackage{url}            %
\usepackage{booktabs}       %
\usepackage{amsfonts}       %
\usepackage{nicefrac}       %
\usepackage{microtype}      %
\usepackage[dvipsnames]{xcolor}         %

\usepackage{colortbl}
\usepackage{graphicx,overpic}
\usepackage[export]{adjustbox}
\usepackage{amsmath, amssymb}
\usepackage{multirow}
\usepackage[super]{nth}
\usepackage{comment}
\usepackage{nicematrix}

\usepackage{enumitem}
\usepackage{caption}

\usepackage{tikz}
\usepackage{pgfplots}
\usepackage{pgfplotstable}
\pgfplotsset{compat=1.13}
\usetikzlibrary{pgfplots.groupplots}
\usetikzlibrary{calc,arrows}

\newcommand\myparagraph[1]{\vspace{0.2ex}\noindent\textbf{#1}\hspace{0.5em}}
\newcommand\pmnum[1]{\small$\pm$#1}

\usepackage{tabularx,stackengine,collcell}
\let\endminwd\relax
\newcolumntype{L}[1]{>{\collectcell\xminwd l{#1}}l<{\endminwd\endcollectcell}}
\newcolumntype{C}[1]{>{\collectcell\xminwd c{#1}}c<{\endminwd\endcollectcell}}
\newcolumntype{R}[1]{>{\collectcell\xminwd r{#1}}r<{\endminwd\endcollectcell}}
\def\minwd#1#2#3\endminwd{\stackengine{0pt}{#3}{\rule{#2}{0pt}}{O}{#1}{F}{F}{L}}
\newcommand\xminwd[1]{\minwd#1}

\usepackage{pifont}
\newcommand{\mj}{$\mathcal{J}$}
\newcommand{\mf}{$\mathcal{F}$}
\newcommand{\mjf}{$\mathcal{J}\&\mathcal{F}$}

\newcommand{\mjs}{$\mathcal{J}_s$}
\newcommand{\mfs}{$\mathcal{F}_s$}
\newcommand{\mju}{$\mathcal{J}_u$}
\newcommand{\mfu}{$\mathcal{F}_u$}
\newcommand{\mg}{$\mathcal{G}$}

\hypersetup{
	colorlinks,
	linkcolor={red!80!black},
	citecolor={blue!80!black},
	urlcolor={blue!80!black},
    pdftitle={Putting the Object Back into Video Object Segmentation},
}

\definecolor{defaultColor}{RGB}{230, 230, 250}

\newcommand{\beginsupplement}{
	\setcounter{table}{0}
	\renewcommand{\thetable}{S\arabic{table}}%
	\setcounter{figure}{0}
	\renewcommand{\thefigure}{S\arabic{figure}}%
	\setcounter{equation}{0}
	\renewcommand{\theequation}{S\arabic{equation}}
}

\title{Putting the Object Back into Video Object Segmentation} 

\author{Ho Kei Cheng\textsuperscript{1} \hspace{1em} Seoung Wug Oh\textsuperscript{2} \hspace{1em} Brian Price\textsuperscript{2} \hspace{1em} Joon-Young Lee\textsuperscript{2} \hspace{1em} Alexander Schwing\textsuperscript{1}\\
\textsuperscript{1}University of Illinois Urbana-Champaign \hspace{2em} \textsuperscript{2}Adobe Research\\
{\tt\small\{hokeikc2,aschwing\}@illinois.edu, \{seoh,bprice,jolee\}@adobe.com}
}

\begin{document}

\maketitle

\begin{abstract}
We present Cutie, a video object segmentation (VOS) network with object-level memory reading, which puts the object representation from memory back into the video object segmentation result. 
Recent works on VOS employ bottom-up pixel-level memory reading which struggles due to matching noise, especially in the presence of distractors, resulting in lower performance in more challenging data.
In contrast, Cutie performs top-down object-level memory reading by adapting a small set of object queries.
Via those, it interacts with the bottom-up pixel features iteratively with a \textbf{q}uery-based object \textbf{t}ransformer (\textbf{qt}, hence Cutie).
The object queries act as a high-level summary of the target object, while high-resolution feature maps are retained for accurate segmentation. 
Together with foreground-background masked attention, Cutie cleanly separates the semantics of the foreground object from the background.
On the challenging MOSE dataset, Cutie improves by 8.7 \mjf~over XMem with a similar running time and improves by 4.2 \mjf~over DeAOT while being three times faster.
Code is available at: {\href{https://hkchengrex.github.io/Cutie}{\nolinkurl{hkchengrex.github.io/Cutie}.}}
\end{abstract}

\section{Introduction}

Video Object Segmentation (VOS), specifically the ``semi-supervised'' setting, requires tracking and segmenting objects from an open vocabulary specified in a first-frame annotation. 
VOS methods are broadly applicable in  robotics~\cite{petrik2022learning}, video editing~\cite{cheng2021mivos}, reducing costs in data annotation~\cite{athar2023burst}, and can also be combined with Segment Anything Models (SAMs)~\cite{kirillov2023segment} for universal video segmentation (e.g.,\ Tracking Anything~\cite{cheng2023tracking,yang2023track,cheng2023segment}). 

Recent VOS approaches employ a memory-based paradigm~\cite{oh2019videoSTM,cheng2022xmem,yang2021associating,bekuzarov2023xmem++}.
A memory representation is computed from past segmented frames (either given as input or segmented by the model), and any new query frame ``reads'' from this memory to retrieve features for segmentation. 
Importantly, these approaches mainly use \emph{pixel-level matching} for memory reading, either with one~\cite{oh2019videoSTM} or multiple matching layers~\cite{yang2021associating}, and generate the segmentation bottom-up from the pixel memory readout. 
Pixel-level matching maps every query pixel independently to a linear combination of memory pixels (e.g., with an attention layer). 
Consequently, pixel-level matching lacks high-level consistency and is prone to matching noise, especially in the presence of distractors. This leads to lower performance in challenging scenarios with occlusions and frequent distractors. Concretely, the performance of recent approaches~\cite{cheng2022xmem,yang2021associating} is more than 20 points in \mjf~lower when evaluating on the recently proposed challenging MOSE~\cite{ding2023mose} dataset rather than the simpler DAVIS-2017~\cite{perazzi2016benchmark} dataset.

\begin{figure}
    \centering
    \includegraphics[width=0.8\linewidth]{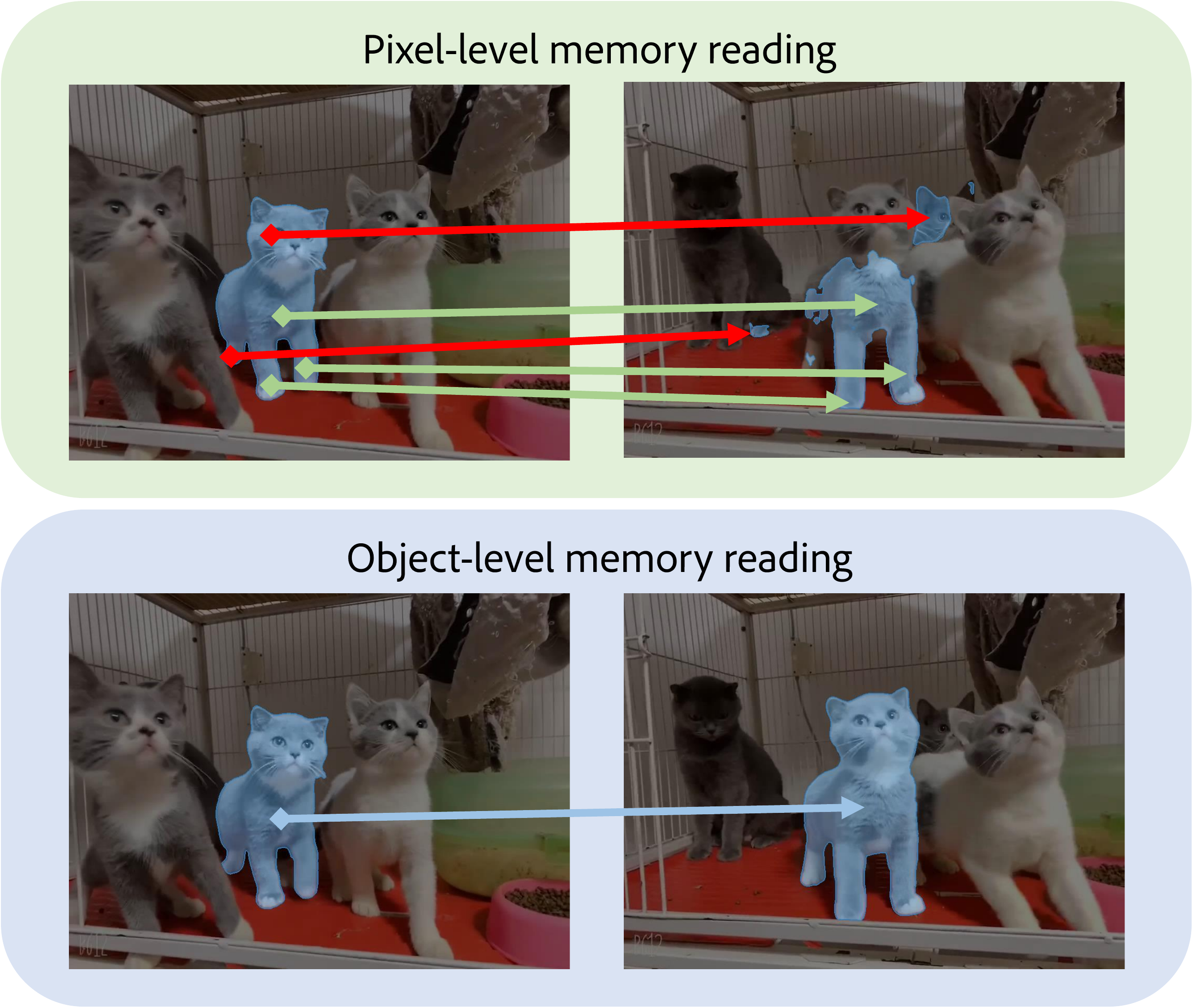}
    \caption{Comparison of pixel-level memory reading v.s.\ object-level memory reading. 
    In each box, the left is the reference frame, and the right is the query frame to be segmented. Red arrows indicate wrong matches.
    Low-level pixel matching (e.g., XMem~\cite{cheng2022xmem}) can be noisy in the presence of distractors. 
    We propose object-level memory reading for more robust video object segmentation.}
    \label{fig:compare-matching}
\end{figure}

We think this unsatisfactory result in challenging scenarios is caused by the lack of object-level reasoning. 
To address this, we propose \emph{object-level memory reading}, which effectively puts the object from a memory back into the query frame (Figure~\ref{fig:compare-matching}). 
Inspired by recent query-based object detection/segmentation~\cite{carion2020end,cheng2022masked,wu2022defense,athar2023tarvis,athar2022hodor} that represent objects as ``object queries,'' we implement our object-level memory reading with an object transformer.
This object transformer uses a small set of end-to-end trained object queries to 1) iteratively probe and calibrate a feature map (initialized by a pixel-level memory readout), and 2) encode object-level information.
This approach simultaneously keeps a high-level/global object query representation and a low-level/high-resolution feature map, enabling bidirectional top-down/bottom-up communication.
This communication is parameterized with a sequence of attention layers, including a proposed \emph{foreground-background masked attention}.
The masked attention, extended from foreground-only masked attention~\cite{cheng2022masked}, lets part of the object queries attend only to the foreground while the remainders attend only to the background -- allowing both global feature interaction and clean separation of foreground/background semantics. 
Moreover, we introduce a compact \emph{object memory} (in addition to a pixel memory) to summarize the features of target objects, enhancing end-to-end object queries with target-specific features.

In experiments, the proposed approach, \emph{Cutie}, is significantly more robust in challenging scenarios (e.g., +8.7 \mjf~in MOSE~\cite{ding2023mose} over XMem~\cite{cheng2022xmem}) than existing approaches while remaining competitive in standard datasets (i.e., DAVIS~\cite{perazzi2016benchmark} and YouTubeVOS~\cite{xu2018youtubeVOS}) in both accuracy and efficiency. In summary, 
\begin{itemize}[topsep=1ex,itemsep=1pt,partopsep=1pt, parsep=1pt]
    \item We develop \emph{Cutie}, which uses high-level top-down queries with pixel-level bottom-up features for robust video object segmentation in challenging scenarios.
    \item We extend masked attention to include foreground \emph{and} background for both rich features and a clean semantic separation between the target object and distractors.
    \item We construct a compact \emph{object memory} to summarize object features in the long term, which are retrieved as target-specific object-level representations during querying.
\end{itemize}

\section{Related Works}

\myparagraph{Memory-Based VOS.}
Since semi-supervised Video Object Segmentation (VOS) involves a directional propagation of information, many existing approaches employ a feature memory representation that stores past features for segmenting future frames.
This includes online learning that finetunes a network on the first-frame segmentation for every video during inference~\cite{caelles2017one,voigtlaender2017online,maninis2018video,bhat2020learning,robinson2020learning}. However, finetuning is slow during test-time.
Recurrent approaches~\cite{perazzi2017learning,hu2017maskrnn,hu2018motion,oh2018fast,wang2019fast,zhang2019fast,ventura2019rvos} are faster but lack context for tracking under occlusion. 
Recent approaches use more context~\cite{hu2018videomatch,voigtlaender2019feelvos,oh2019videoSTM,wang2019ranet,Duarte2019Capsule,yang2020collaborativeCFBI,li2020fastGlobalContext,zhang2020transductive,seong2020kernelizedMemory,lu2020videoGraphMem,Liang2020AFBURR,huang2020fast,liang2021video,xu2021reliable,ge2021video,hu2021learning,wang2021swiftnet,xie2021efficient,seong2021hierarchical,mao2021joint,cheng2021stcn,liu2022learning,yu2022batman,miao2022region,li2022recurrent,park2022per,liu2022global,zhang2023boosting,xu2022towards,cho2022pixel,xu2022accelerating,miles2023mobilevos,yan2023two,cheng2023tracking,bekuzarov2023xmem++,sun2023alignment} via pixel-level feature matching and integration, with some exploring the modeling of background features -- either explicitly~\cite{yang2020collaborativeCFBI,yang2021collaborativeplus} or implicitly~\cite{cheng2021stcn}.
XMem~\cite{cheng2022xmem} uses multiple types of memory for better performance and efficiency but still struggles with noise from low-level pixel matching.
While we adopt the memory reading of XMem~\cite{cheng2022xmem}, we develop an object reading mechanism to integrate the pixel features at an object level which permits Cutie to attain much better performance in challenging scenarios.

\myparagraph{Transformers in VOS.}
Transformer-based~\cite{vaswani2017attention} approaches have been developed for pixel matching with memory in video object segmentation~\cite{duke2021sstvos,yang2021associating,mao2021joint,yang2022decoupling,yu2022batman,wu2023scalable,zhang2023joint}. 
However, they compute attention between spatial feature maps (as cross-attention, self-attention, or both), which is computationally expensive with $O(n^4)$ time/space complexity, where $n$ is the image side length. 
SST~\cite{duke2021sstvos} proposes sparse attention but performs worse than state-of-the-art methods.
AOT approaches~\cite{yang2021associating,yang2022decoupling} use an identity bank for processing multiple objects in a single forward pass to improve efficiency, but are not permutation equivariant with respect to object ID and do not scale well to longer videos.
Concurrent approaches~\cite{wu2023scalable,zhang2023joint} use a single vision transformer network to jointly model the reference frames and the query frame without explicit memory reading operations. They attain high accuracy but require large-scale pretraining (e.g., MAE~\cite{he2021masked}) and have a much lower inference speed ($<4$ frames per second).
Cutie is carefully designed to \textit{not} compute any (costly) attention between spatial feature maps in our object transformer while facilitating efficient global communication via a small set of object queries -- allowing Cutie to be real-time.

\myparagraph{Object-Level Reasoning.}
Early VOS algorithms~\cite{li2018video,luiten2018premvos,xu2022towards} that attempt to reason at the object level use either re-identification or k-means clustering to obtain object features and have a lower performance on standard benchmarks.
HODOR~\cite{athar2022hodor}, and its follow-up work TarViS~\cite{athar2023tarvis}, approach VOS with object-level descriptors which allow for greater flexibility (e.g., training on static images only~\cite{athar2022hodor} or extending to different video segmentation tasks~\cite{athar2023tarvis,yan2023universal,yan2022towards}) but fall short on VOS segmentation accuracy (e.g., \cite{yan2023universal} is 6.9~\mjf~behind state-of-the-art methods in DAVIS 2017~\cite{perazzi2016benchmark}) due to under-using high-resolution features.
ISVOS~\cite{wang2022look} proposes to inject features from a pre-trained instance segmentation network (i.e., Mask2Former~\cite{cheng2022masked}) into a memory-based VOS method~\cite{cheng2021stcn}. Cutie has a similar motivation but is crucially different in three ways: 1) Cutie learns object-level information end-to-end, without needing to pre-train on instance segmentation tasks/datasets, 2) Cutie allows bi-directional communication between pixel-level features and object-level features for an integrated framework, and 3) Cutie is a one-stage method that does not perform separate instance segmentation while ISVOS does -- this allows Cutie to run six times (estimated) faster.
Moreover, ISVOS does not release code while we open source code for the community which facilitates follow-up work.

\myparagraph{Automatic Video Segmentation.}
Recently, video object segmentation methods have been used as an integral component in automatic video segmentation pipelines, such as open-vocabulary/universal video segmentation (e.g.,\ Tracking Anything~\cite{cheng2023tracking,yang2023track}, DEVA~\cite{cheng2023segment}) and unsupervised video segmentation~\cite{garg2021mask}.
We believe the robustness and efficiency of Cutie are beneficial for these applications.

\begin{figure*}
    \centering
    \includegraphics[width=0.95\linewidth]{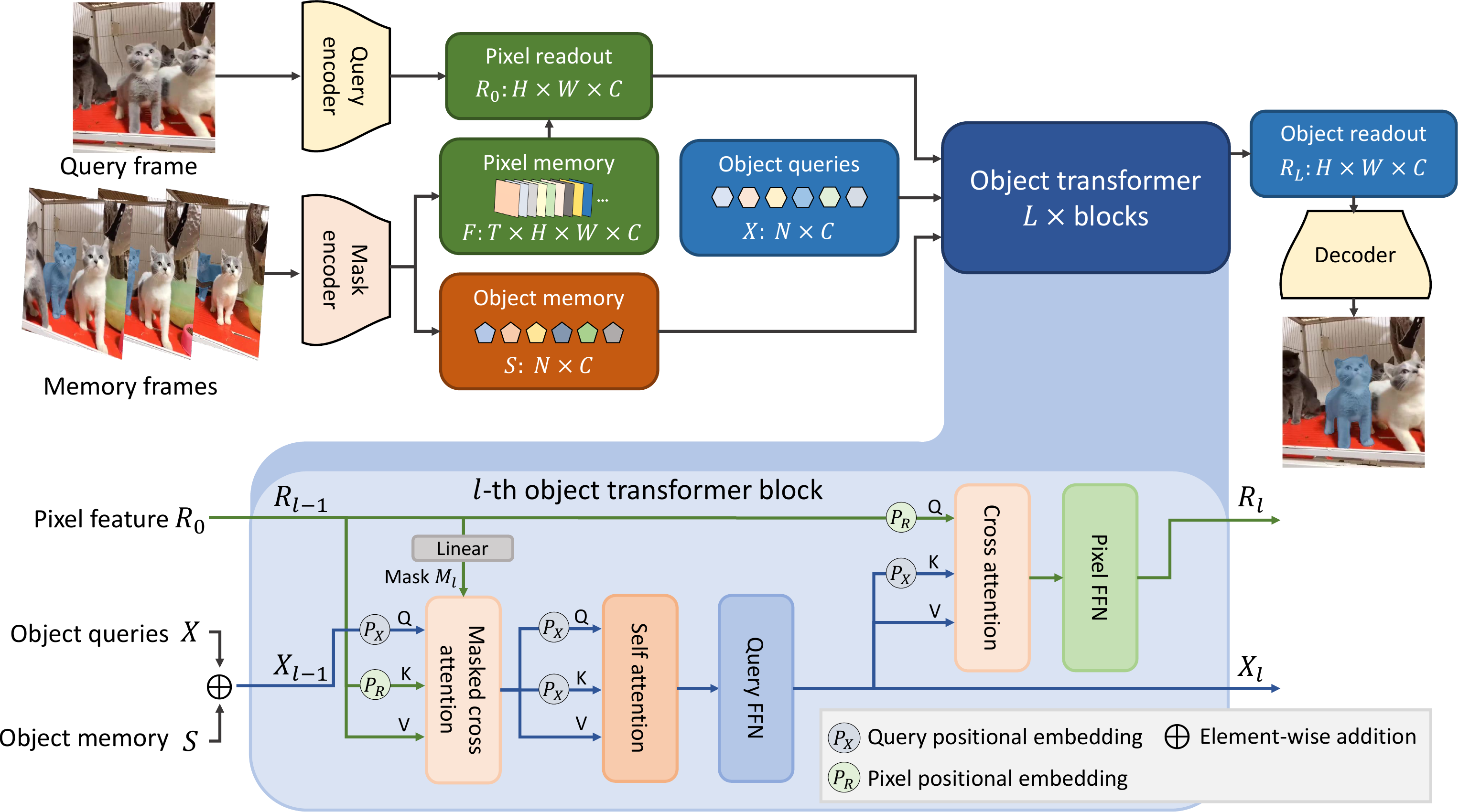}
    \caption{Overview of Cutie. We store pixel memory $F$ and object memory $S$ representations from past segmented (memory) frames. Pixel memory is retrieved for the query frame as pixel readout $R_0$, which bidirectionally interacts with object queries $X$ and object memory $S$ in the object transformer. The $L$ object transformer blocks enrich the pixel feature with object-level semantics and produce the final $R_L$ object readout for decoding into the output mask.
    Standard residual connections, layer normalization, and skip-connections from the query encoder to the decoder are omitted for readability.}
    \label{fig:overview}
\end{figure*}

\section{Cutie}

\subsection{Overview}
We provide an overview of Cutie in Figure~\ref{fig:overview}. 
For readability, following prior works~\cite{oh2019videoSTM,cheng2022xmem}, we consider a single target object as the extension to multiple objects is straightforward (see supplement).
Following the standard semi-supervised video object segmentation (VOS) setting, Cutie takes a first-frame segmentation of target objects as input and segments subsequent frames sequentially in a streaming fashion.
First, Cutie encodes segmented frames (given as input or segmented by the model) into a high-resolution pixel memory~$F$~(Section~\ref{sec:pix-mem}) and a high-level object memory~$S$~(Section~\ref{sec:obj-mem}) and stores them for segmenting future frames. 
To segment a new query frame, Cutie retrieves an initial pixel readout $R_0$ from the pixel memory using encoded query features.
This initial readout $R_0$ is computed via low-level pixel matching and is therefore often noisy. We enrich it with object-level semantics by augmenting $R_0$ with information from the object memory $S$ and a set of object queries $X$ through an \textit{object transformer} with $L$ transformer blocks~(Section~\ref{sec:obj-trans}).
The enriched output of the object transformer, $R_L$, or the object readout, is passed to the decoder for generating the final output mask.
In the following, we will first describe the three main contributions of Cutie: object transformer, masked attention, and object memory. 
Note, we derive the pixel memory from existing works~\cite{cheng2022xmem}, which we only describe as implementation details in Section~\ref{sec:pix-mem} without claiming any contribution.

\subsection{Object Transformer}\label{sec:obj-trans}

\subsubsection{Overview}
The bottom of Figure~\ref{fig:overview} illustrates the object transformer. The object transformer takes an initial readout $R_0\in\mathbb{R}^{HW\times C}$, a set of $N$ end-to-end trained object queries $X\in\mathbb{R}^{N\times C}$, and object memory $S\in\mathbb{R}^{N\times C}$ as input, and integrates them with $L$ transformer blocks. Note $H$ and $W$ are image dimensions after encoding with stride 16.
Before the first block, we sum the static object queries with the dynamic object memory for better adaptation, i.e., $X_0=X+S$.
Each transformer block bidirectionally allows the object queries $X_{l-1}$ to attend to the readout $R_{l-1}$, and vice versa, producing updated queries $X_l$ and readout $R_l$ as the output of the $l$-th block.
The last block's readout, $R_L$, is the final output of the object transformer.
 
Within each block, we first compute masked cross-attention, letting the object queries $X_{l-1}$ read from the pixel features $R_{l-1}$.
The masked attention focuses half of the object queries on the foreground region while the other half is targeted towards the background (details in Section~\ref{sec:masked-attn}). 
Then, we pass the object queries into standard self-attention and feed-forward layers~\cite{vaswani2017attention} for object-level reasoning. 
Next, we update the pixel features with a reversed cross-attention layer, \textit{putting the object} semantics from object queries $X_l$ \emph{back} into pixel features $R_{l-1}$. We then pass the pixel features into a feed-forward network while skipping the computationally expensive self-attention in a standard transformer~\cite{vaswani2017attention}.
Throughout, positional embeddings are added to the queries and keys following~\cite{carion2020end,cheng2022masked} (Section~\ref{sec:pos-emb}).
Residual connections and layer normalizations are used in every attention and feed-forward layer following~\cite{xiong2020layer}.
All attention layers are implemented with multi-head scaled dot product attention~\cite{vaswani2017attention}.
Importantly, 
\begin{enumerate}[topsep=0pt,itemsep=0pt,partopsep=0pt, parsep=0pt]
    \item We carefully avoid any direct attention between high-resolution spatial features (e.g., $R$), as they are intensive in both memory and compute. 
    Despite this, these spatial features can still interact globally via object queries, making each transformer block efficient and expressive.
    \item The object queries restructure the pixel features with a residual contribution without discarding the high-resolution pixel features. This avoids irreversible dimensionality reductions (would be over 100$\times$) and keeps those high-resolution features for accurate segmentation.
\end{enumerate}

Next, we describe the core components in our object transformer blocks: foreground/background masked attention and the construction of the positional embeddings.

\subsubsection{Foreground-Background Masked Attention}\label{sec:masked-attn}

\begin{figure*}
    \centering
    \begin{tabular}{c@{\hspace{1pt}}c@{\hspace{1pt}}c@{\hspace{1pt}}c@{\hspace{1pt}}c@{\hspace{1pt}}c}
\includegraphics[width=0.16\linewidth]{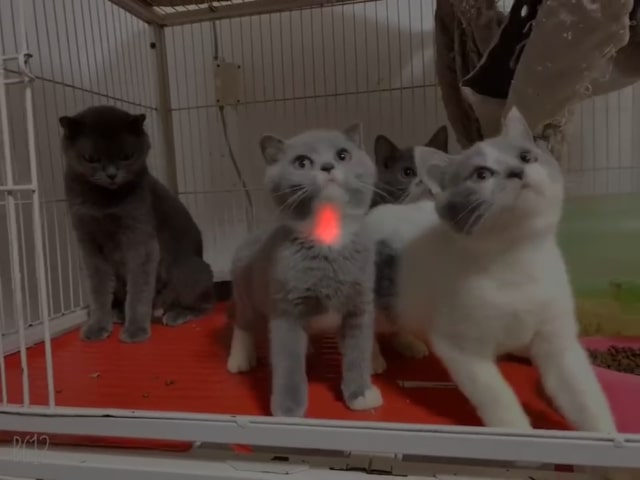} & 
      \includegraphics[width=0.16\linewidth]{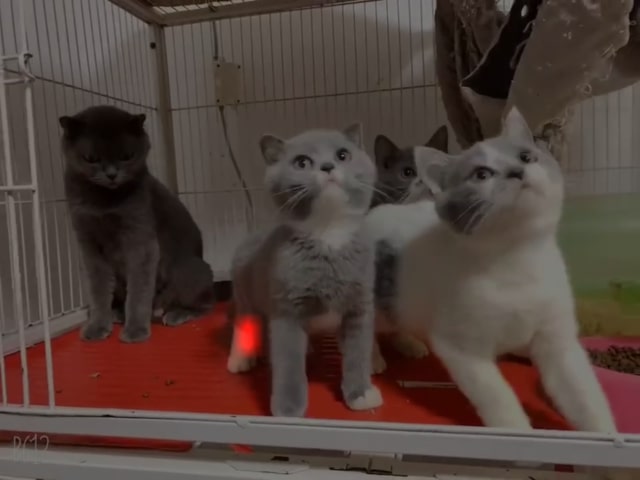} & 
      \fcolorbox{red}{red}{\includegraphics[width=0.16\linewidth]{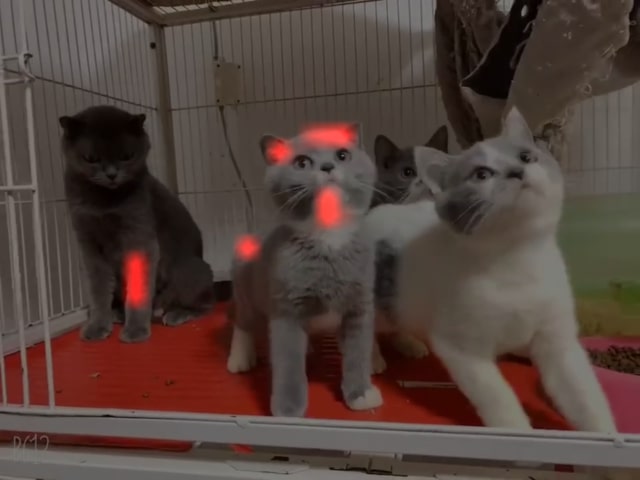}} & 
      \includegraphics[width=0.16\linewidth]{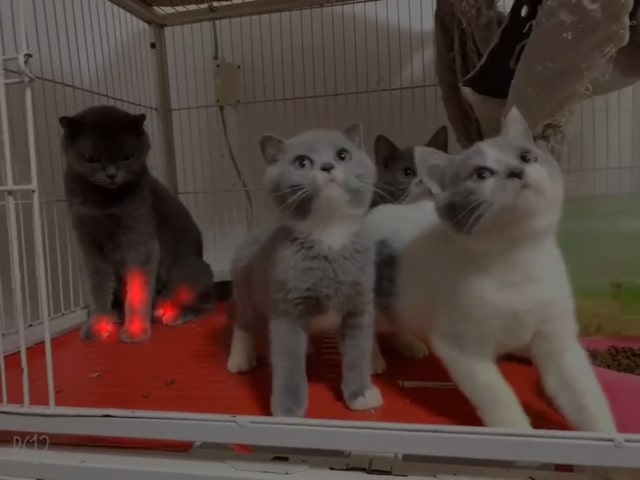} & 
      \fcolorbox{red}{red}{\includegraphics[width=0.16\linewidth]{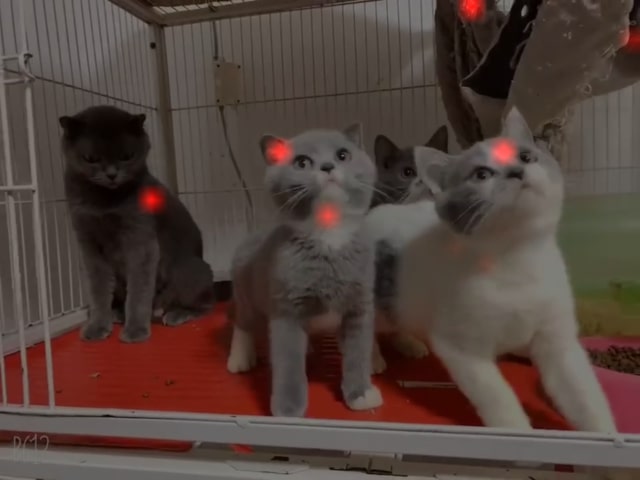}} & 
      \fcolorbox{red}{red}{\includegraphics[width=0.16\linewidth]{img/attn-nomask/00040_obj1_q15_head3.jpg}} \\
    \includegraphics[width=0.16\linewidth]{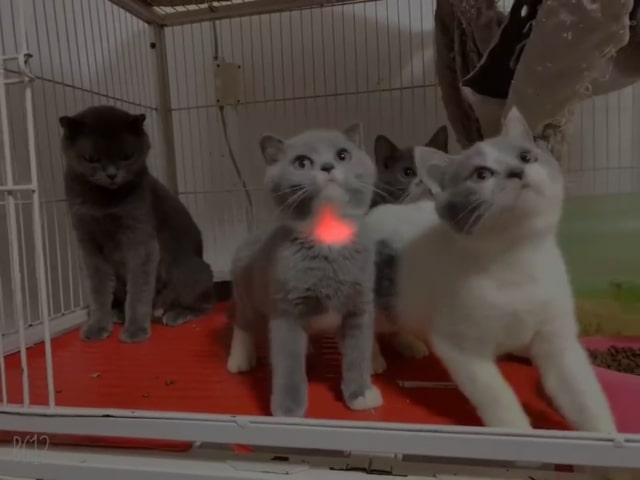} & 
    \includegraphics[width=0.16\linewidth]{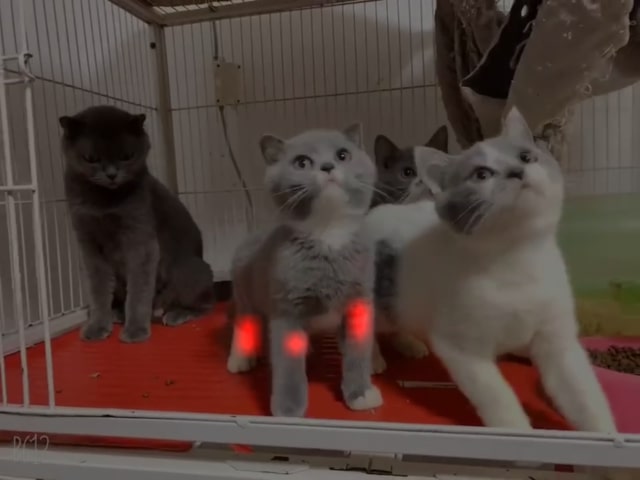} & 
      \includegraphics[width=0.16\linewidth]{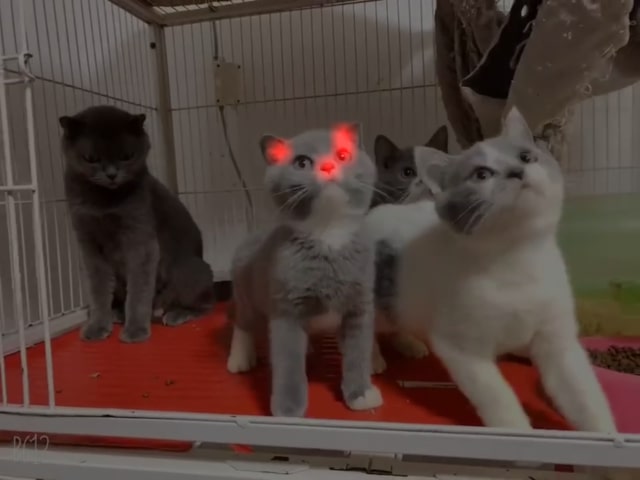} & 
      \includegraphics[width=0.16\linewidth]{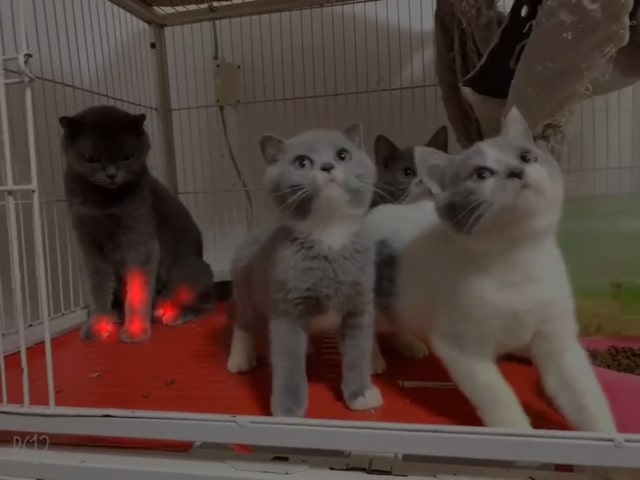} & 
      \includegraphics[width=0.16\linewidth]{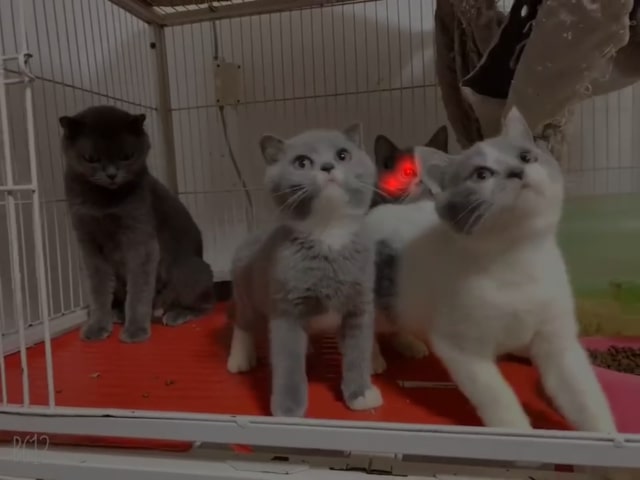} &
      \includegraphics[width=0.16\linewidth]{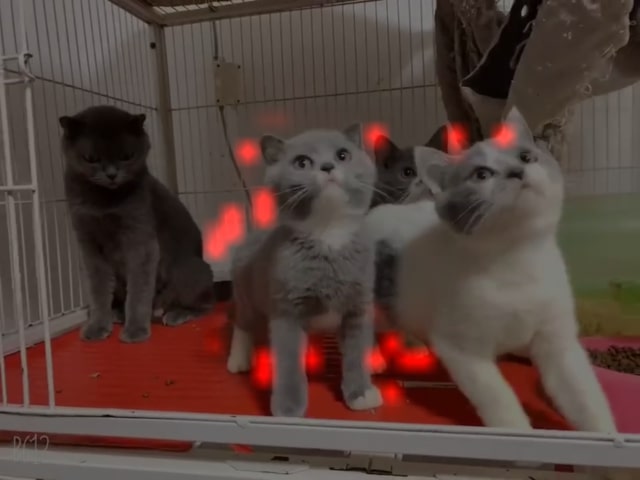} \\
 \end{tabular}

    \caption{Visualization of cross-attention weights (rows of $A_L$) in the object transformer.
    The middle cat is the target object.
    Top: without foreground-background masking -- some queries mix semantics from foreground and background (framed in red).
    Bottom: with foreground-background masking. The leftmost three are foreground queries, and the rightmost three are background queries. Semantics is thus cleanly separated. The f.g./b.g.\ queries can communicate in the subsequent self-attention layer.
    Note the queries attend to different foreground regions, distractors, and background regions.}
    \label{fig:fg-bg-query-attention}
\end{figure*}

In our (pixel-to-query) cross-attention, we aim to update the object queries $X_l\in\mathbb{R}^{N\times C}$ by attending over the pixel features $R_l\in\mathbb{R}^{HW\times C}$. Standard cross-attention with the residual path finds
\begin{equation}
 X_l' = A_l V_l + X_l = \text{softmax}(Q_l K_l^T) V_l + X_l, 
\end{equation}
where $Q_l$ is a learned linear transformation of $X_l$, and $K_l, V_l$ are learned linear transformations of $R_l$. The rows of the affinity matrix $A_l\in\mathbb{R}^{N\times HW}$ describe the attention of each object query over the entire feature map. 
We note that there are distinctly different attention patterns for different object queries -- some focus on different foreground parts, some on the background, and some on distractors (top of Figure~\ref{fig:fg-bg-query-attention}). These object queries collect information from different regions of interest and integrate them in subsequent self-attention/feed-forward layers. 
However, the soft nature of attention makes this process noisy and less reliable -- queries that mainly attend to the foreground might have small weights distributed in the background and vice versa.
Inspired by~\cite{cheng2022masked}, we deploy masked attention to aid the clean separation of semantics between foreground and background. 
Different from~\cite{cheng2022masked}, which only attends to the foreground, we find it helpful to also attend to the background, especially in challenging tracking scenarios with distractors. 
In practice, we let the first half of the object queries (i.e., foreground queries) always attend to the foreground and the second half (i.e., background queries) attend to the background.
This masking is shared across all attention heads.

Formally, our foreground-background masked cross-attention finds
\begin{equation}
 X_l' = \text{softmax}(\mathcal{M}_l + Q_l K_l^T) V_l + X_l, 
\end{equation}
where $\mathcal{M}_l\in\{ 0, -\infty \}^{N\times HW}$ controls the attention masking -- specifically,  $\mathcal{M}_l(q,i)$ determines whether the $q$-th query is allowed ($=0$) or not allowed ($=-\infty$) to attend to the $i$-th pixel.
To compute $\mathcal{M}_l$, we first find a mask prediction at the current layer $M_l$, which is linearly projected from the last pixel feature $R_{l-1}$ and activated with the sigmoid function. Then, $\mathcal{M}_l$ is computed as
\begin{equation}
    \mathcal{M}_l(q, i) = 
    \begin{cases}
    0, \quad &\text{if } q\leq N/2 \text{ and } M_l(i)\geq0.5 \\
    0, \quad &\text{if } q> N/2 \text{ and } M_l(i)<0.5 \\
    -\infty, \quad &\text{otherwise }  \\
    \end{cases}, 
\end{equation}
where the first case is for foreground attention and the second is for background attention.
Figure~\ref{fig:fg-bg-query-attention} (bottom) visualizes the attention maps after this foreground-background masking. Note, despite the hard foreground-background separation, the object queries communicate in the subsequent self-attention layer for potential global feature interaction.
Next, we discuss the positional embeddings used in object queries and pixel features that allow location-based attention.

\subsubsection{Positional Embeddings}\label{sec:pos-emb}
Since vanilla attention operations are permutation equivariant, positional embeddings are used to provide additional features about the position of each token~\cite{vaswani2017attention}. 
Following prior transformer-based vision networks~\cite{carion2020end,cheng2022masked}, we add the positional embedding to the query and key features at every attention layer (Figure~\ref{fig:overview}), and not to the value.

For the object queries, we use a positional embedding $P_X\in\mathbb{R}^{N\times C}$ that combines an end-to-end learnable embedding $E_X\in\mathbb{R}^{N\times C}$ and the dynamic object memory $S\in\mathbb{R}^{N\times C}$  via
\begin{equation}
    P_X = E_X + f_\text{ObjEmbed}(S), 
\end{equation}
where $f_\text{ObjEmbed}$ is a trainable linear projection.

For the pixel feature, the positional embedding $P_R\in\mathbb{R}^{HW\times C}$ combines a fixed 2D sinusoidal positional embedding $R_{\sin}$~\cite{carion2020end} that encodes absolute pixel coordinates and the initial readout $R_0\in\mathbb{R}^{HW\times C}$ via 
\begin{equation}
    P_R = R_{\sin} + f_\text{PixEmbed}(R_0), 
\end{equation}
where $f_\text{PixEmbed}$ is another trainable linear projection. Note that the sinusoidal embedding $R_{\sin}$ operates on normalized coordinates and is scaled accordingly to different image sizes at test time.

\subsection{Object Memory}\label{sec:obj-mem}

In the object memory $S\in\mathbb{R}^{N\times C}$, we store a compact set of $N$ vectors which make up a high-level summary of the target object.
This object memory is used in the object transformer (Section~\ref{sec:obj-trans}) to provide target-specific features.
At a high level, we compute $S$ by mask-pooling over all encoded object features with $N$ different masks. Concretely, given object features $U\in \mathbb{R}^{THW\times C}$ and $N$ pooling masks $\{W_q\in [0, 1]^{THW}, 0<q\leq N\}$, where $T$ is the number of memory frames, the $q$-th object memory $S_q\in\mathbb{R}^C$ is computed by
\begin{equation}
    S_q = \frac{\sum^{THW}_{i=1}{U(i) W_q(i)}}{\sum^{THW}_{i=1}{W_q(i)}}.
\end{equation}
During inference, we use a classic streaming average algorithm such that this operation takes constant time and memory with respect to the video length. See the supplement for details.
Note, an object memory vector $S_q$ would not be modified if the corresponding pooling weights are zero, i.e., $\sum^{HW}_{i=1}W^t_q(i)=0$, preventing feature drifting when the corresponding object region is not visible (e.g., occluded).

To find $U$ and $W$ for a memory frame, we first encode the corresponding image $I$ and the segmentation mask $M$ with the mask encoder for memory feature $F\in \mathbb{R}^{THW\times C}$.
We use a 2-layer, $C$-dimensional MLP $f_\text{ObjFeat}$ to obtain the object feature $U$ via
\begin{equation}
    U = f_\text{ObjFeat} (F).
\end{equation}
For the $N$ pooling masks $\{W_q\in [0, 1]^{THW}, 0<q\leq N\}$, we additionally apply foreground-background separation as detailed in Section~\ref{sec:masked-attn} and augment it with a fixed 2D sinusoidal positional embedding $R_{\sin}$ (as mentioned in Section~\ref{sec:pos-emb}). The separation allows it to aggregate clean semantics during pooling, while the positional embedding enables location-aware pooling.
Formally, we compute the $i$-th pixel of the $q$-th pooling mask via
\begin{equation}
    W_q(i) = 
    \begin{cases}
    0, \quad \text{if } q\leq N/2 \text{ and } M(i)<0.5 \\
    0, \quad \text{if } q > N/2 \text{ and } M(i)\geq 0.5 \\
    \sigma(f_{\text{PoolWeight}}(F(i) + R_{\sin}(i))), \quad \text{otherwise}  \\
    \end{cases}, 
\end{equation}
where $\sigma$ is the sigmoid function, $f_\text{PoolWeight}$ is a 2-layer, $N$-dimensional MLP, and the segmentation mask $M$ is downsampled to match the feature stride of $F$.

\subsection{Implementation Details}

\subsubsection{Pixel Memory}\label{sec:pix-mem}
Our pixel memory design, which provides the pixel feature $R_0$ (see Figure~\ref{fig:overview}), is derived from XMem~\cite{cheng2022xmem,cheng2023tracking} working and sensory memory. We do not claim contributions.
Here, we present the high-level algorithm and defer details to the supplementary material.
The pixel memory is composed of an attentional component (with keys $\mathbf{k}\in\mathbb{R}^{THW\times C^\text{k}}$ and values $\mathbf{v}\in\mathbb{R}^{THW\times C}$) and a recurrent component (with hidden state $\mathbf{h}^{HW\times C}$).
Long-term memory~\cite{cheng2022xmem} can be optionally included in the attentional component without re-training for better performance on long videos.
The keys and values consist of low-level appearance features for matching while the hidden state provides temporally consistent features. 
To retrieve a pixel readout $R_0$, we first encode the query frame to obtain query feature $\mathbf{q}^{HW\times C}$, and compute the query-to-memory affinity $A^\text{pix}\in[0,1]^{HW\times THW}$ via
\begin{equation}
    A^\text{pix}_{ij} = \frac{\exp\left(d(\mathbf{q}_i, \mathbf{k}_j)\right)}{ \sum_m  \exp\left(d(\mathbf{q}_i, \mathbf{k}_m)\right)}, 
\end{equation}
where $d(\cdot,\cdot)$ is the anisotropic L2 function~\cite{cheng2022xmem} which is proportional to the similarity between the two inputs. 
Finally, we find the pixel readout $R_0$ by combining the attention readout with the hidden state:
\begin{equation}
    R_0 = f_{\text{fuse}}\left(A^\text{pix}\mathbf{v} + \mathbf{h}\right), 
\end{equation}
where $f_{\text{fuse}}$ is a small network consisting of two $C$-dimension convolutional residual blocks with channel attention~\cite{wang2019eca}.

\subsubsection{Network Architecture}
We study two model variants: `small' and `base' with different query encoder backbones, otherwise sharing the same configuration: $C=256$ channels with $L=3$ object transformer blocks and $N=16$ object queries. 

\myparagraph{ConvNets.}
We parameterize the query encoder and the mask encoder with ResNets~\cite{he2016deepResNet}. Following~\cite{oh2019videoSTM,cheng2022xmem}, we discard the last convolutional stage and use the stride 16 feature. 
For the query encoder, we use ResNet-18 for the small model and ResNet-50 for the base model. 
For the mask encoder, we use ResNet-18.
`Cutie-base' thus shares the same backbone configuration as XMem.
We find that Cutie works well with a lighter decoder -- we use a similar iterative upsampling architecture as in XMem but halve the number of channels in all upsampling blocks for better efficiency.

\myparagraph{Feed-Forward Networks (FFN).} 
We use query FFN and pixel FFN in our object transformer block (Figure~\ref{fig:overview}). 
For the query FFN, we use a 2-layer MLP with a hidden size of $8C=2048$. 
For the pixel FFN, we use two $3\times3$ convolutions with a smaller hidden size of $C=256$ to reduce computation. 
As we do not use self-attention on the pixel features, we compensate by using efficient channel attention~\cite{wang2019eca} after the second convolution of the pixel FFN.
Layer normalizations are applied to the query FFN following~\cite{xiong2020layer} and not to the pixel FFN, as we observe no empirical benefits. 
ReLU is used as the activation function.

\subsubsection{Training}
\myparagraph{Data.}
Following~\cite{oh2019videoSTM,yang2021associating,cheng2022xmem}, we first pretrain our network on static images~\cite{shi2015hierarchicalECSSD,wang2017DUTS,li2020fss,zeng2019towardsHRSOD,cheng2020cascadepsp} by generating three-frame sequences with synthetic deformation.
Next, we perform the main training on video datasets DAVIS~\cite{perazzi2016benchmark} and YouTubeVOS~\cite{xu2018youtubeVOS} by sampling eight frames following~\cite{cheng2022xmem}.
We optionally also train on MOSE~\cite{ding2023mose} (combined with DAVIS and YouTubeVOS), as we notice the training sets of YouTubeVOS and DAVIS have become too easy for our model to learn from (>93\% IoU during training). 
For every setting, we use one trained model and do not finetune for specific datasets.
We additionally introduce a `MEGA' setting with BURST~\cite{athar2023burst} and OVIS~\cite{qi2022occluded} included in training (+1.6~\mjf~in MOSE).
Details are provided in the supplementary material.

\myparagraph{Optimization.}
We use the AdamW~\cite{loshchilov2017decoupledAdamW} optimizer with a learning rate of $1\mathrm{e}{-4}$, a batch size of 16, and a weight decay of 0.001.
Pretraining lasts for 80K iterations with no learning rate decay. 
Main training lasts for 125K iterations, with the learning rate reduced by 10 times after 100K and 115K iterations. 
The query encoder has a learning rate multiplier of 0.1 following~\cite{cheng2022masked,yang2021associating,cheng2023tracking} to mitigate overfitting.
Following the bag of tricks from DEVA~\cite{cheng2023tracking}, we clip the global gradient norm to 3 throughout and use stable data augmentation.
The entire training process takes approximately 30 hours on four A100 GPUs for the small model.

\myparagraph{Losses.}
Following~\cite{cheng2022masked}, we adopt point supervision which computes the loss only at $K$ sampled points instead of the whole mask.
We use importance sampling~\cite{kirillov2020pointrend} and set $K=8192$ during pretraining and $K=12544$ during main training.
We use a combined loss function of cross-entropy and soft dice loss with equal weighting following~\cite{cheng2022xmem,yang2021associating,cheng2023tracking}.
In addition to the loss applied to the final segmentation output, we adopt auxiliary losses in the same form (scaled by 0.01) to the intermediate masks $M_l$ in the object transformer.

\subsubsection{Inference}\label{sec:inference}
During testing, we encode a memory frame for updating the pixel memory and the object memory every $r$-th frame. $r$ defaults to 5 following~\cite{cheng2022xmem}.
For the keys $\mathbf{k}$ and values $\mathbf{v}$ in the attention component of the pixel memory, we always keep features from the first frame (as it is given by the user) and use a First-In-First-Out (FIFO) approach for other memory frames to ensure the total number of memory frames $T$ is less than or equal to a pre-defined limit $T_{\max}=5$. 
For processing long videos (e.g., BURST~\cite{athar2023burst} or LVOS~\cite{hong2022lvos} with over a thousand frames per video), we use the long-term memory~\cite{cheng2022xmem} instead of FIFO without re-training, following the default parameters in~\cite{cheng2022xmem}.
For the pixel memory, we use top-$k$ filtering~\cite{cheng2021mivos} with $k=30$.
Inference is fully online, can be streamed, and uses a constant amount of compute per frame and memory with respect to the sequence length.

\section{Experiments}
For evaluation, we use standard metrics: Jaccard index \mj, contour accuracy \mf, and their average \mjf~\cite{perazzi2016benchmark}.
In YouTubeVOS~\cite{xu2018youtubeVOS}, \mj~and \mf~are computed for ``seen'' and ``unseen'' categories separately. \mg~is the averaged \mjf~for both seen and unseen classes.
For BURST~\cite{athar2023burst}, we assess Higher Order Tracking Accuracy (HOTA)~\cite{luiten2021hota} on common and uncommon object classes separately.
For our models, unless otherwise specified, we resize the inputs such that the shorter edge has no more than 480 pixels and rescale the model's prediction back to the original resolution.

\begin{table*}
    \centering
\begin{NiceTabular}
{l@{\hspace{10pt}}C{2.2em}@{}C{2.2em}@{}C{2.2em}@{\hspace{5pt}}C{2.2em}@{}C{2.2em}@{}C{2.2em}@{\hspace{5pt}}C{2.2em}@{}C{2.2em}@{}C{2.2em}@{\hspace{5pt}}C{2.2em}@{}C{2.2em}@{}C{2.2em}@{}C{2.2em}@{}C{2.2em}@{}R{2em}}[colortbl-like]
\toprule
& \multicolumn{3}{c}{\small MOSE} & \multicolumn{3}{c}{\small DAVIS-17 val} & \multicolumn{3}{c}{\small DAVIS-17 test} & \multicolumn{6}{c}{\small YouTubeVOS-2019 val} \\
\cmidrule(lr{\dimexpr 4\tabcolsep+8pt}){2-5} \cmidrule(lr{\dimexpr 4\tabcolsep+8pt}){5-8} \cmidrule(lr{\dimexpr 4\tabcolsep+8pt}){8-11} \cmidrule(lr){11-16}
Method & \mjf & \mj & \mf & \mjf & \mj & \mf & \mjf & \mj & \mf & \mg & \mjs & \mfs & \mju & \mfu & FPS \\
\toprule
\multicolumn{10}{l}{\textbf{\textit{Trained without MOSE}}} \\
\midrule
STCN~\cite{cheng2021stcn} & 52.5 & 48.5 & 56.6 & 85.4 & 82.2 & 88.6 & 76.1 & 72.7 & 79.6 & 82.7 & 81.1 & 85.4 & 78.2 & 85.9 & 13.2 \\
AOT-R50~\cite{yang2021associating} & 58.4 & 54.3 & 62.6 & 84.9 & 82.3 & 87.5 & 79.6 & 75.9 & 83.3 & 85.3 & 83.9 & 88.8 & 79.9 & 88.5 & 6.4 \\
RDE~\cite{li2022recurrent} & 46.8 & 42.4 & 51.3 & 84.2 & 80.8 & 87.5 & 77.4 & 73.6 & 81.2 & 81.9 & 81.1 & 85.5 & 76.2 & 84.8 & 24.4 \\
XMem~\cite{cheng2022xmem} & 56.3 & 52.1 & 60.6 & 86.2 & 82.9 & 89.5 & 81.0 & 77.4 & 84.5 & 85.5 & 84.3 & 88.6 & 80.3 & 88.6 & 22.6 \\
DeAOT-R50~\cite{yang2022decoupling} & 59.0 & 54.6 & 63.4 & 85.2 & 82.2 & 88.2 & 80.7 & 76.9 & 84.5 & 85.6 & 84.2 & 89.2 & 80.2 & 88.8 & 11.7 \\
\color{gray}
SimVOS-B~\cite{wu2023scalable} & \color{gray}- & \color{gray}- & \color{gray}- & \color{gray}81.3 & \color{gray}78.8 & \color{gray}83.8 & \color{gray}- & \color{gray}- & \color{gray}- & \color{gray}- & \color{gray}- & \color{gray}- & \color{gray}- & \color{gray}- & \color{gray}3.3 \\
\color{gray}
JointFormer~\cite{zhang2023joint} & \color{gray}- & \color{gray}- & \color{gray}- & \color{gray}- & \color{gray}- & \color{gray}- & \color{gray}65.6 & \color{gray}61.7 & \color{gray}69.4 & \color{gray}73.3 & \color{gray}75.2 & \color{gray}78.5 & \color{gray}65.8 & \color{gray}73.6 & \color{gray}3.0 \\
\color{gray}
ISVOS~\cite{wang2022look} & \color{gray}- & \color{gray}- & \color{gray}- & \color{gray}80.0 & \color{gray}76.9 & \color{gray}83.1 & \color{gray}- & \color{gray}- & \color{gray}- & \color{gray}- & \color{gray}- & \color{gray}- & \color{gray}- & \color{gray}- & \color{gray}5.8$^\ast$ \\
DEVA~\cite{cheng2023tracking} & 60.0 & 55.8 & 64.3 & 86.8 & 83.6 & 90.0 & 82.3 & 78.7 & 85.9 & 85.5 & 85.0 & 89.4 & 79.7 & 88.0 & 25.3 \\
\rowcolor{defaultColor}
Cutie-small & 62.2 & 58.2 & 66.2 & 87.2 & 84.3 & 90.1 & 84.1 & 80.5 & 87.6 & \textbf{86.2} & 85.3 & 89.6 & \textbf{80.9} & \textbf{89.0} & \textbf{45.5} \\
\rowcolor{defaultColor}
Cutie-base & \textbf{64.0} & \textbf{60.0} & \textbf{67.9} & \textbf{88.8} & \textbf{85.4} & \textbf{92.3} & \textbf{84.2} & \textbf{80.6} & \textbf{87.7} & 86.1 & \textbf{85.5} & \textbf{90.0} & 80.6 & 88.3 & 36.4 \\
\toprule
\multicolumn{10}{l}{\textbf{\textit{Trained with MOSE}}} \\
\midrule
XMem~\cite{cheng2022xmem} & 59.6 & 55.4 & 63.7 & 86.0 & 82.8 & 89.2 & 79.6 & 76.1 & 83.0 & 85.6 & 84.1 & 88.5 & 81.0 & 88.9 & 22.6 \\
DeAOT-R50~\cite{yang2022decoupling} & 64.1 & 59.5 & 68.7 & 86.0 & 83.1 & 88.9 & 82.8 & 79.1 & 86.5 & 85.3 & 84.2 & 89.0 & 79.9 & 88.2 & 11.7 \\
DEVA~\cite{cheng2023tracking} & 66.0 & 61.8 & 70.3 & 87.0 & 83.8 & 90.2 & 82.6 & 78.9 & 86.4 & 85.4 & 84.9 & 89.4 & 79.6 & 87.8 & 25.3 \\
\rowcolor{defaultColor}
Cutie-small & 67.4 & 63.1 & 71.7 & 86.5 & 83.5 & 89.5 & 83.8 & 80.2 & 87.5 & 86.3 & 85.2 & 89.7 & 81.1 & 89.2 & \textbf{45.5} \\
\rowcolor{defaultColor}
Cutie-base & \textbf{68.3} & \textbf{64.2} & \textbf{72.3} & \textbf{88.8} & \textbf{85.6} & \textbf{91.9} & \textbf{85.3} & \textbf{81.4} & \textbf{89.3} & \textbf{86.5} & \textbf{85.4} & \textbf{90.0} & \textbf{81.3} & \textbf{89.3} & 36.4 \\
\midrule
\bottomrule
\end{NiceTabular}

    \caption{Quantitative comparison on video object segmentation benchmarks. All algorithms with available code are re-run on our hardware for a fair comparison.
    We could not obtain the code for~\cite{wang2022look,wu2023scalable,zhang2023joint} at the time of writing, and thus they cannot be reproduced on datasets that they do not report results on. 
    For a fair comparison, all methods in this table use ImageNet~\cite{deng2009imagenet} pre-training only or are trained from scratch. 
    We compare methods with external pre-training (e.g., MAE~\cite{he2021masked} pre-training) in the supplement.
    $^\ast$estimated FPS.
    }
    \label{tab:main-results}
\end{table*}

\begin{table}
    \centering
\small
\begin{NiceTabular}{l@{\space}l@{\hspace{3pt}}c@{\hspace{3pt}}c@{\hspace{3pt}}c@{\hspace{6pt}}c@{\hspace{3pt}}c@{\hspace{3pt}}c@{\hspace{6pt}}c}[colortbl-like]
    \toprule
     & & \multicolumn{3}{c}{BURST val} & \multicolumn{3}{c}{BURST test} & \\
     \cmidrule(lr{\dimexpr 4\tabcolsep+3pt}){3-6}
     \cmidrule(lr{\dimexpr 4\tabcolsep+12pt}){6-9}
     Method && All & Com. & Unc. & All & Com. & Unc. & Mem. \\
     \midrule
     DeAOT~\cite{yang2022decoupling} & FIFO & 51.3 & 56.3 & 50.0 & 53.2 & 53.5 & 53.2 & 10.8G \\
     DeAOT~\cite{yang2022decoupling} & INF & 56.4 & 59.7 & 55.5 & 57.9 & 56.7 & 58.1 & 34.9G \\
     XMem~\cite{cheng2022xmem} & FIFO & 52.9 & 56.0 & 52.1 & 55.9 & 57.6 & 55.6 & 3.03G \\
     XMem~\cite{cheng2022xmem} & LT & 55.1 & 57.9 & 54.4 & 58.2 & 59.5 & 58.0 & 3.34G \\
     \rowcolor{defaultColor}
     Cutie-small & FIFO & 56.8 & 61.1 & 55.8 & 61.1 & 62.4 & 60.8 & \textbf{1.35G} \\
     \rowcolor{defaultColor}
     Cutie-small & LT & 58.3 & 61.5 & \textbf{57.5} & 61.6 & 63.1 & 61.3 & 2.28G \\
     \rowcolor{defaultColor}
     Cutie-base & LT & \textbf{58.4} & \textbf{61.8} & \textbf{57.5} & \textbf{62.6} & \textbf{63.8} & \textbf{62.3} & 2.36G \\
     \midrule
     \bottomrule
\end{NiceTabular}

    \caption{Comparisons of performance on long videos on the BURST dataset~\cite{athar2023burst}. 
    Mem.: maximum GPU memory usage. FIFO: first-in-first-out memory bank; INF: unbounded memory; LT: long-term memory~\cite{cheng2022xmem}. DeAOT~\cite{yang2022decoupling} is not compatible with long-term memory. All methods are trained with the MOSE~\cite{ding2023mose} dataset.}
    \label{tab:burst-results}
\end{table}

\subsection{Main Results}
We compare with several state-of-the-art approaches on recent standard benchmarks: DAVIS 2017 validation/test-dev~\cite{perazzi2016benchmark} and YouTubeVOS validation~\cite{xu2018youtubeVOS}.
To assess the robustness of VOS algorithms, we also report results on MOSE validation~\cite{ding2023mose}, which contains heavy occlusions and crowded environments for evaluation.
DAVIS 2017~\cite{perazzi2016benchmark} contains annotated videos at 24 frames per second (fps), while YouTubeVOS contains videos at 30fps but is only annotated at 6fps. 
For a fair comparison, we evaluate all algorithms at full fps whenever possible, which is crucial for video editing and for having a smooth user-interaction experience.
For this, we re-run (De)AOT~\cite{yang2021associating,yang2022decoupling} with their official code at 30fps on YouTubeVOS. 
We also retrain XMem~\cite{cheng2022xmem}, DeAOT~\cite{yang2022decoupling}, and DEVA~\cite{cheng2023tracking} with their official code to include MOSE as training data (in addition to YouTubeVOS and DAVIS).
For long video evaluation, we test on BURST~\cite{athar2023burst} and LVOS~\cite{hong2022lvos} and experiment with the long-term memory~\cite{cheng2022xmem} in addition to our default FIFO memory strategy.
See supplement for details.
We compare with DeAOT~\cite{yang2022decoupling} and XMem~\cite{cheng2022xmem} under the same setting.

Table~\ref{tab:main-results} and Table~\ref{tab:burst-results} list our findings. 
Our method is highlighted with \colorbox{defaultColor}{lavender}.
FPS is recorded on YouTubeVOS with a V100.
Results on YouTubeVOS-18 and LVOS~\cite{hong2022lvos} are provided in the supplement.
Cutie achieves better results than state-of-the-art methods, especially on the challenging MOSE dataset, while remaining efficient.

\subsection{Ablations}
Here, we study various design choices of our algorithm. We use the small model variant with MOSE~\cite{ding2023mose} training data.
We highlight our default configuration with \colorbox{defaultColor}{lavender}.
For ablations, we report the $\mathcal{J}\&\mathcal{F}$ for MOSE validation and FPS on YouTubeVOS-2019 validation when applicable.
Due to resource constraints, we train a selected subset of ablations three times with different random seeds and report mean$\pm$std. The baseline is trained five times. 
In tables that do not report std, we present our performance with the default random seed only.

\begin{table*}[t]
\vspace{0pt}
\begin{minipage}[t]{0.25\linewidth}
\vspace{0pt}
\centering
\begin{tabular}{lcc}
    \toprule
     Setting & \mjf & FPS  \\
     \toprule
     \multicolumn{3}{l}{\textit{Number of transformer blocks}} \\
     \midrule
     $L=0$ & 65.2 & \textbf{56.6} \\
     $L=1$ & 66.0 & 51.1 \\
     \rowcolor{defaultColor}
     $L=3$ & 67.4 & 45.5 \\
     $L=5$ & \textbf{67.8} & 37.1 \\
     \toprule
     \multicolumn{3}{l}{\textit{Number of object queries}} \\
     \midrule
     $N=8$ & \textbf{67.6} & \textbf{45.5} \\
     \rowcolor{defaultColor}
     $N=16$ & \textbf{67.4} & \textbf{45.5} \\
     $N=32$ & 67.2 & \textbf{45.5} \\
     \toprule
     \multicolumn{3}{l}{\textit{Memory interval}} \\
     \midrule
     $r=3$ & \textbf{68.9} & 43.2\\
     \rowcolor{defaultColor}
     $r=5$ & 67.4 & 45.5 \\
     $r=7$ & 67.0 & \textbf{46.4} \\
     \toprule
     \multicolumn{3}{l}{\textit{Max. memory frames}} \\
     \midrule
     $T_{\max}=3$ & 66.9 & \textbf{48.5} \\
     \rowcolor{defaultColor}
     $T_{\max}=5$ & \textbf{67.4} & 45.5 \\
     $T_{\max}=10$ & \textbf{67.6} & 37.4 \\
     \midrule
     \bottomrule
\end{tabular}

\caption{Performance comparison with different choices of hyperparameters. }
\label{tab:hyperparameter}
\end{minipage}
\quad
\begin{minipage}[t]{0.7\linewidth}
\vspace{0pt}
\begin{minipage}[t]{0.45\linewidth}
\vspace{0pt}
\centering
\begin{tabular}{lcc}
    \toprule
     Setting & \mjf & FPS  \\
     \midrule
     \rowcolor{defaultColor}
     Both & \textbf{67.3}\pmnum{0.36} & 45.5 \\
     Bottom-up only & 65.0\pmnum{0.44} & \textbf{56.6} \\
     Top-down only & 40.7\pmnum{1.62}  & 46.9 \\
     \midrule
     \bottomrule
\end{tabular}

\caption{Comparison of our approach with bottom-up-only (no object transformer) and top-down-only (no pixel memory).}
\label{tab:bottom-up-top-down}
\end{minipage}
\quad
\begin{minipage}[t]{0.5\linewidth}
    \vspace{0pt}
\centering
\begin{tabular}{lc}
    \toprule
     Setting & \mjf  \\
     \midrule
     \rowcolor{defaultColor}
     With both & \textbf{67.3}\pmnum{0.36}\\
     No object memory ($X$) & 66.9\pmnum{0.26} \\
     No object query ($S$) & \textbf{67.2}\pmnum{0.10} \\
     \midrule
     \bottomrule
\end{tabular}

\caption{Ablations on the dynamic object memory and the static object query. Running times are similar.}
\label{tab:object-block-config-2}
\end{minipage}
\begin{minipage}[t]{0.55\linewidth}
    \vspace{0pt}
\centering
\begin{tabular}{lcc}
    \toprule
     Setting & \mjf & FPS  \\
     \midrule
     \rowcolor{defaultColor}
     f.g.-b.g.\ masked attn. & \textbf{67.3}\pmnum{0.36} & 45.5 \\
     f.g.\ masked attn. only & 66.7\pmnum{0.21} & 45.5 \\
     No masked attn. & 63.8\pmnum{1.06} & \textbf{46.3} \\
     \midrule
     \bottomrule
\end{tabular}

\caption{Ablations on foreground-background masked attention and object memory.}
\label{tab:object-block-config}
\end{minipage}
\begin{minipage}[t]{0.4\linewidth}
    \vspace{0pt}
\centering
\begin{tabular}{lcc}
    \toprule
     Setting & \mjf \\
     \midrule
     \rowcolor{defaultColor}
     With both p.e. & \textbf{67.4}\\
     Without query p.e. & 66.5 \\
     Without pixel p.e. & 66.2 \\
     With neither & 66.1 \\
     \midrule
     \bottomrule
\end{tabular}

\caption{Ablations on positional embeddings. Running times are similar.}
\label{tab:positional-embeddings}
\end{minipage}
\small
\begin{tabular}{c@{\hspace{1pt}}c@{\hspace{1pt}}c@{\hspace{1pt}}c@{\hspace{1pt}}c}
     \includegraphics[width=0.19\linewidth]{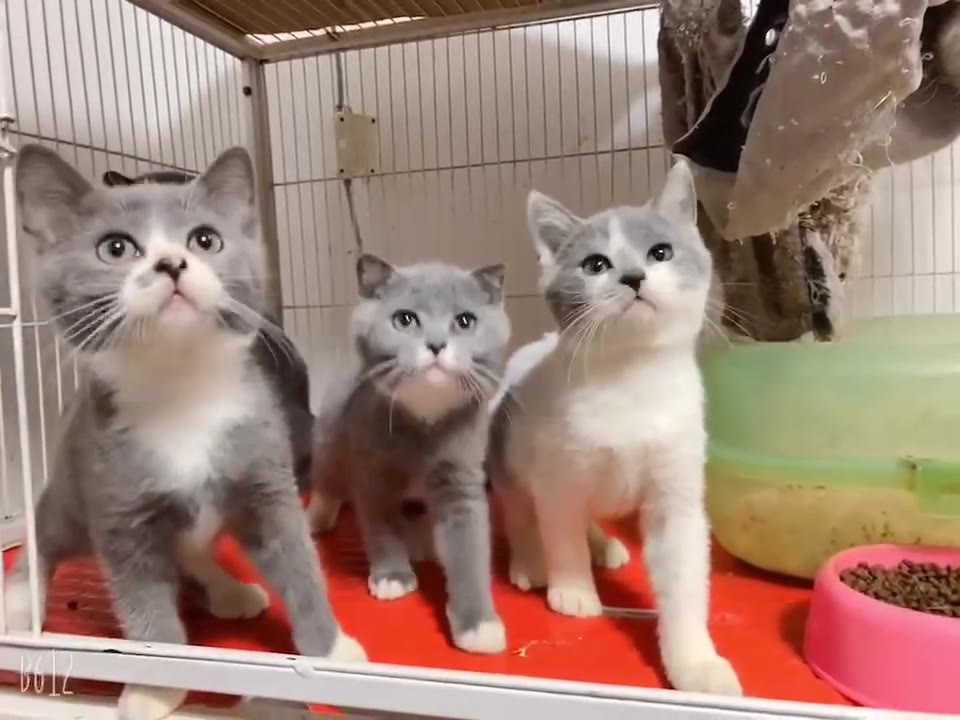} & 
     \includegraphics[width=0.19\linewidth]{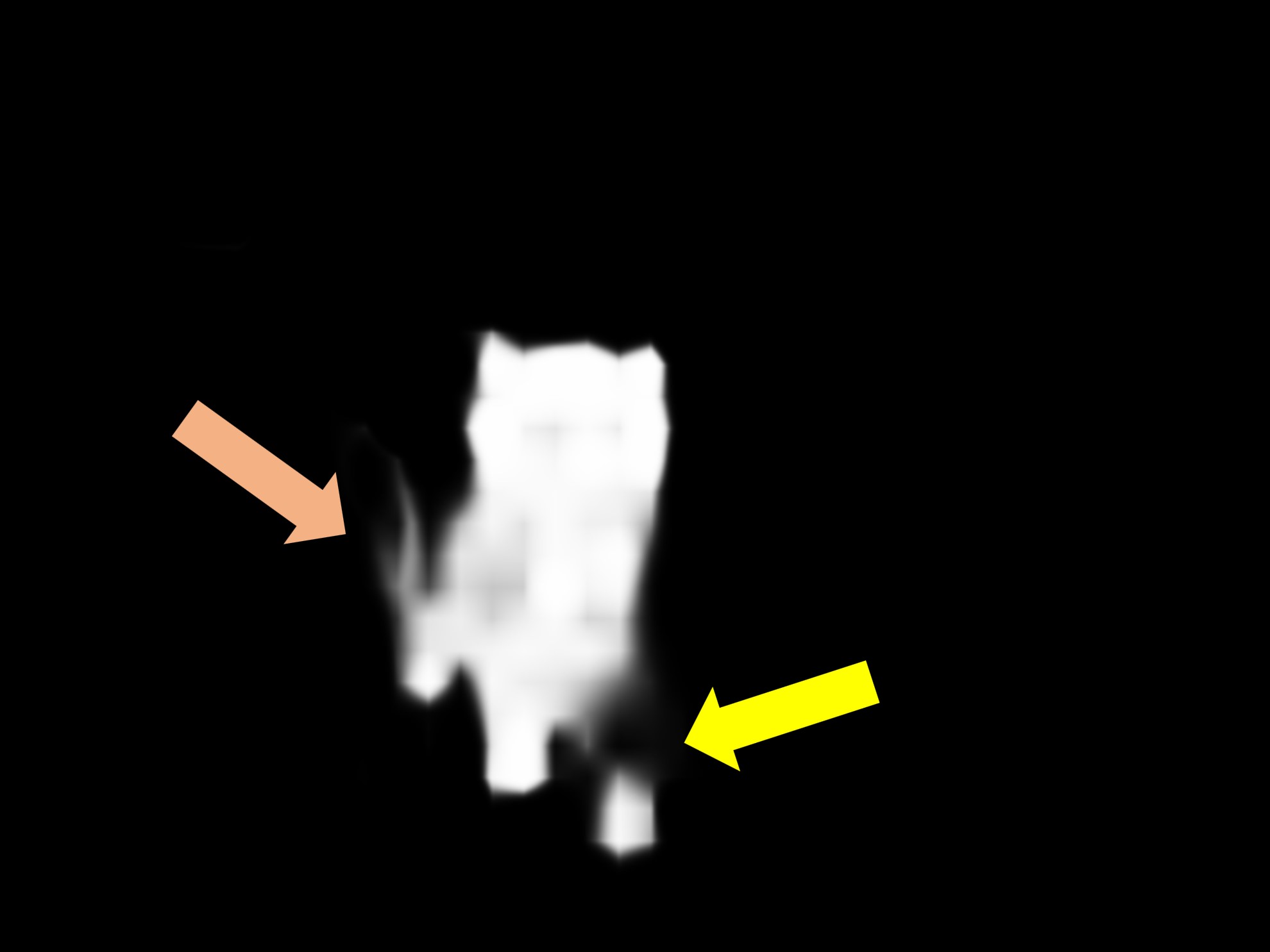} & 
     \includegraphics[width=0.19\linewidth]{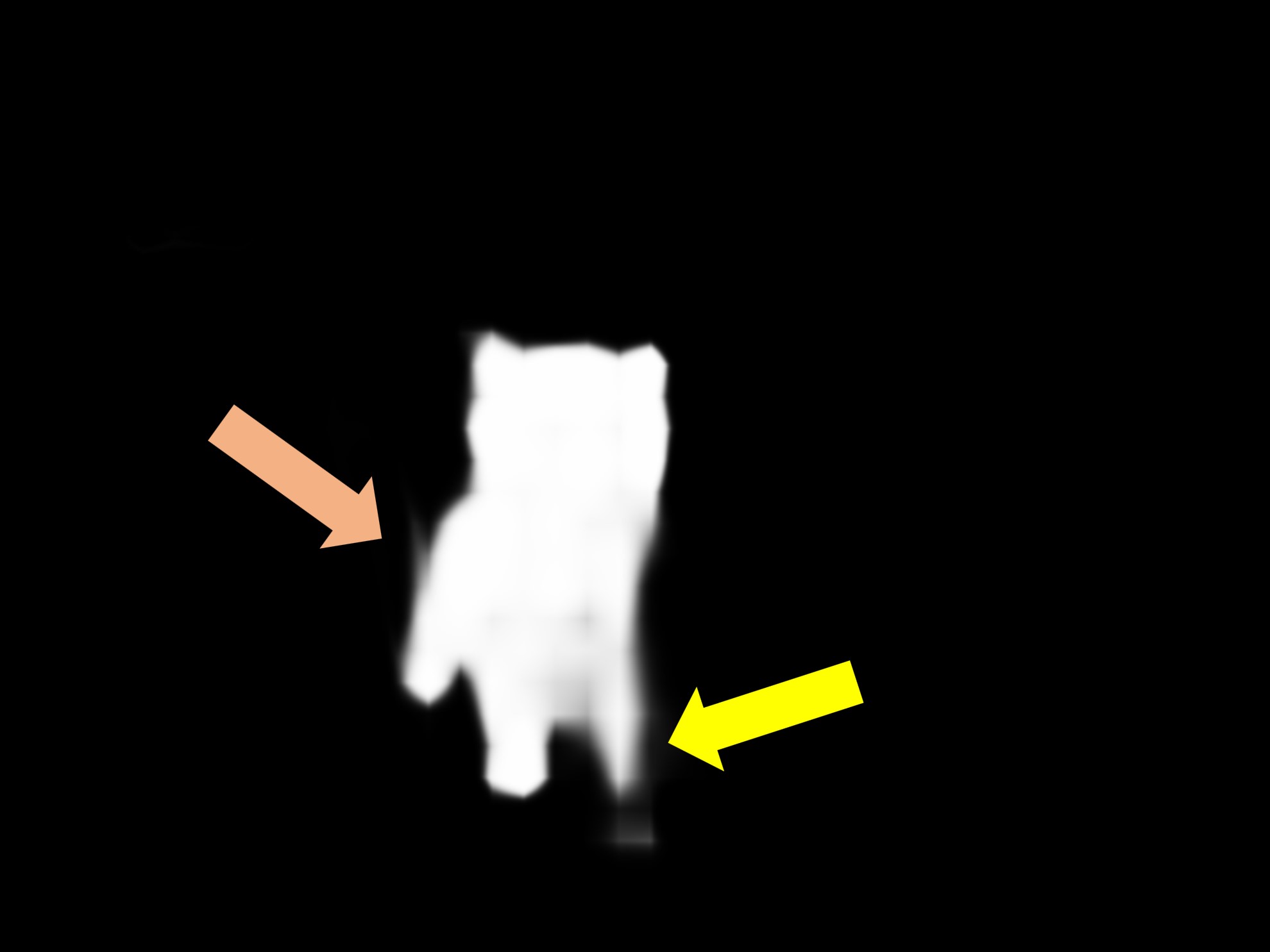} & 
     \includegraphics[width=0.19\linewidth]{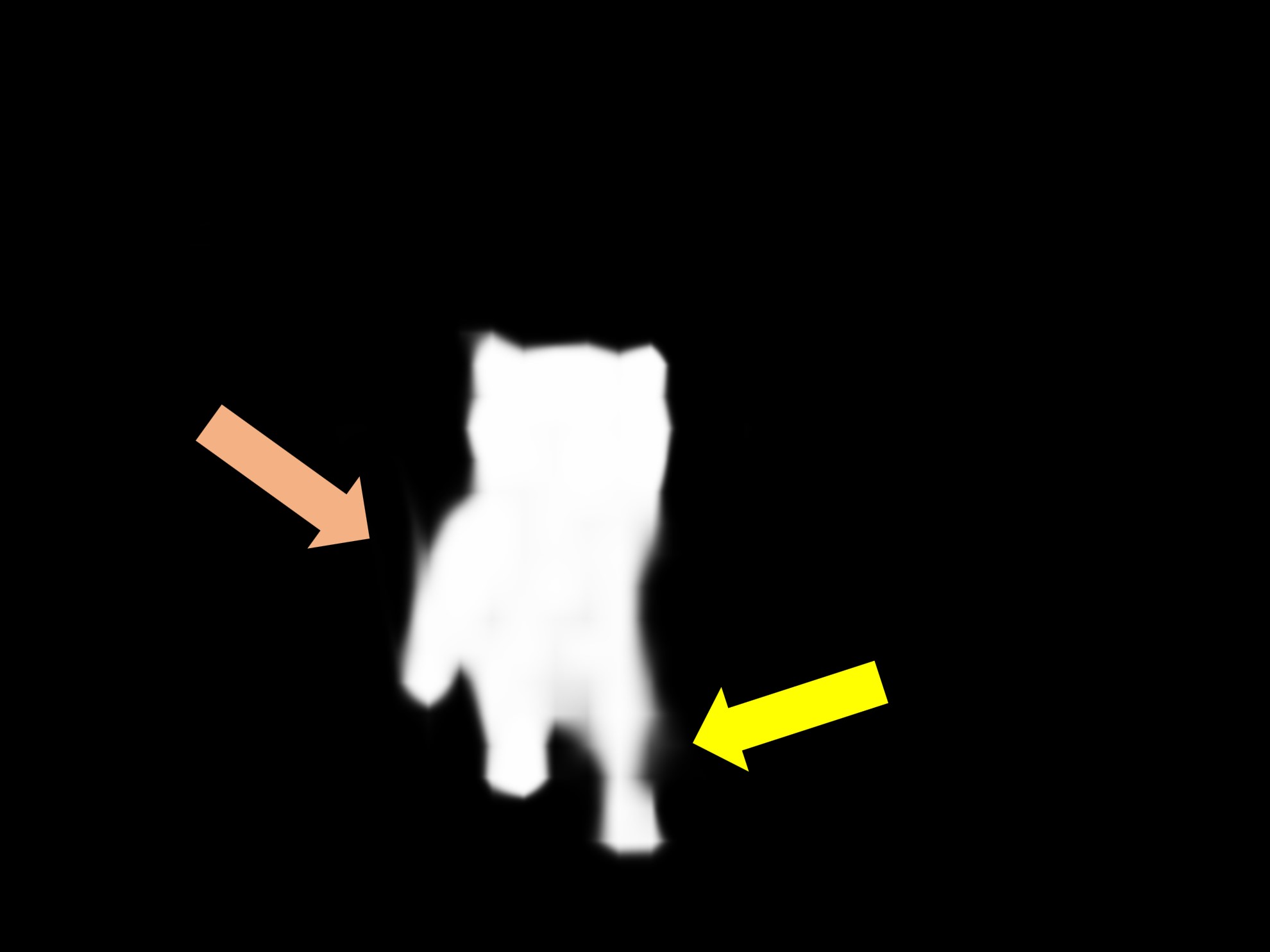} & 
     \includegraphics[width=0.19\linewidth]{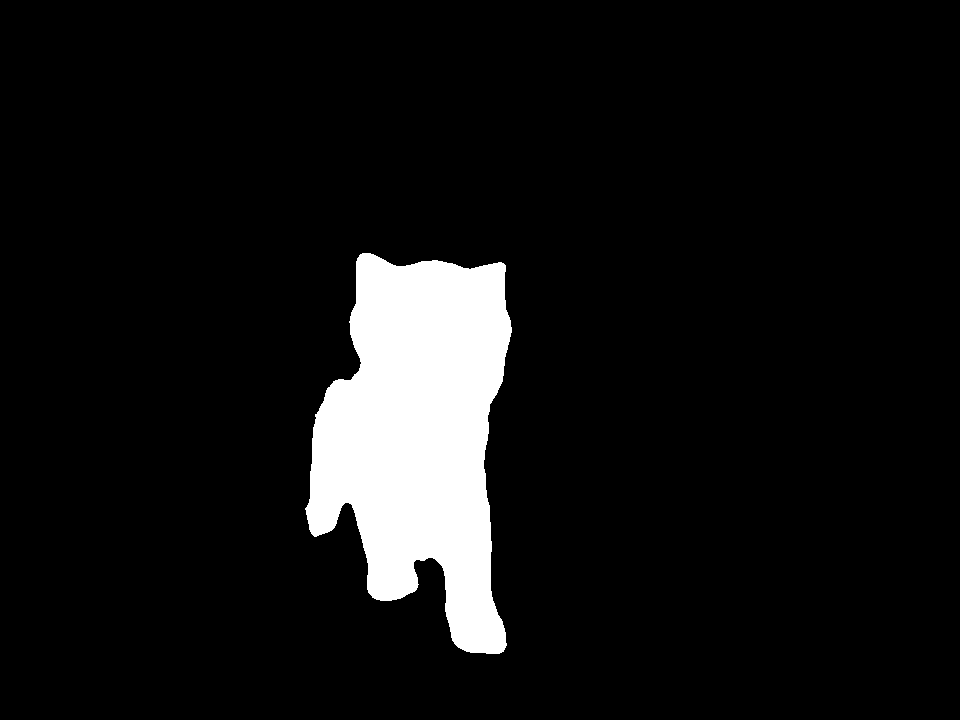} \\
    Image & 
$M_1$ &
$M_2$ & 
$M_3$ & 
Final mask \\
\end{tabular}

\captionof{figure}{Visualization of auxiliary masks ($M_l$) at different layers of the object transformer. At every layer, noises are suppressed (pink arrows) and the target object becomes more coherent (yellow arrows).}
\label{fig:vis-query-blocks}
\end{minipage}
\end{table*}

\myparagraph{Hyperparameter Choices.}
Table~\ref{tab:hyperparameter} compares our results with different choices of hyperparameters: number of object transformer blocks $L$, number of object queries $N$, interval between memory frames $r$, and maximum number of memory frames $T_{\max}$. 
Note that $L=0$ is equivalent to not having an object transformer.
We visualize the progression of pixel features in Figure~\ref{fig:vis-query-blocks}. We find that the object transformer blocks effectively suppress noises from distractors and produce more coherent object masks. 
Cutie is insensitive to the number of object queries -- we think this is because 8 queries are sufficient to model the foreground/background of a single target object. 
As these queries execute in parallel, we find no noticeable differences in running time.
Cutie benefits from having a shorter memory interval and a larger memory bank at the cost of a slower running time (e.g., +2.2~\mjf~on MOSE with half the speed) -- we explore this speed-accuracy trade-off (as Cutie+) without re-training in the supplement.

\myparagraph{Bottom-Up v.s.\ Top-Down Feature.}
Table~\ref{tab:bottom-up-top-down} reports our findings. 
We compare a bottom-up-only approach (similar to XMem~\cite{cheng2022xmem} with the training tricks~\cite{cheng2023tracking} and a lighter backbone) without the object transformer, a top-down-only approach without the pixel memory, and our approach with both. 
Ours, integrating both features, performs the best.

\myparagraph{Masked Attention}
Table~\ref{tab:object-block-config} shows our results with different masked attention configurations.
Masked attention is crucial for good performance -- we hypothesize that using full attention produces confusing signals (especially in cluttered settings, see supplement), which leads to poor generalization.
We note that using full attention also leads to rather unstable training.
We experimented with different distributions of f.g./b.g.\ queries with no significant observed effects.

\myparagraph{Object Memory and Positional Embeddings.}
Table~\ref{tab:object-block-config-2} and Table~\ref{tab:positional-embeddings} ablate on the object memory ($X$), the object query ($S$), and the positional embeddings in the object transformer.
We note that the object query, while standard, is not useful for Cutie in the presence of the object memory.
Positional embeddings are commonly used and do help.

\subsection{Limitations}
Despite being more robust, Cutie often fails when highly similar objects move in close proximity or occlude each other. This problem is not unique to Cutie. 
We suspect that, in these cases, neither the pixel memory nor the object memory is able to pick up sufficiently discriminative features for the object transformer to operate on.
We provide visualizations in the supplementary material.

\section{Conclusion}
We present Cutie, an end-to-end network with object-level memory reading for robust video object segmentation in challenging scenarios.
Cutie efficiently integrates top-down and bottom-up features, achieving new state-of-the-art results in several benchmarks, especially on the challenging MOSE dataset.
We hope to draw more attention to object-centric video segmentation and to enable more accessible universal video segmentation methods via integration with segment-anything models~\cite{kirillov2023segment,cheng2023tracking}.

{
\small
\noindent\textbf{Acknowledgments}. Work supported in part by NSF grants 2008387, 2045586, 2106825, MRI 1725729 (HAL~\cite{kindratenko2020hal}), and NIFA award 2020-67021-32799.
}

{\small
\bibliographystyle{unsrt}
\bibliography{ref}

\begin{thebibliography}{10}

\bibitem{petrik2022learning}
Vladim{\'\i}r Petr{\'\i}k, Mohammad~Nomaan Qureshi, Josef Sivic, and Makar
  Tapaswi.
\newblock Learning object manipulation skills from video via approximate
  differentiable physics.
\newblock In {\em IROS}, 2022.

\bibitem{cheng2021mivos}
Ho~Kei Cheng, Yu-Wing Tai, and Chi-Keung Tang.
\newblock Modular interactive video object segmentation: Interaction-to-mask,
  propagation and difference-aware fusion.
\newblock In {\em CVPR}, 2021.

\bibitem{athar2023burst}
Ali Athar, Jonathon Luiten, Paul Voigtlaender, Tarasha Khurana, Achal Dave,
  Bastian Leibe, and Deva Ramanan.
\newblock Burst: A benchmark for unifying object recognition, segmentation and
  tracking in video.
\newblock In {\em WACV}, 2023.

\bibitem{kirillov2023segment}
Alexander Kirillov, Eric Mintun, Nikhila Ravi, Hanzi Mao, Chloe Rolland, Laura
  Gustafson, Tete Xiao, Spencer Whitehead, Alexander~C Berg, Wan-Yen Lo, et~al.
\newblock Segment anything.
\newblock In {\em arXiv}, 2023.

\bibitem{cheng2023tracking}
Ho~Kei Cheng, Seoung~Wug Oh, Brian Price, Alexander Schwing, and Joon-Young
  Lee.
\newblock Tracking anything with decoupled video segmentation.
\newblock In {\em ICCV}, 2023.

\bibitem{yang2023track}
Jinyu Yang, Mingqi Gao, Zhe Li, Shang Gao, Fangjing Wang, and Feng Zheng.
\newblock Track anything: Segment anything meets videos.
\newblock In {\em arXiv}, 2023.

\bibitem{cheng2023segment}
Yangming Cheng, Liulei Li, Yuanyou Xu, Xiaodi Li, Zongxin Yang, Wenguan Wang,
  and Yi~Yang.
\newblock Segment and track anything.
\newblock In {\em arXiv}, 2023.

\bibitem{oh2019videoSTM}
Seoung~Wug Oh, Joon-Young Lee, Ning Xu, and Seon~Joo Kim.
\newblock Video object segmentation using space-time memory networks.
\newblock In {\em ICCV}, 2019.

\bibitem{cheng2022xmem}
Ho~Kei Cheng and Alexander~G Schwing.
\newblock {XMem}: Long-term video object segmentation with an atkinson-shiffrin
  memory model.
\newblock In {\em ECCV}, 2022.

\bibitem{yang2021associating}
Zongxin Yang, Yunchao Wei, and Yi~Yang.
\newblock Associating objects with transformers for video object segmentation.
\newblock In {\em NeurIPS}, 2021.

\bibitem{bekuzarov2023xmem++}
Maksym Bekuzarov, Ariana Bermudez, Joon-Young Lee, and Hao Li.
\newblock Xmem++: Production-level video segmentation from few annotated
  frames.
\newblock In {\em ICCV}, 2023.

\bibitem{ding2023mose}
Henghui Ding, Chang Liu, Shuting He, Xudong Jiang, Philip~HS Torr, and Song
  Bai.
\newblock {MOSE}: A new dataset for video object segmentation in complex
  scenes.
\newblock In {\em arXiv}, 2023.

\bibitem{perazzi2016benchmark}
Federico Perazzi, Jordi Pont-Tuset, Brian McWilliams, Luc Van~Gool, Markus
  Gross, and Alexander Sorkine-Hornung.
\newblock A benchmark dataset and evaluation methodology for video object
  segmentation.
\newblock In {\em CVPR}, 2016.

\bibitem{carion2020end}
Nicolas Carion, Francisco Massa, Gabriel Synnaeve, Nicolas Usunier, Alexander
  Kirillov, and Sergey Zagoruyko.
\newblock End-to-end object detection with transformers.
\newblock In {\em ECCV}, 2020.

\bibitem{cheng2022masked}
Bowen Cheng, Ishan Misra, Alexander~G Schwing, Alexander Kirillov, and Rohit
  Girdhar.
\newblock Masked-attention mask transformer for universal image segmentation.
\newblock In {\em CVPR}, 2022.

\bibitem{wu2022defense}
Junfeng Wu, Qihao Liu, Yi~Jiang, Song Bai, Alan Yuille, and Xiang Bai.
\newblock In defense of online models for video instance segmentation.
\newblock In {\em ECCV}, 2022.

\bibitem{athar2023tarvis}
Ali Athar, Alexander Hermans, Jonathon Luiten, Deva Ramanan, and Bastian Leibe.
\newblock Tarvis: A unified approach for target-based video segmentation.
\newblock {\em arXiv preprint arXiv:2301.02657}, 2023.

\bibitem{athar2022hodor}
Ali Athar, Jonathon Luiten, Alexander Hermans, Deva Ramanan, and Bastian Leibe.
\newblock Hodor: High-level object descriptors for object re-segmentation in
  video learned from static images.
\newblock In {\em CVPR}, 2022.

\bibitem{xu2018youtubeVOS}
Ning Xu, Linjie Yang, Yuchen Fan, Dingcheng Yue, Yuchen Liang, Jianchao Yang,
  and Thomas Huang.
\newblock Youtube-vos: A large-scale video object segmentation benchmark.
\newblock In {\em ECCV}, 2018.

\bibitem{caelles2017one}
Sergi Caelles, Kevis-Kokitsi Maninis, Jordi Pont-Tuset, Laura Leal-Taix{\'e},
  Daniel Cremers, and Luc Van~Gool.
\newblock One-shot video object segmentation.
\newblock In {\em CVPR}, 2017.

\bibitem{voigtlaender2017online}
Paul Voigtlaender and Bastian Leibe.
\newblock Online adaptation of convolutional neural networks for video object
  segmentation.
\newblock In {\em BMVC}, 2017.

\bibitem{maninis2018video}
K-K Maninis, Sergi Caelles, Yuhua Chen, Jordi Pont-Tuset, Laura Leal-Taix{\'e},
  Daniel Cremers, and Luc Van~Gool.
\newblock Video object segmentation without temporal information.
\newblock In {\em PAMI}, 2018.

\bibitem{bhat2020learning}
Goutam Bhat, Felix~J{\"a}remo Lawin, Martin Danelljan, Andreas Robinson,
  Michael Felsberg, Luc Van~Gool, and Radu Timofte.
\newblock Learning what to learn for video object segmentation.
\newblock In {\em ECCV}, 2020.

\bibitem{robinson2020learning}
Andreas Robinson, Felix~Jaremo Lawin, Martin Danelljan, Fahad~Shahbaz Khan, and
  Michael Felsberg.
\newblock Learning fast and robust target models for video object segmentation.
\newblock In {\em CVPR}, 2020.

\bibitem{perazzi2017learning}
Federico Perazzi, Anna Khoreva, Rodrigo Benenson, Bernt Schiele, and Alexander
  Sorkine-Hornung.
\newblock Learning video object segmentation from static images.
\newblock In {\em CVPR}, 2017.

\bibitem{hu2017maskrnn}
Yuan-Ting Hu, Jia-Bin Huang, and Alexander Schwing.
\newblock Maskrnn: Instance level video object segmentation.
\newblock In {\em NIPS}, 2017.

\bibitem{hu2018motion}
Ping Hu, Gang Wang, Xiangfei Kong, Jason Kuen, and Yap-Peng Tan.
\newblock Motion-guided cascaded refinement network for video object
  segmentation.
\newblock In {\em CVPR}, 2018.

\bibitem{oh2018fast}
Seoung~Wug Oh, Joon-Young Lee, Kalyan Sunkavalli, and Seon Joo~Kim.
\newblock Fast video object segmentation by reference-guided mask propagation.
\newblock In {\em CVPR}, 2018.

\bibitem{wang2019fast}
Qiang Wang, Li~Zhang, Luca Bertinetto, Weiming Hu, and Philip~HS Torr.
\newblock Fast online object tracking and segmentation: A unifying approach.
\newblock In {\em CVPR}, 2019.

\bibitem{zhang2019fast}
Lu~Zhang, Zhe Lin, Jianming Zhang, Huchuan Lu, and You He.
\newblock Fast video object segmentation via dynamic targeting network.
\newblock In {\em ICCV}, 2019.

\bibitem{ventura2019rvos}
Carles Ventura, Miriam Bellver, Andreu Girbau, Amaia Salvador, Ferran Marques,
  and Xavier Giro-i Nieto.
\newblock Rvos: End-to-end recurrent network for video object segmentation.
\newblock In {\em CVPR}, 2019.

\bibitem{hu2018videomatch}
Yuan-Ting Hu, Jia-Bin Huang, and Alexander~G Schwing.
\newblock Videomatch: Matching based video object segmentation.
\newblock In {\em ECCV}, 2018.

\bibitem{voigtlaender2019feelvos}
Paul Voigtlaender, Yuning Chai, Florian Schroff, Hartwig Adam, Bastian Leibe,
  and Liang-Chieh Chen.
\newblock Feelvos: Fast end-to-end embedding learning for video object
  segmentation.
\newblock In {\em CVPR}, 2019.

\bibitem{wang2019ranet}
Ziqin Wang, Jun Xu, Li~Liu, Fan Zhu, and Ling Shao.
\newblock Ranet: Ranking attention network for fast video object segmentation.
\newblock In {\em ICCV}, 2019.

\bibitem{Duarte2019Capsule}
Kevin Duarte, Yogesh~S. Rawat, and Mubarak Shah.
\newblock Capsulevos: Semi-supervised video object segmentation using capsule
  routing.
\newblock In {\em ICCV}, 2019.

\bibitem{yang2020collaborativeCFBI}
Zongxin Yang, Yunchao Wei, and Yi~Yang.
\newblock Collaborative video object segmentation by foreground-background
  integration.
\newblock In {\em ECCV}, 2020.

\bibitem{li2020fastGlobalContext}
Yu~Li, Zhuoran Shen, and Ying Shan.
\newblock Fast video object segmentation using the global context module.
\newblock In {\em ECCV}, 2020.

\bibitem{zhang2020transductive}
Yizhuo Zhang, Zhirong Wu, Houwen Peng, and Stephen Lin.
\newblock A transductive approach for video object segmentation.
\newblock In {\em CVPR}, 2020.

\bibitem{seong2020kernelizedMemory}
Hongje Seong, Junhyuk Hyun, and Euntai Kim.
\newblock Kernelized memory network for video object segmentation.
\newblock In {\em ECCV}, 2020.

\bibitem{lu2020videoGraphMem}
Xiankai Lu, Wenguan Wang, Danelljan Martin, Tianfei Zhou, Jianbing Shen, and
  Van~Gool Luc.
\newblock Video object segmentation with episodic graph memory networks.
\newblock In {\em ECCV}, 2020.

\bibitem{Liang2020AFBURR}
Yongqing Liang, Xin Li, Navid Jafari, and Jim Chen.
\newblock Video object segmentation with adaptive feature bank and
  uncertain-region refinement.
\newblock In {\em NeurIPS}, 2020.

\bibitem{huang2020fast}
Xuhua Huang, Jiarui Xu, Yu-Wing Tai, and Chi-Keung Tang.
\newblock Fast video object segmentation with temporal aggregation network and
  dynamic template matching.
\newblock In {\em CVPR}, 2020.

\bibitem{liang2021video}
Shuxian Liang, Xu~Shen, Jianqiang Huang, and Xian-Sheng Hua.
\newblock Video object segmentation with dynamic memory networks and adaptive
  object alignment.
\newblock In {\em ICCV}, 2021.

\bibitem{xu2021reliable}
Xiaohao Xu, Jinglu Wang, Xiao Li, and Yan Lu.
\newblock Reliable propagation-correction modulation for video object
  segmentation.
\newblock In {\em AAAI}, 2022.

\bibitem{ge2021video}
Wenbin Ge, Xiankai Lu, and Jianbing Shen.
\newblock Video object segmentation using global and instance embedding
  learning.
\newblock In {\em CVPR}, 2021.

\bibitem{hu2021learning}
Li~Hu, Peng Zhang, Bang Zhang, Pan Pan, Yinghui Xu, and Rong Jin.
\newblock Learning position and target consistency for memory-based video
  object segmentation.
\newblock In {\em CVPR}, 2021.

\bibitem{wang2021swiftnet}
Haochen Wang, Xiaolong Jiang, Haibing Ren, Yao Hu, and Song Bai.
\newblock Swiftnet: Real-time video object segmentation.
\newblock In {\em CVPR}, 2021.

\bibitem{xie2021efficient}
Haozhe Xie, Hongxun Yao, Shangchen Zhou, Shengping Zhang, and Wenxiu Sun.
\newblock Efficient regional memory network for video object segmentation.
\newblock In {\em CVPR}, 2021.

\bibitem{seong2021hierarchical}
Hongje Seong, Seoung~Wug Oh, Joon-Young Lee, Seongwon Lee, Suhyeon Lee, and
  Euntai Kim.
\newblock Hierarchical memory matching network for video object segmentation.
\newblock In {\em ICCV}, 2021.

\bibitem{mao2021joint}
Yunyao Mao, Ning Wang, Wengang Zhou, and Houqiang Li.
\newblock Joint inductive and transductive learning for video object
  segmentation.
\newblock In {\em ICCV}, 2021.

\bibitem{cheng2021stcn}
Ho~Kei Cheng, Yu-Wing Tai, and Chi-Keung Tang.
\newblock Rethinking space-time networks with improved memory coverage for
  efficient video object segmentation.
\newblock In {\em NeurIPS}, 2021.

\bibitem{liu2022learning}
Yong Liu, Ran Yu, Fei Yin, Xinyuan Zhao, Wei Zhao, Weihao Xia, and Yujiu Yang.
\newblock Learning quality-aware dynamic memory for video object segmentation.
\newblock In {\em ECCV}, 2022.

\bibitem{yu2022batman}
Ye~Yu, Jialin Yuan, Gaurav Mittal, Li~Fuxin, and Mei Chen.
\newblock Batman: Bilateral attention transformer in motion-appearance
  neighboring space for video object segmentation.
\newblock In {\em ECCV}, 2022.

\bibitem{miao2022region}
Bo~Miao, Mohammed Bennamoun, Yongsheng Gao, and Ajmal Mian.
\newblock Region aware video object segmentation with deep motion modeling.
\newblock In {\em arXiv}, 2022.

\bibitem{li2022recurrent}
Mingxing Li, Li~Hu, Zhiwei Xiong, Bang Zhang, Pan Pan, and Dong Liu.
\newblock Recurrent dynamic embedding for video object segmentation.
\newblock In {\em CVPR}, 2022.

\bibitem{park2022per}
Kwanyong Park, Sanghyun Woo, Seoung~Wug Oh, In~So Kweon, and Joon-Young Lee.
\newblock Per-clip video object segmentation.
\newblock In {\em CVPR}, 2022.

\bibitem{liu2022global}
Yong Liu, Ran Yu, Jiahao Wang, Xinyuan Zhao, Yitong Wang, Yansong Tang, and
  Yujiu Yang.
\newblock Global spectral filter memory network for video object segmentation.
\newblock In {\em ECCV}, 2022.

\bibitem{zhang2023boosting}
Yurong Zhang, Liulei Li, Wenguan Wang, Rong Xie, Li~Song, and Wenjun Zhang.
\newblock Boosting video object segmentation via space-time correspondence
  learning.
\newblock In {\em CVPR}, 2023.

\bibitem{xu2022towards}
Xiaohao Xu, Jinglu Wang, Xiang Ming, and Yan Lu.
\newblock Towards robust video object segmentation with adaptive object
  calibration.
\newblock In {\em ACM MM}, 2022.

\bibitem{cho2022pixel}
Suhwan Cho, Heansung Lee, Minjung Kim, Sungjun Jang, and Sangyoun Lee.
\newblock Pixel-level bijective matching for video object segmentation.
\newblock In {\em WACV}, 2022.

\bibitem{xu2022accelerating}
Kai Xu and Angela Yao.
\newblock Accelerating video object segmentation with compressed video.
\newblock In {\em CVPR}, 2022.

\bibitem{miles2023mobilevos}
Roy Miles, Mehmet~Kerim Yucel, Bruno Manganelli, and Albert Saa-Garriga.
\newblock Mobilevos: Real-time video object segmentation contrastive learning
  meets knowledge distillation.
\newblock In {\em CVPR}, 2023.

\bibitem{yan2023two}
Kun Yan, Xiao Li, Fangyun Wei, Jinglu Wang, Chenbin Zhang, Ping Wang, and Yan
  Lu.
\newblock Two-shot video object segmentation.
\newblock In {\em CVPR}, 2023.

\bibitem{sun2023alignment}
Rui Sun, Yuan Wang, Huayu Mai, Tianzhu Zhang, and Feng Wu.
\newblock Alignment before aggregation: trajectory memory retrieval network for
  video object segmentation.
\newblock In {\em ICCV}, 2023.

\bibitem{yang2021collaborativeplus}
Zongxin Yang, Yunchao Wei, and Yi~Yang.
\newblock Collaborative video object segmentation by multi-scale
  foreground-background integration.
\newblock In {\em TPAMI}, 2021.

\bibitem{vaswani2017attention}
Ashish Vaswani, Noam Shazeer, Niki Parmar, Jakob Uszkoreit, Llion Jones,
  Aidan~N Gomez, {\L}ukasz Kaiser, and Illia Polosukhin.
\newblock Attention is all you need.
\newblock In {\em NeurIPS}, 2017.

\bibitem{duke2021sstvos}
Brendan Duke, Abdalla Ahmed, Christian Wolf, Parham Aarabi, and Graham~W
  Taylor.
\newblock Sstvos: Sparse spatiotemporal transformers for video object
  segmentation.
\newblock In {\em CVPR}, 2021.

\bibitem{yang2022decoupling}
Zongxin Yang and Yi~Yang.
\newblock Decoupling features in hierarchical propagation for video object
  segmentation.
\newblock In {\em NeurIPS}, 2022.

\bibitem{wu2023scalable}
Qiangqiang Wu, Tianyu Yang, Wei Wu, and Antoni Chan.
\newblock Scalable video object segmentation with simplified framework.
\newblock In {\em ICCV}, 2023.

\bibitem{zhang2023joint}
Jiaming Zhang, Yutao Cui, Gangshan Wu, and Limin Wang.
\newblock Joint modeling of feature, correspondence, and a compressed memory
  for video object segmentation.
\newblock In {\em arXiv}, 2023.

\bibitem{he2021masked}
Kaiming He, Xinlei Chen, Saining Xie, Yanghao Li, Piotr Doll’ar, and Ross~B
  Girshick.
\newblock Masked autoencoders are scalable vision learners. 2022 ieee.
\newblock In {\em CVPR}, 2021.

\bibitem{li2018video}
Xiaoxiao Li and Chen~Change Loy.
\newblock Video object segmentation with joint re-identification and
  attention-aware mask propagation.
\newblock In {\em ECCV}, 2018.

\bibitem{luiten2018premvos}
Jonathon Luiten, Paul Voigtlaender, and Bastian Leibe.
\newblock Premvos: Proposal-generation, refinement and merging for video object
  segmentation.
\newblock In {\em ACCV}, 2018.

\bibitem{yan2023universal}
Bin Yan, Yi~Jiang, Jiannan Wu, Dong Wang, Ping Luo, Zehuan Yuan, and Huchuan
  Lu.
\newblock Universal instance perception as object discovery and retrieval.
\newblock In {\em CVPR}, 2023.

\bibitem{yan2022towards}
Bin Yan, Yi~Jiang, Peize Sun, Dong Wang, Zehuan Yuan, Ping Luo, and Huchuan Lu.
\newblock Towards grand unification of object tracking.
\newblock In {\em ECCV}, 2022.

\bibitem{wang2022look}
Junke Wang, Dongdong Chen, Zuxuan Wu, Chong Luo, Chuanxin Tang, Xiyang Dai,
  Yucheng Zhao, Yujia Xie, Lu~Yuan, and Yu-Gang Jiang.
\newblock Look before you match: Instance understanding matters in video object
  segmentation.
\newblock In {\em CVPR}, 2023.

\bibitem{garg2021mask}
Shubhika Garg and Vidit Goel.
\newblock Mask selection and propagation for unsupervised video object
  segmentation.
\newblock In {\em WCAV}, 2021.

\bibitem{xiong2020layer}
Ruibin Xiong, Yunchang Yang, Di~He, Kai Zheng, Shuxin Zheng, Chen Xing,
  Huishuai Zhang, Yanyan Lan, Liwei Wang, and Tieyan Liu.
\newblock On layer normalization in the transformer architecture.
\newblock In {\em ICLR}, 2020.

\bibitem{wang2019eca}
Qilong Wang, Banggu Wu, Pengfei Zhu, Peihua Li, Wangmeng Zuo, and Qinghua Hu.
\newblock Eca-net: efficient channel attention for deep convolutional neural
  networks.
\newblock In {\em CVPR}, 2020.

\bibitem{he2016deepResNet}
Kaiming He, Xiangyu Zhang, Shaoqing Ren, and Jian Sun.
\newblock Deep residual learning for image recognition.
\newblock In {\em CVPR}, 2016.

\bibitem{shi2015hierarchicalECSSD}
Jianping Shi, Qiong Yan, Li~Xu, and Jiaya Jia.
\newblock Hierarchical image saliency detection on extended cssd.
\newblock In {\em TPAMI}, 2015.

\bibitem{wang2017DUTS}
Lijun Wang, Huchuan Lu, Yifan Wang, Mengyang Feng, Dong Wang, Baocai Yin, and
  Xiang Ruan.
\newblock Learning to detect salient objects with image-level supervision.
\newblock In {\em CVPR}, 2017.

\bibitem{li2020fss}
Xiang Li, Tianhan Wei, Yau~Pun Chen, Yu-Wing Tai, and Chi-Keung Tang.
\newblock Fss-1000: A 1000-class dataset for few-shot segmentation.
\newblock In {\em CVPR}, 2020.

\bibitem{zeng2019towardsHRSOD}
Yi~Zeng, Pingping Zhang, Jianming Zhang, Zhe Lin, and Huchuan Lu.
\newblock Towards high-resolution salient object detection.
\newblock In {\em ICCV}, 2019.

\bibitem{cheng2020cascadepsp}
Ho~Kei Cheng, Jihoon Chung, Yu-Wing Tai, and Chi-Keung Tang.
\newblock Cascadepsp: Toward class-agnostic and very high-resolution
  segmentation via global and local refinement.
\newblock In {\em CVPR}, 2020.

\bibitem{qi2022occluded}
Jiyang Qi, Yan Gao, Yao Hu, Xinggang Wang, Xiaoyu Liu, Xiang Bai, Serge
  Belongie, Alan Yuille, Philip~HS Torr, and Song Bai.
\newblock Occluded video instance segmentation: A benchmark.
\newblock In {\em IJCV}, 2022.

\bibitem{loshchilov2017decoupledAdamW}
Ilya Loshchilov and Frank Hutter.
\newblock Decoupled weight decay regularization.
\newblock In {\em ICLR}, 2019.

\bibitem{kirillov2020pointrend}
Alexander Kirillov, Yuxin Wu, Kaiming He, and Ross Girshick.
\newblock Pointrend: Image segmentation as rendering.
\newblock In {\em CVPR}, 2020.

\bibitem{hong2022lvos}
Lingyi Hong, Wenchao Chen, Zhongying Liu, Wei Zhang, Pinxue Guo, Zhaoyu Chen,
  and Wenqiang Zhang.
\newblock Lvos: A benchmark for long-term video object segmentation.
\newblock In {\em ICCV}, 2023.

\bibitem{luiten2021hota}
Jonathon Luiten, Aljosa Osep, Patrick Dendorfer, Philip Torr, Andreas Geiger,
  Laura Leal-Taix{\'e}, and Bastian Leibe.
\newblock Hota: A higher order metric for evaluating multi-object tracking.
\newblock In {\em IJCV}, 2021.

\bibitem{deng2009imagenet}
Jia Deng, Wei Dong, Richard Socher, Li-Jia Li, Kai Li, and Li~Fei-Fei.
\newblock Imagenet: A large-scale hierarchical image database.
\newblock In {\em CVPR}, 2009.

\bibitem{kindratenko2020hal}
Volodymyr Kindratenko, Dawei Mu, Yan Zhan, John Maloney, Sayed~Hadi Hashemi,
  Benjamin Rabe, Ke~Xu, Roy Campbell, Jian Peng, and William Gropp.
\newblock Hal: Computer system for scalable deep learning.
\newblock In {\em PEARC}, 2020.

\bibitem{lin2014microsoft}
Tsung-Yi Lin, Michael Maire, Serge Belongie, James Hays, Pietro Perona, Deva
  Ramanan, Piotr Doll{\'a}r, and C~Lawrence Zitnick.
\newblock Microsoft coco: Common objects in context.
\newblock In {\em ECCV}, 2014.

\bibitem{PyTorch}
Adam Paszke, Sam Gross, Francisco Massa, Adam Lerer, James Bradbury, Gregory
  Chanan, Trevor Killeen, Zeming Lin, Natalia Gimelshein, Luca Antiga, Alban
  Desmaison, Andreas Kopf, Edward Yang, Zachary DeVito, Martin Raison, Alykhan
  Tejani, Sasank Chilamkurthy, Benoit Steiner, Lu~Fang, Junjie Bai, and Soumith
  Chintala.
\newblock Pytorch: An imperative style, high-performance deep learning library.
\newblock In {\em NeurIPS}, 2019.

\bibitem{duchon1977splines}
Jean Duchon.
\newblock Splines minimizing rotation-invariant semi-norms in sobolev spaces.
\newblock In {\em Constructive Theory of Functions of Several Variables}, 1977.

\bibitem{ghiasi2021simple}
Golnaz Ghiasi, Yin Cui, Aravind Srinivas, Rui Qian, Tsung-Yi Lin, Ekin~D Cubuk,
  Quoc~V Le, and Barret Zoph.
\newblock Simple copy-paste is a strong data augmentation method for instance
  segmentation.
\newblock In {\em CVPR}, 2021.

\bibitem{cho2014propertiesGRU}
Kyunghyun Cho, Bart Van~Merri{\"e}nboer, Dzmitry Bahdanau, and Yoshua Bengio.
\newblock On the properties of neural machine translation: Encoder-decoder
  approaches.
\newblock In {\em arXiv}, 2014.

\bibitem{sofiiuk2022reviving}
Konstantin Sofiiuk, Ilya~A Petrov, and Anton Konushin.
\newblock Reviving iterative training with mask guidance for interactive
  segmentation.
\newblock In {\em ICIP}, 2022.

\end{thebibliography}
}

\clearpage
\appendix
\beginsupplement
\onecolumn

\begin{center}
     \Large\textbf{Supplementary Material\\Putting the Object Back into Video Object Segmentation}
\end{center}

\noindent The supplementary material is structured as follows:

\begin{enumerate}
    \item We first provide visual comparisons of Cutie with state-of-the-art methods in Section~\ref{sec:app:visual-comparisons}.
    \item We then show some highly challenging cases where both Cutie and state-of-the-art methods fail in Section~\ref{sec:app:failure-cases}.
    \item We analyze the running time of XMem and Cutie in Section~\ref{sec:app:running-time}.
    \item To elucidate the workings of the object transformer, we visualize the difference in attention patterns of pixel readout v.s.\ object readout, feature progression within the object transformer, and the qualitative benefits of masked attention/object transformer in Section~\ref{sec:app:visualizations}.
    \item We present additional details on BURST evaluation in Section~\ref{sec:app:burst-details}.
    \item We list options for adjusting the speed-accuracy trade-off without re-training, comparisons with methods that use external training, additional quantitative results on YouTube-VOS 2018~\cite{xu2018youtubeVOS}/LVOS~\cite{hong2022lvos}, and the performance variations with respect to different random seeds in Section~\ref{sec:app:quantitative}.
    \item We give more implementation details on the training process, decoder architecture, and pixel memory in Section~\ref{sec:app:implementation}.
    \item Lastly, we showcase an interactive video segmentation tool powered by Cutie in Section~\ref{sec:app:interactive-tool}. This tool will be open-sourced to help researchers and data annotators.
\end{enumerate}

\section{Visual Comparisons}\label{sec:app:visual-comparisons}
We provide visual comparisons of Cutie with DeAOT-R50~\cite{yang2022decoupling} and XMem~\cite{cheng2022xmem} at \href{https://youtu.be/LGbJ11GT8Ig}{youtu.be/LGbJ11GT8Ig}.
For a fair comparison, we use Cutie-base and train all models with the MOSE dataset.
We visualize the comparisons using sequences from YouTubeVOS-2019 validation, DAVIS 2017 test-dev, and MOSE validation. 
Only the first-frame (not full-video) ground-truth annotations are available in these datasets. 
At the beginning of each sequence, we pause for two seconds to show the first-frame segmentation that initializes all the models.
Our model is more robust to distractors and generates more coherent masks.

\section{Failure Cases}\label{sec:app:failure-cases}
We visualize some failure cases of Cutie at \href{https://youtu.be/PIjXUYRzQ8Q}{youtu.be/PIjXUYRzQ8Q}, following the format discussed in Section~\ref{sec:app:visual-comparisons}.
As discussed in the main paper, Cutie fails in some of the challenging cases where similar objects move in close proximity or occlude each other. 
This problem is not unique to Cutie, as current state-of-the-art methods also fail as shown in the video.

In the first sequence ``elephants'', all models have difficulty tracking the elephants (e.g., light blue mask) behind the big (unannotated) foreground elephant.
In the second sequence ``birds'', all models fail when the bird with the pink mask moves and occludes other birds.

We think that this is due to the lack of useful features from the pixel memory and the object memory, as they fail to disambiguate objects that are similar in both appearance and position.
A potential future work direction for this is to encode three-dimensional spatial understanding (i.e., the bird that occludes is closer to the camera).

\section{Running Time Analysis}\label{sec:app:running-time}
We analyze the total runtime of XMem and Cutie in Tab.~\ref{tab:app:running-time}, testing on a single video with a 2080Ti. 
We synchronize and warm up properly to get accurate timing; small deviations might arise from minor implementation differences and run-time variations.
We note that the speedup is mostly achieved by using a lighter decoder.

\begin{table}[h]
    \centering
\begin{tabular}{lccc}
\toprule
& XMem & Cutie-base & Cutie-small \\
\midrule
Query encoder & 0.861 & 0.851 & 0.295 \\
Mask encoder & 0.143 & 0.145 & 0.142 \\
Pixel memory read & 0.758 & 0.514 & 0.514  \\
Object memory read & - & 0.913 & 0.894 \\
Decoding & 2.749 & 0.725 & 0.700 \\
\midrule
\bottomrule
\end{tabular}

    \caption{Total running time (s) of each component in XMem and Cutie.}
    \label{tab:app:running-time}
\end{table}

\section{Additional Visualizations}\label{sec:app:visualizations}
\subsection{Attention Patterns of Pixel Attention v.s.\ Object Query Attention}
Here, we visualize the attention maps of pixel memory reading and of the object transformer, showing the qualitative difference between the two.

To visualize attention in pixel memory reading, we use ``self-attention'', i.e., by setting $\mathbf{k}=\mathbf{q}\in\mathbb{R}^{HW\times C^{\textbf{k}}}$.
We compute the pixel affinity $A^{\text{pix}}\in[0, 1]^{HW\times HW}$, as in Equation (9) of the main paper.
We then sum over the foreground region along the rows, visualizing the affinity of every pixel to the foreground. Ideally, all the affinity should be in the foreground -- others are distractors that cause erroneous matching. The foreground region is defined by the last auxiliary mask $M_L$ in the object transformer.

To visualize the attention in the object transformer, we inspect the attention weights $A_L\in\mathbb{R}^{N\times HW}$ of the first (pixel-to-query) cross-attention in the last object transformer block.
Similar to how we visualize the pixel attention, we focus on the foreground queries (i.e., first $N/2$ object queries) and sum over the corresponding rows in the affinity matrix.

\begin{figure}
    \centering
    \small
\begin{tabular}{c@{\hspace{4pt}}c@{\hspace{4pt}}c@{\hspace{4pt}}c}
    Input image &  Target object mask & Pixel attention map & Object query attention map \\
    \includegraphics[width=0.24\linewidth]{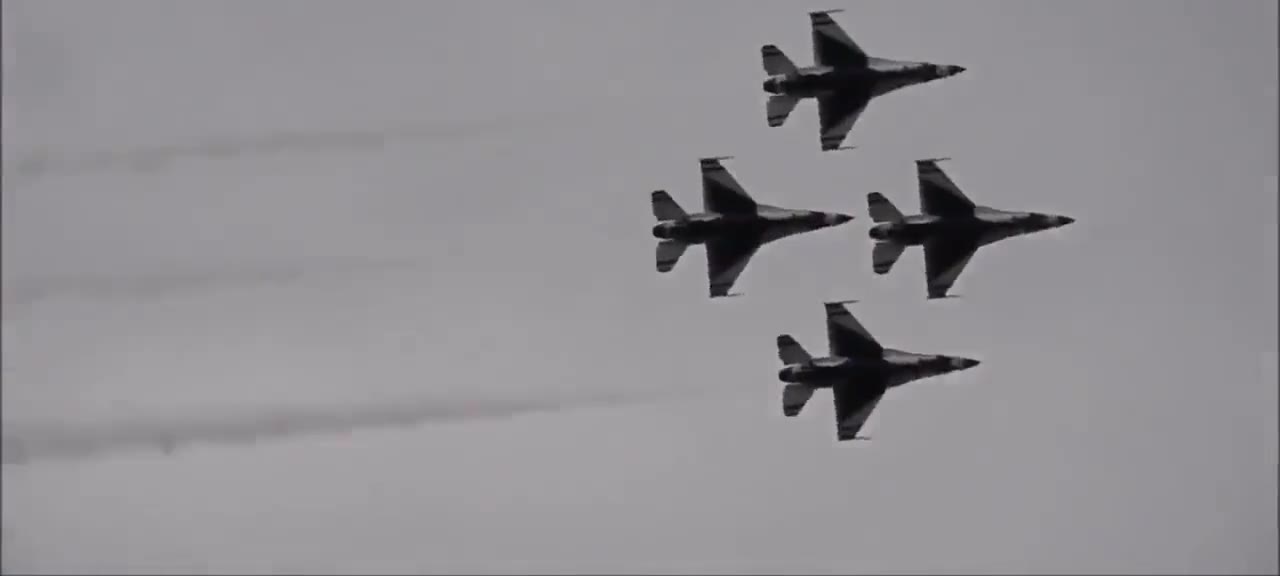} & 
    \includegraphics[width=0.24\linewidth]{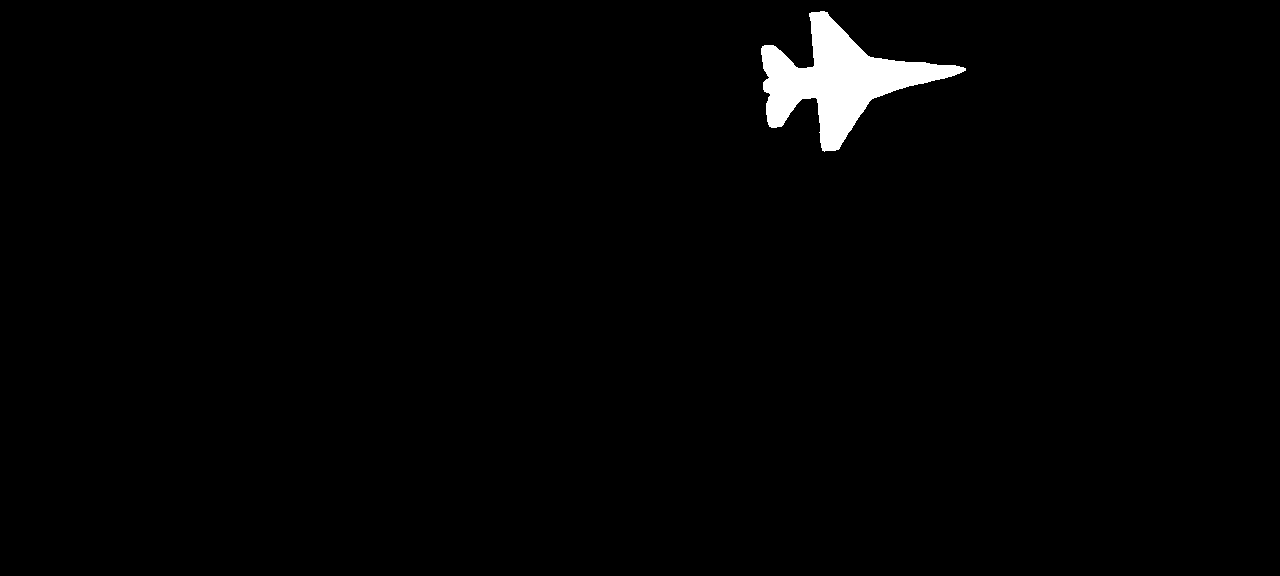} & 
    \includegraphics[width=0.24\linewidth]{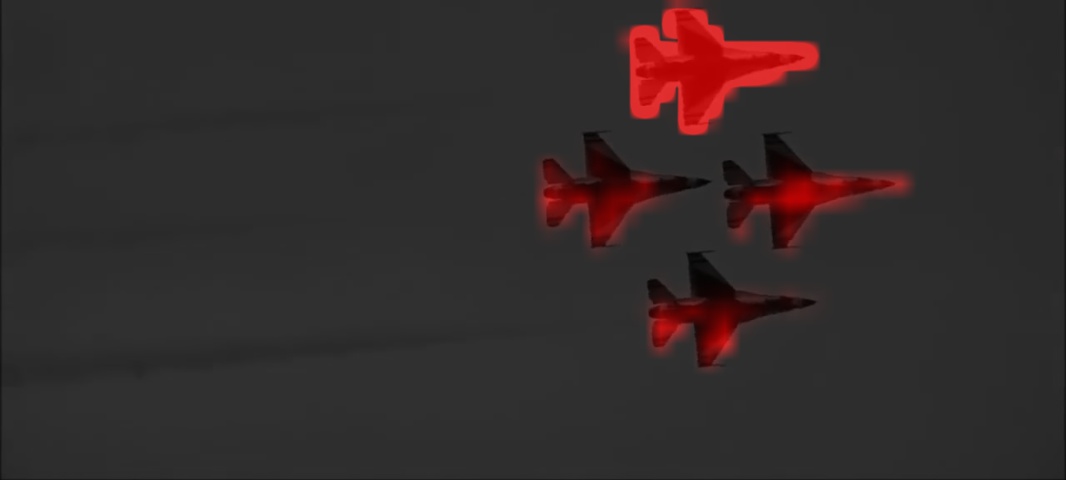} &
    \includegraphics[width=0.24\linewidth]{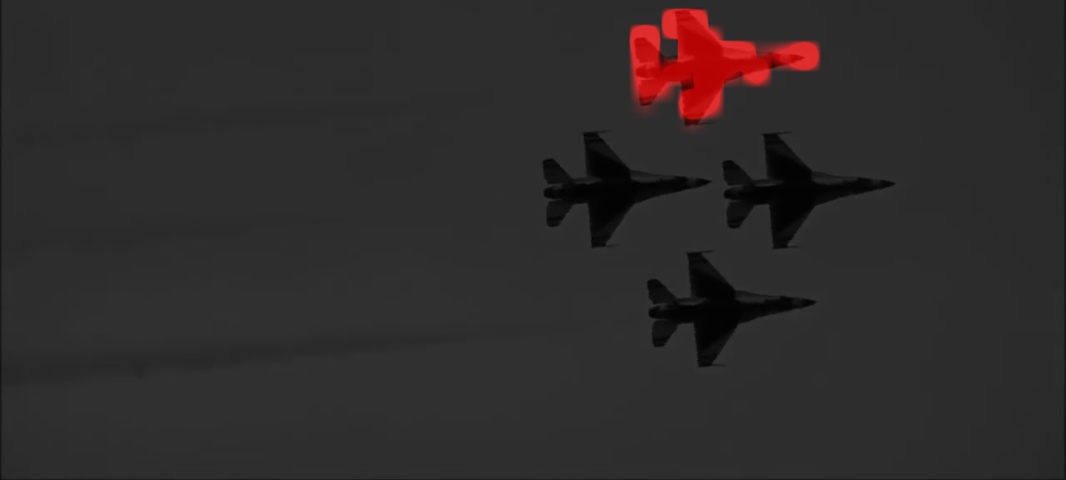} \\ 
    \includegraphics[width=0.24\linewidth]{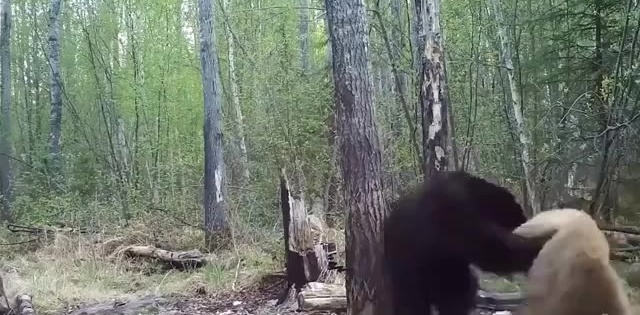} & 
    \includegraphics[width=0.24\linewidth]{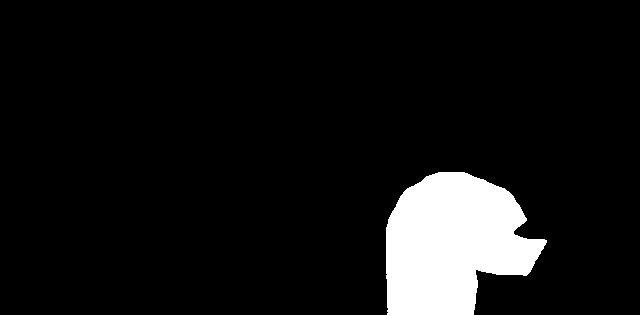} & 
    \includegraphics[width=0.24\linewidth]{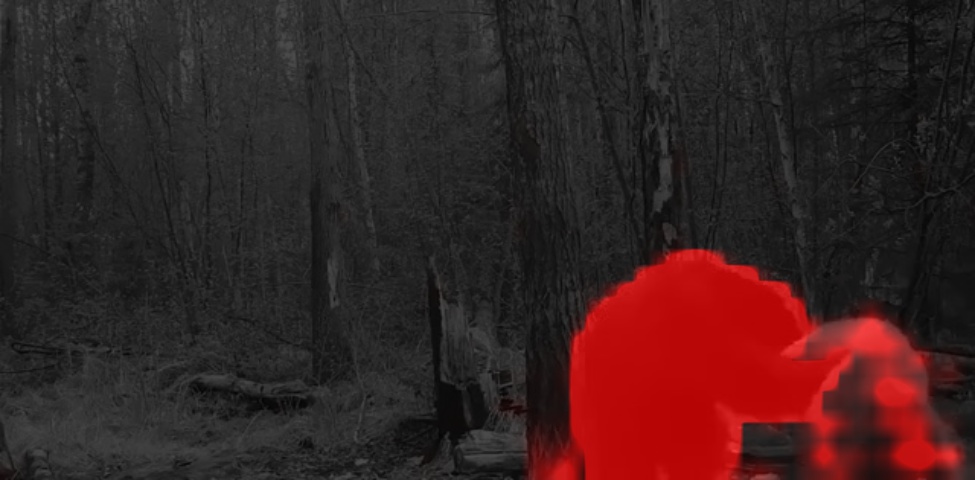} &
    \includegraphics[width=0.24\linewidth]{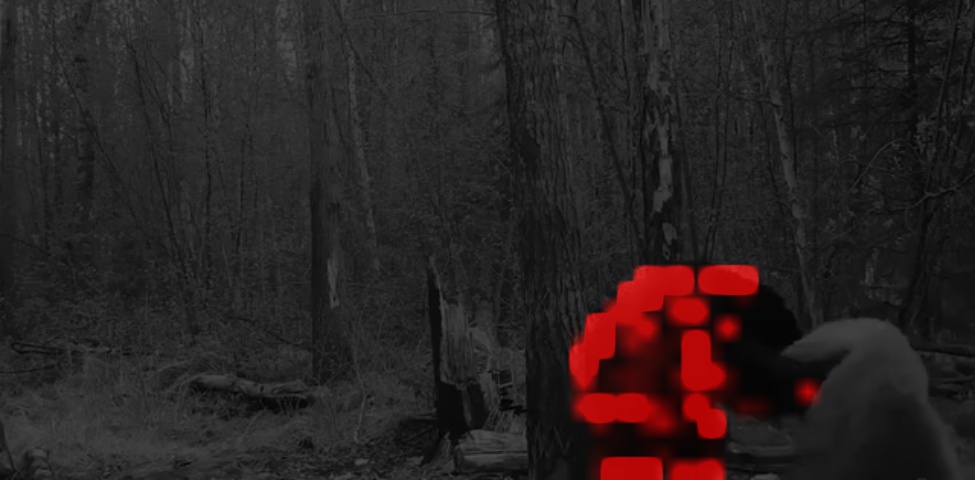} \\
    \includegraphics[width=0.24\linewidth]{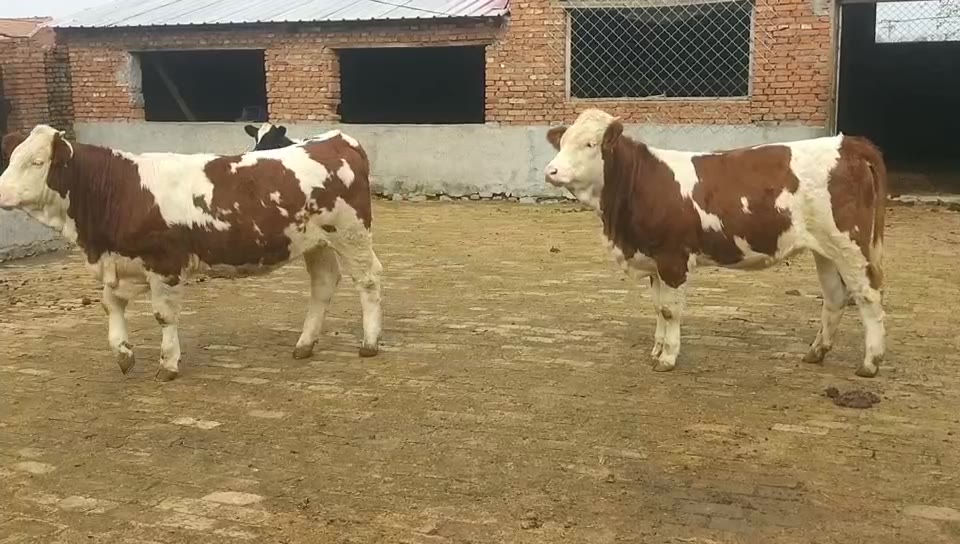} & 
    \includegraphics[width=0.24\linewidth]{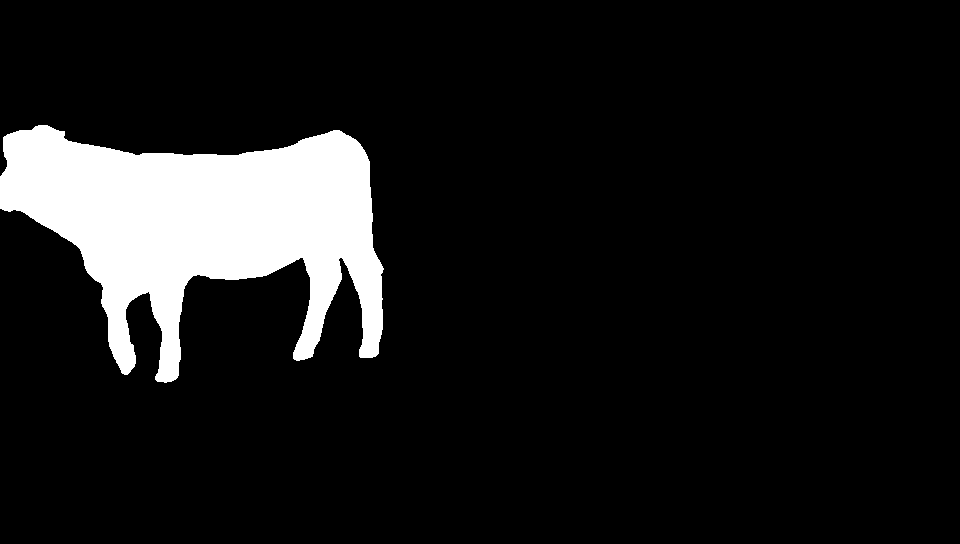} & 
    \includegraphics[width=0.24\linewidth]{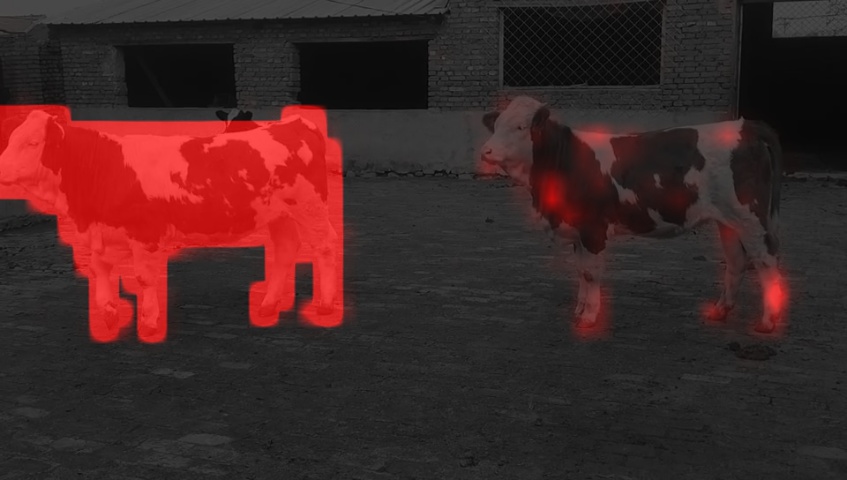} &
    \includegraphics[width=0.24\linewidth]{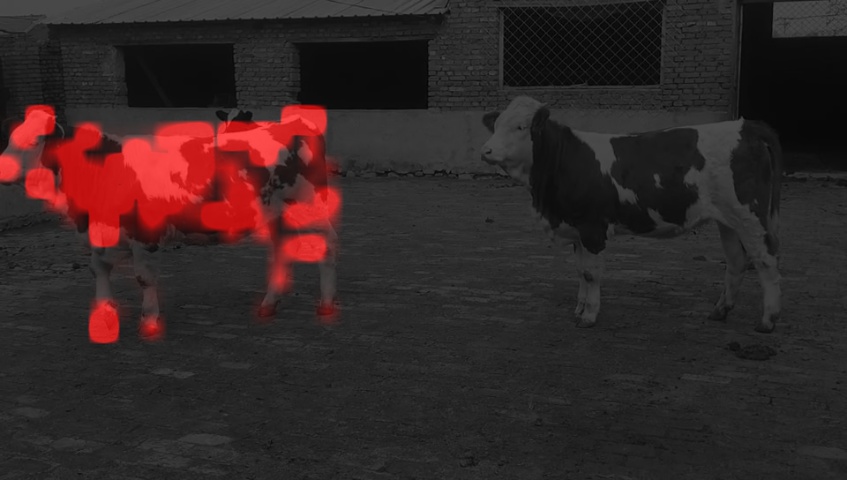} \\
 \end{tabular}

    \caption{Comparison of foreground attention patterns between pixel attention with object query attention. 
    In each of the three examples, the two leftmost columns show the input image and the ground-truth (annotated by us for reference).
    The two rightmost columns show the attention patterns for pixel attention and object query attention respectively.
    Ideally, the attention weights should focus on the foreground object. As shown, the pixel attention has a broader coverage but is easily distracted by similar objects.
    The object query attention's attention is more sparse (as we use a small number of object queries), and is more focused on the foreground. 
    Our object transformer makes use of both: it first initializes with pixel attention and restructures the features iteratively with object query attention.
    }
    \label{fig:app:attn-weights}
\end{figure}

Figure~\ref{fig:app:attn-weights} visualizes the differences between these two types of attention. The pixel attention is more spread out and is easily distracted by similar objects.
The object query attention focuses on the foreground without being distracted.
Our object transformer makes use of both types of attention by using pixel attention for initialization and object query attention for restructuring the feature in every transformer block.

\begin{figure}
    \centering
    \begin{tabular}{c@{\hspace{1pt}}c@{\hspace{1pt}}c@{\hspace{1pt}}c@{\hspace{1pt}}c}
Image & 
$M_1$ &
$M_2$ & 
$M_3$ & 
Ground-truth \\
 \includegraphics[width=0.19\linewidth]{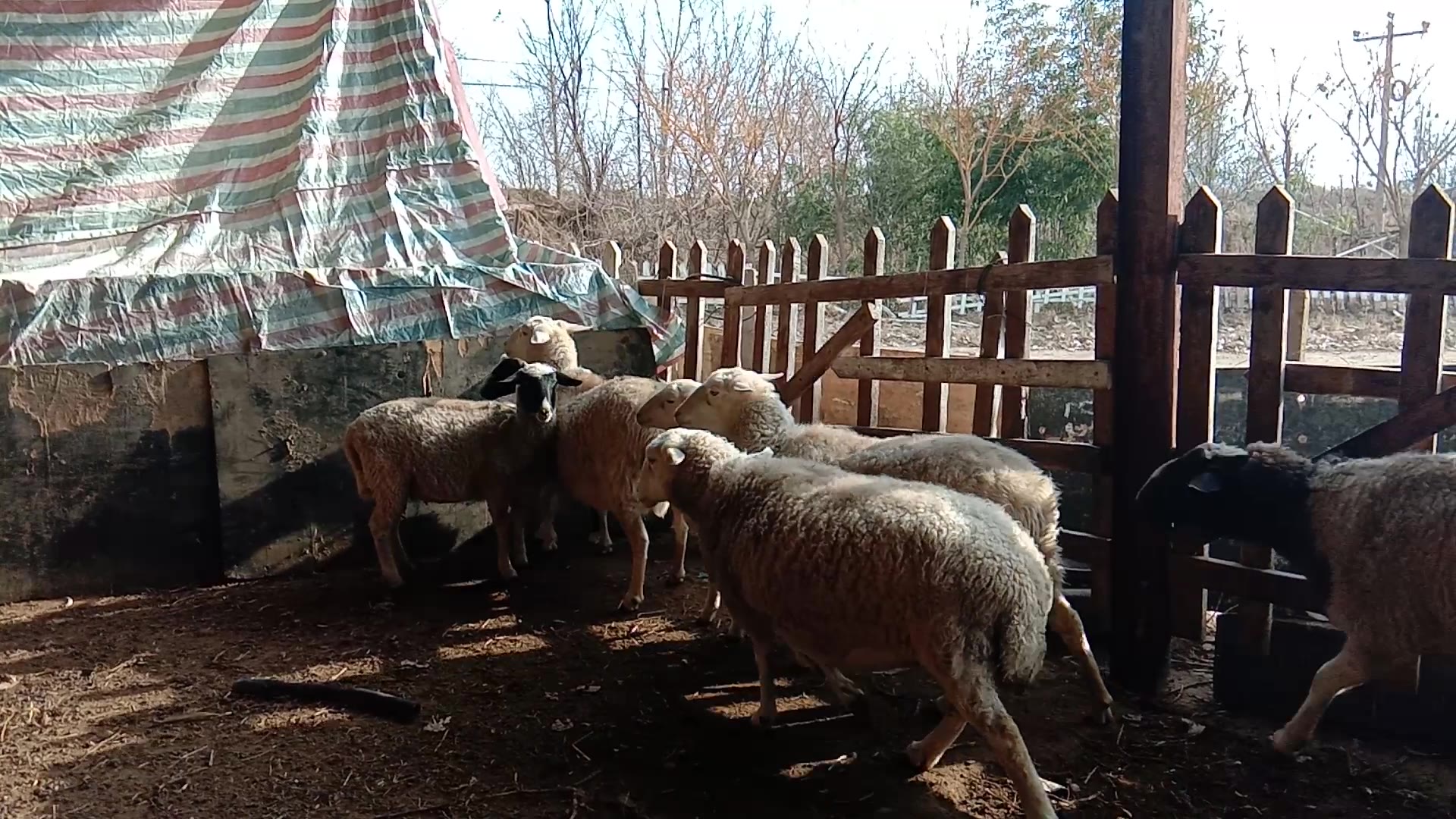} & 
 \includegraphics[width=0.19\linewidth]{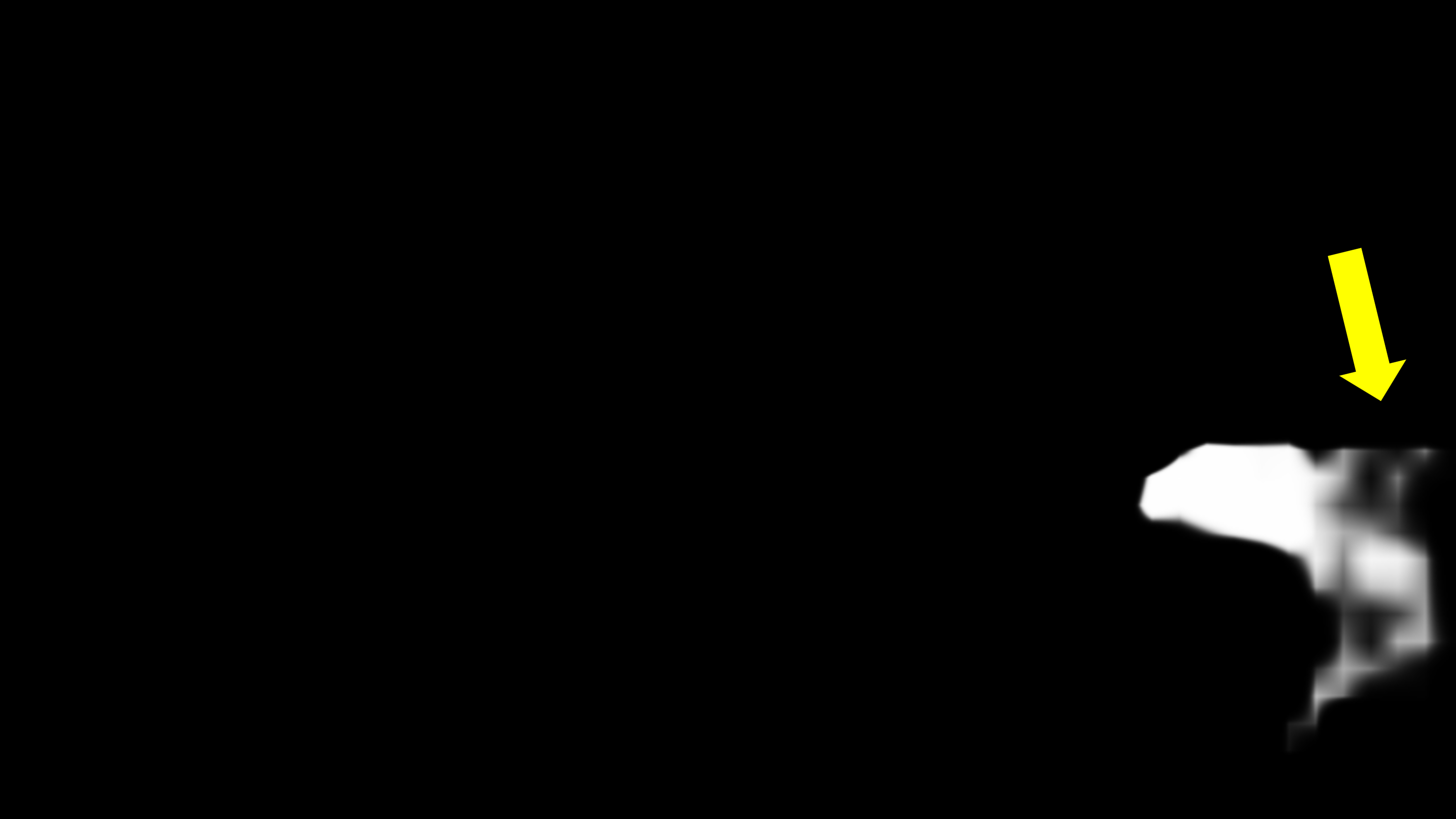} & 
 \includegraphics[width=0.19\linewidth]{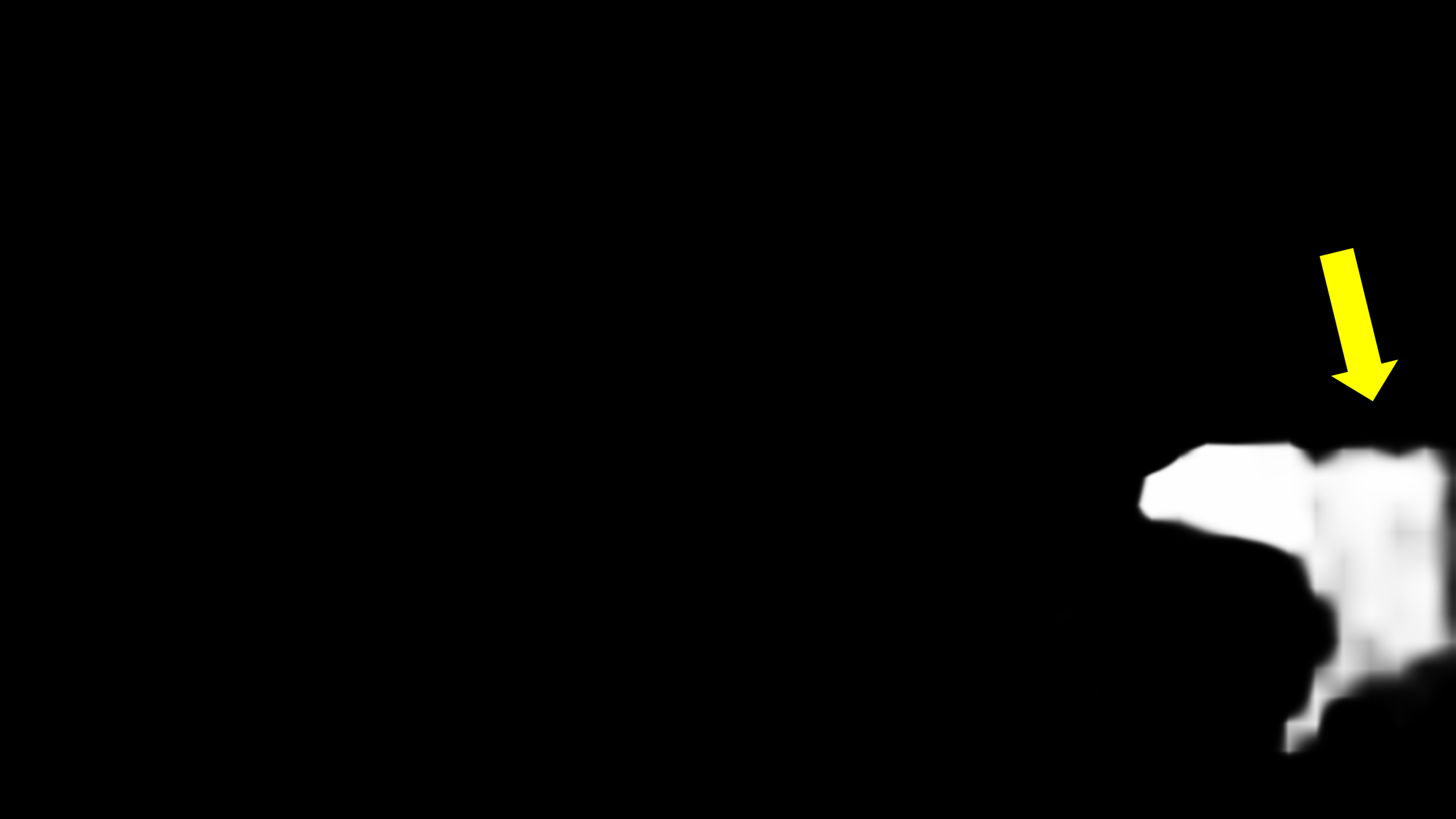} & 
 \includegraphics[width=0.19\linewidth]{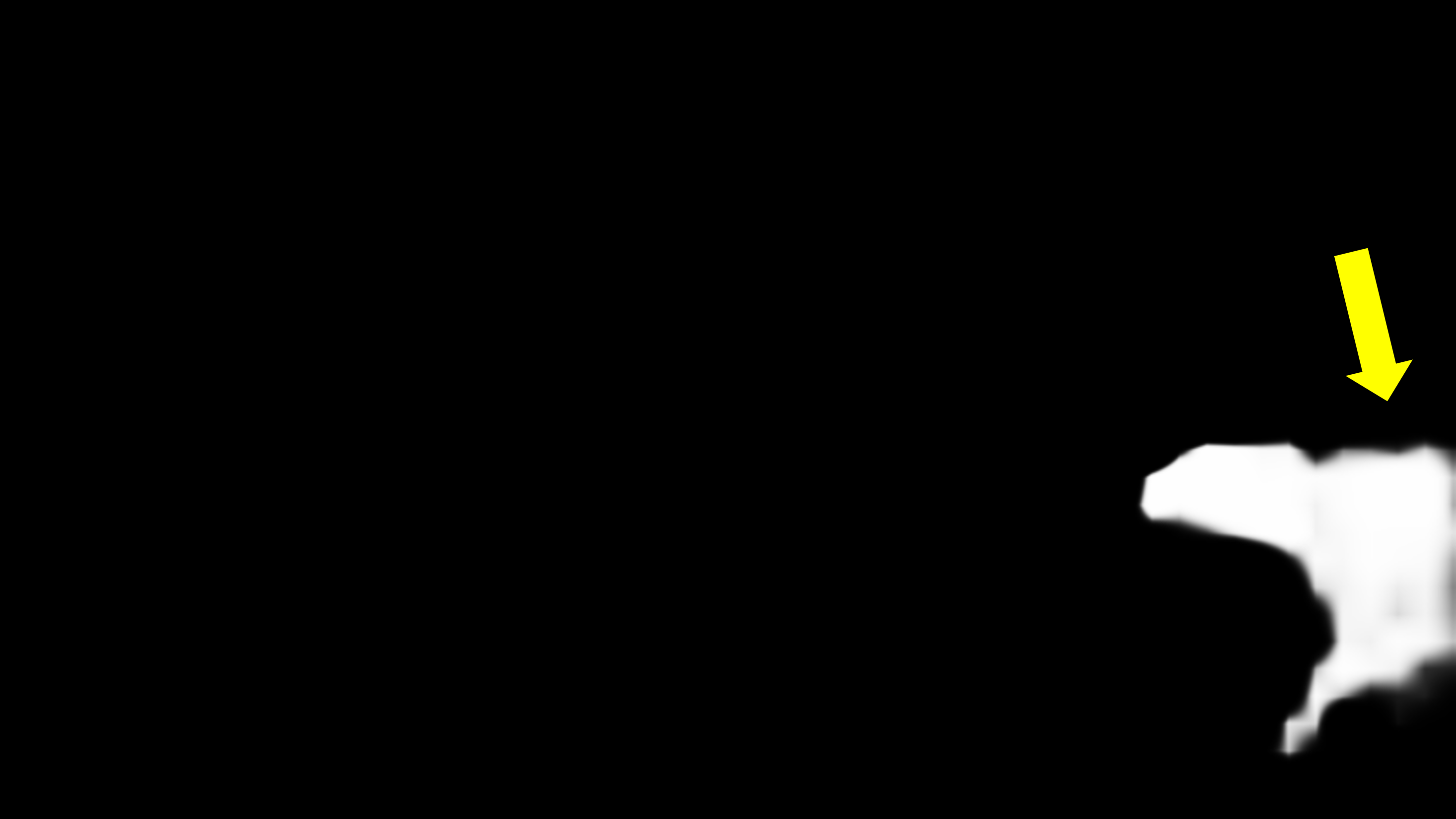} & 
 \includegraphics[width=0.19\linewidth]{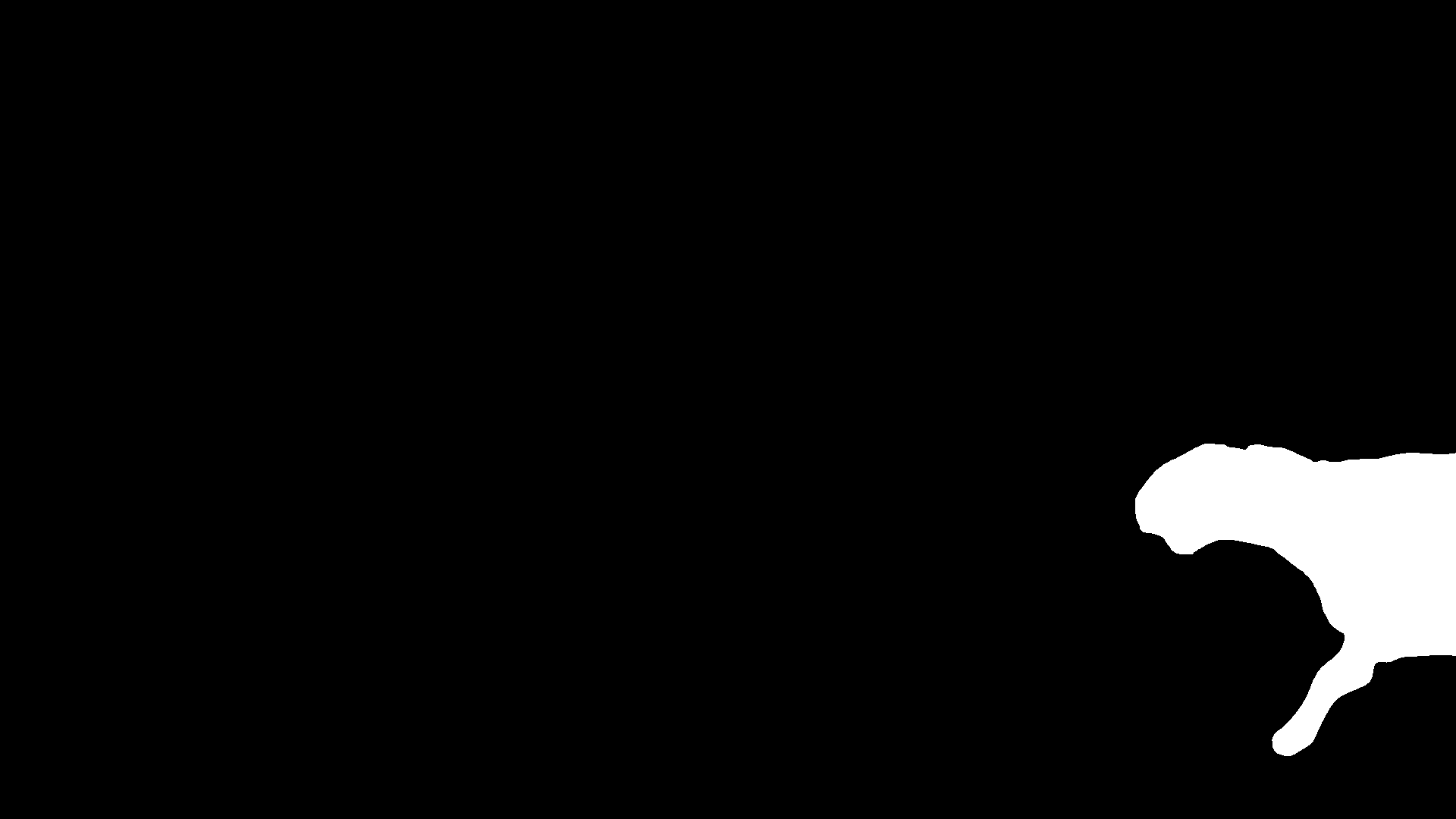} \\
 \includegraphics[width=0.19\linewidth]{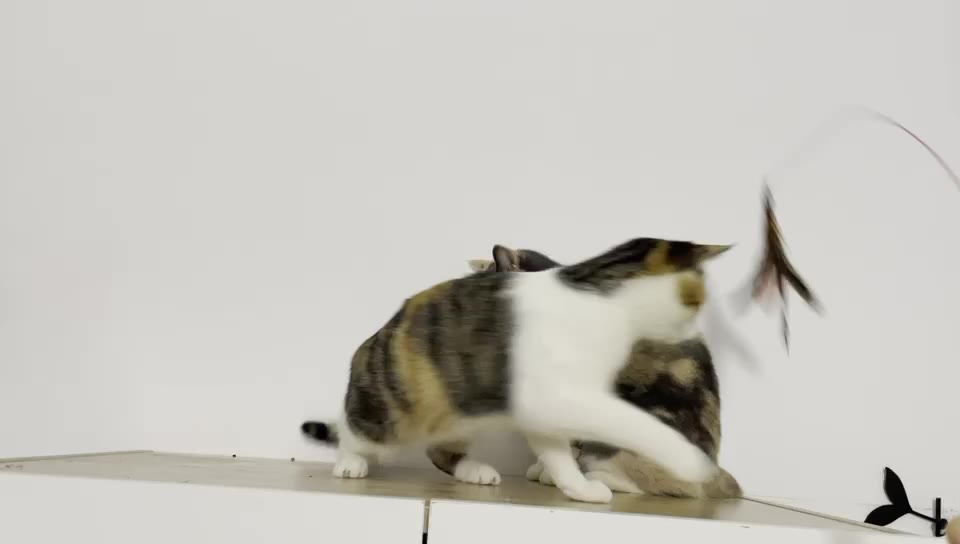} & 
 \includegraphics[width=0.19\linewidth]{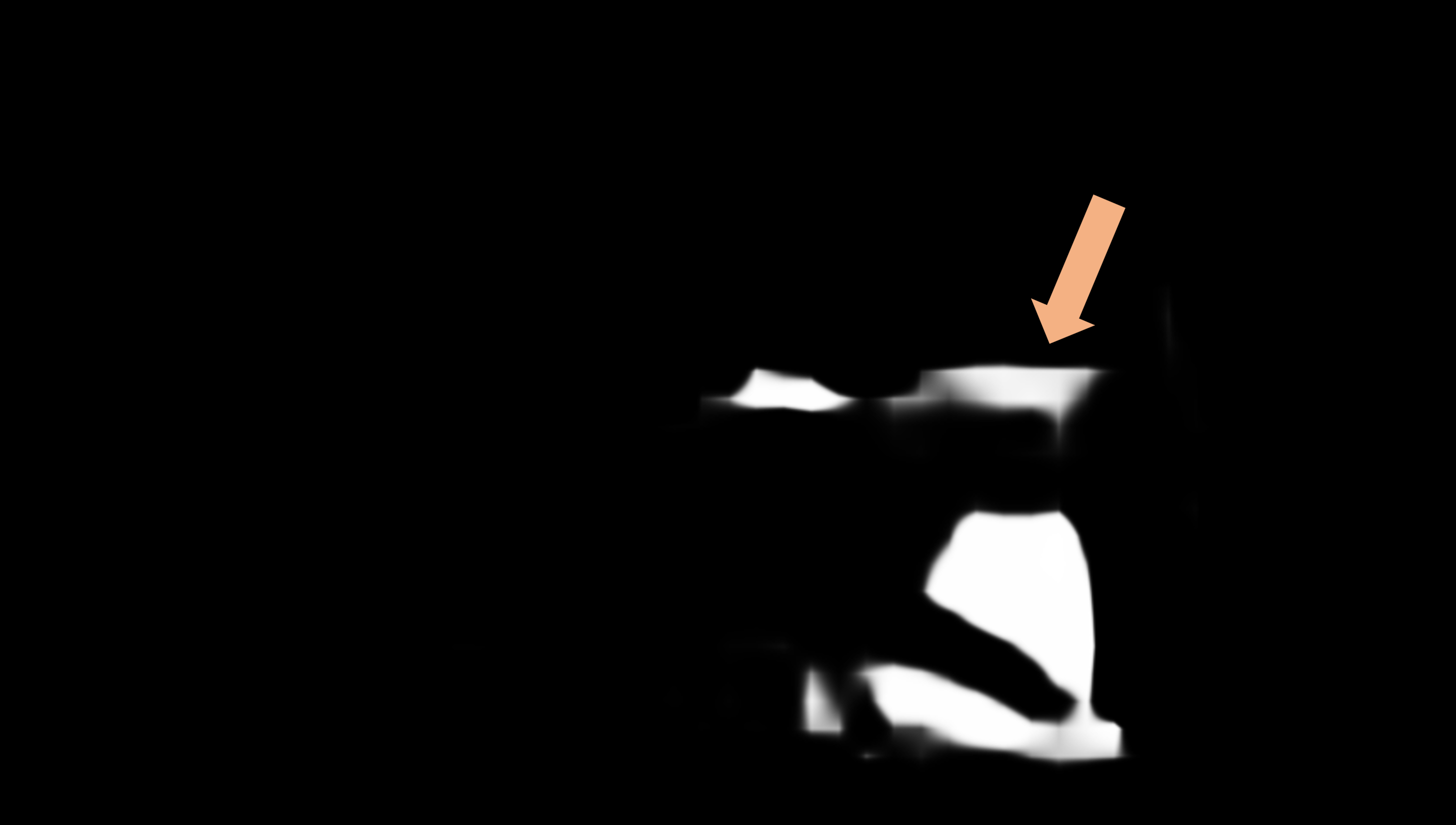} & 
 \includegraphics[width=0.19\linewidth]{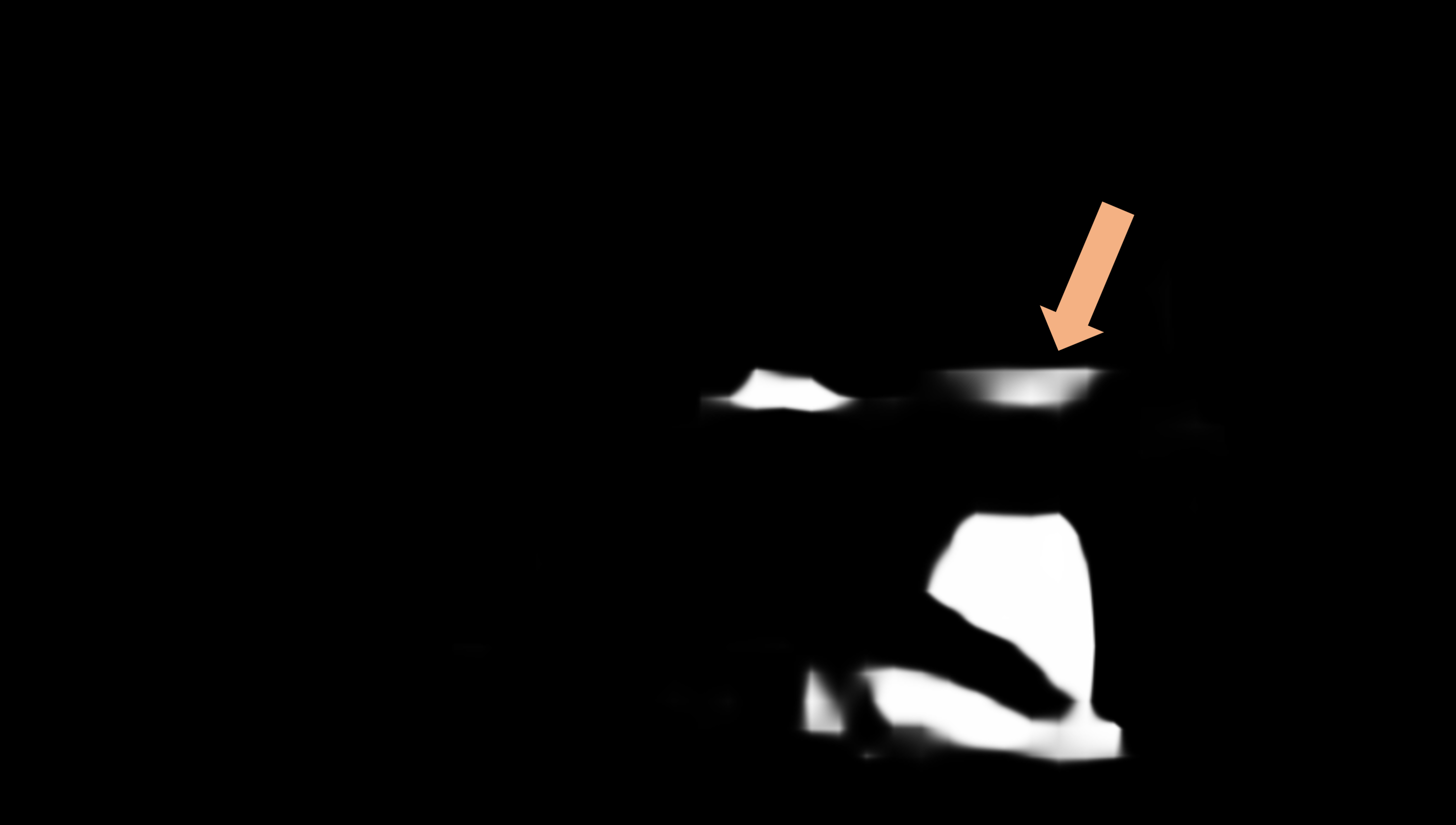} & 
 \includegraphics[width=0.19\linewidth]{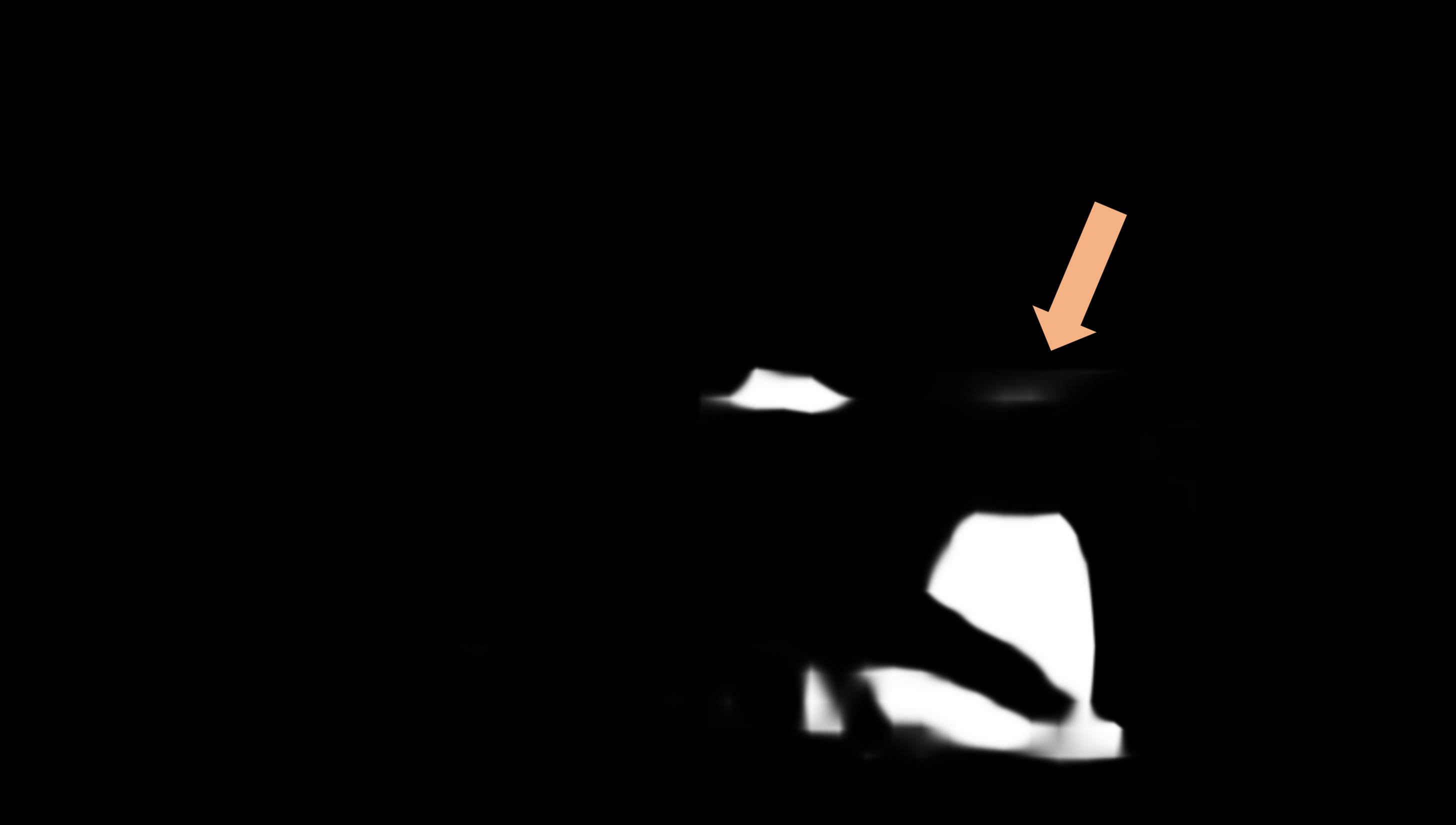} & 
 \includegraphics[width=0.19\linewidth]{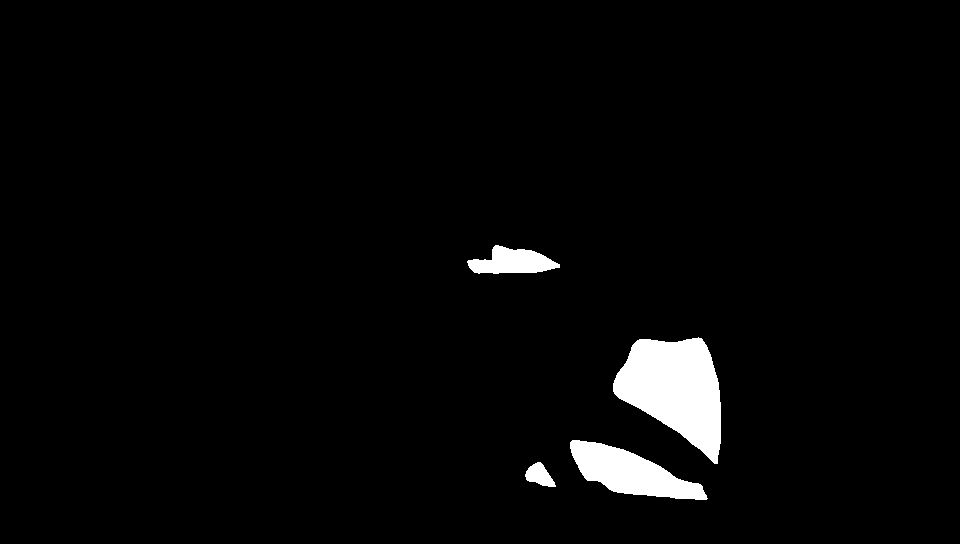} \\
 \includegraphics[width=0.19\linewidth]{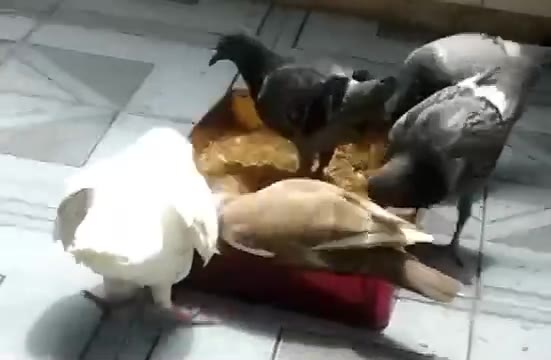} & 
 \includegraphics[width=0.19\linewidth]{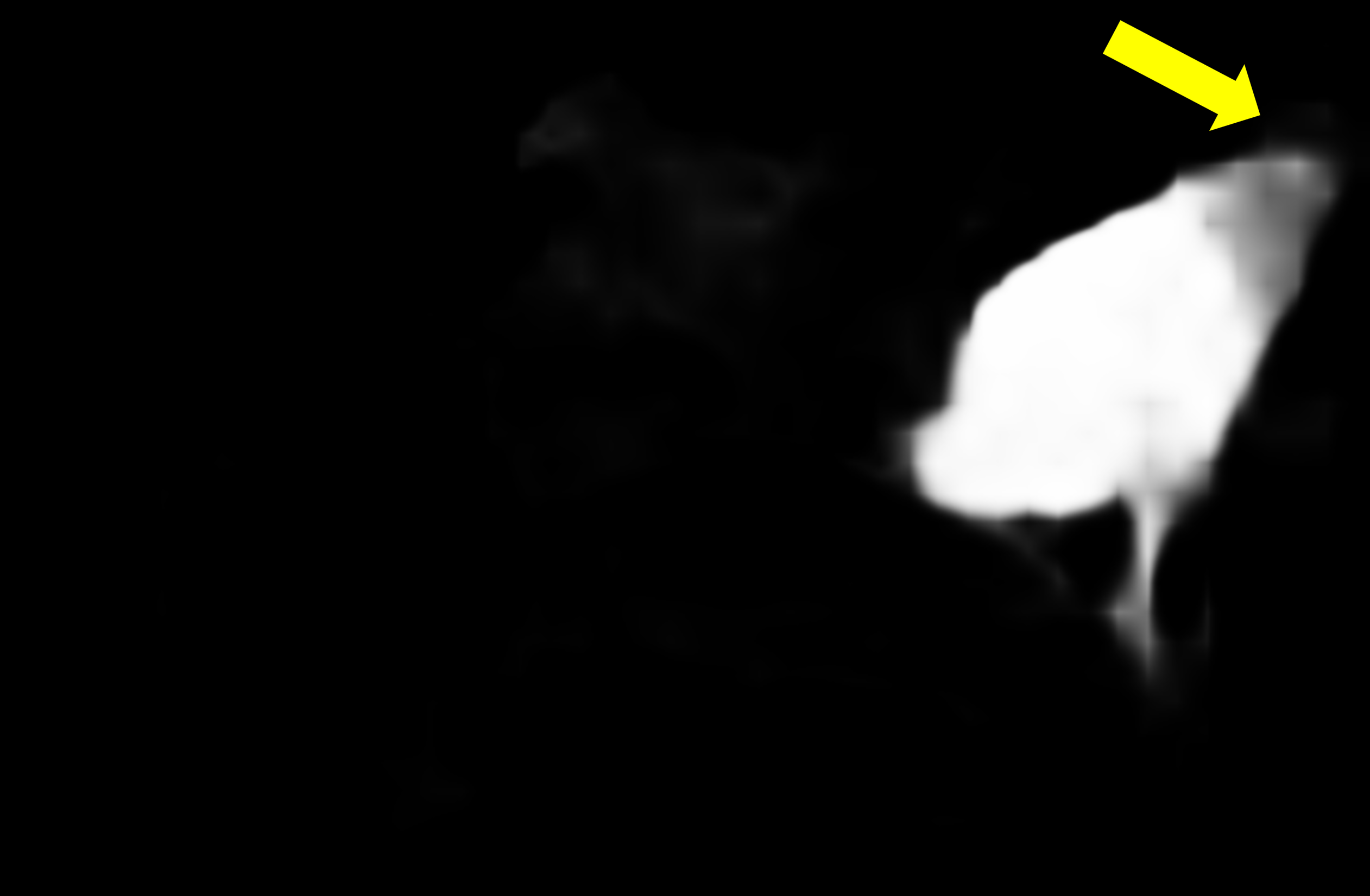} & 
 \includegraphics[width=0.19\linewidth]{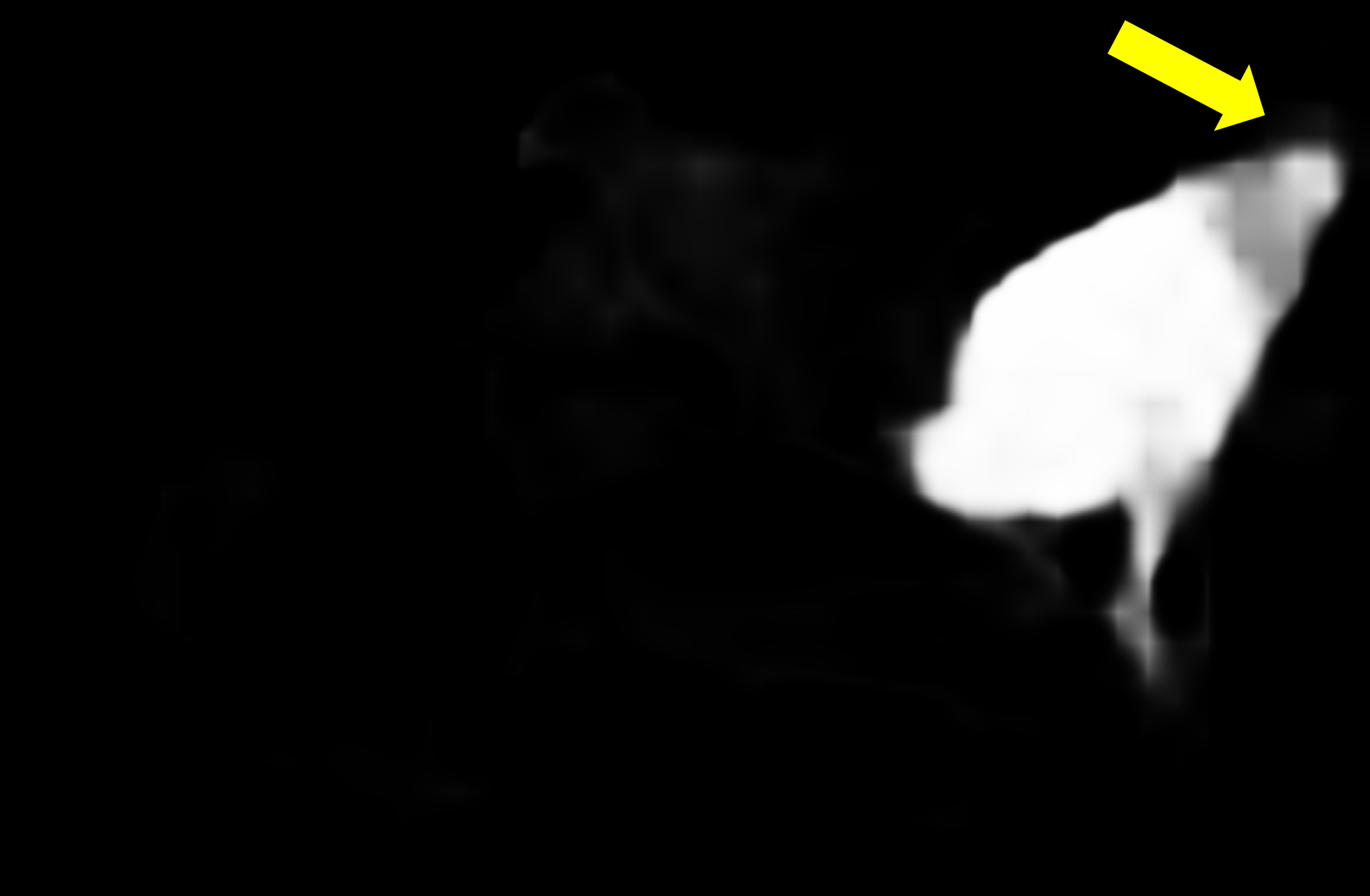} & 
 \includegraphics[width=0.19\linewidth]{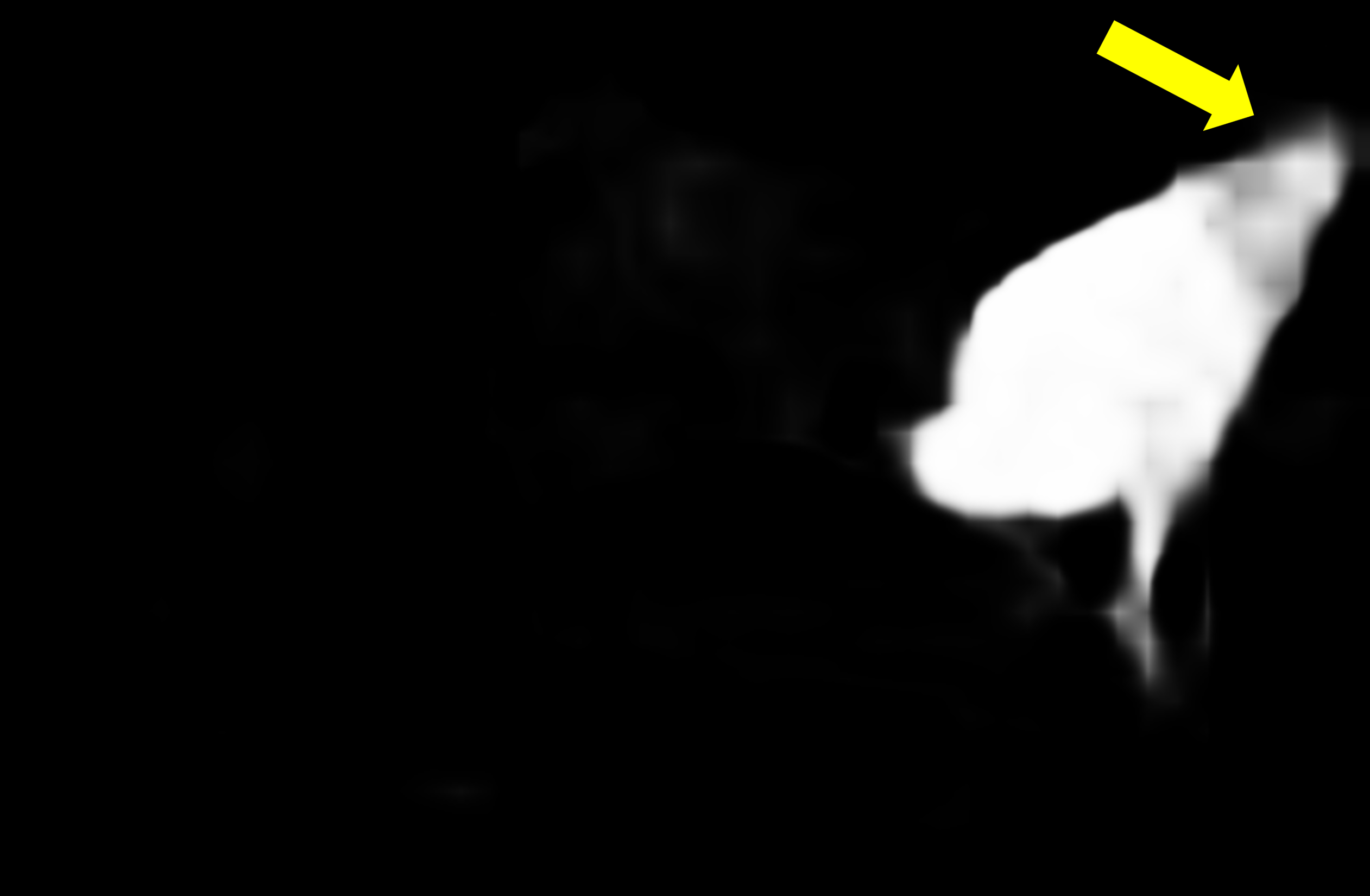} & 
 \includegraphics[width=0.19\linewidth]{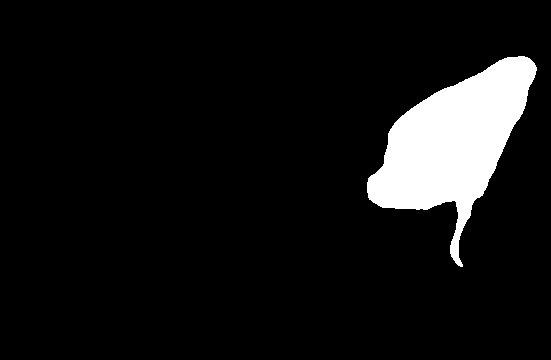} \\
\end{tabular}

    \caption{Visualization of auxiliary masks $M_l$ in the $l$-th object transformer block. At every layer, matching errors are suppressed (pink arrows) and the target object becomes more coherent (yellow arrows). The ground-truth is annotated by us for reference.}
    \label{fig:app:feature-progression}
\end{figure}

\subsection{Feature Progression in the Object Transformer}
Figure~\ref{fig:app:feature-progression} visualizes additional feature progressions within the object transformer (in addition to Figure~4 in the main paper).
The object transformer helps to suppress noises from low-level matching and produces more coherent object-level features.

\subsection{Benefits of Masked Attention/Object Transformer}
\begin{figure}[t]
    \centering
    \centering
\begin{tabular}{c@{\hspace{2pt}}c@{\hspace{2pt}}c}
& w/o masked attn & w/ masked attn \\
\rotatebox[origin=c]{90}{Simple} & 
    \raisebox{-0.5\height}{\includegraphics[width=0.3\linewidth]{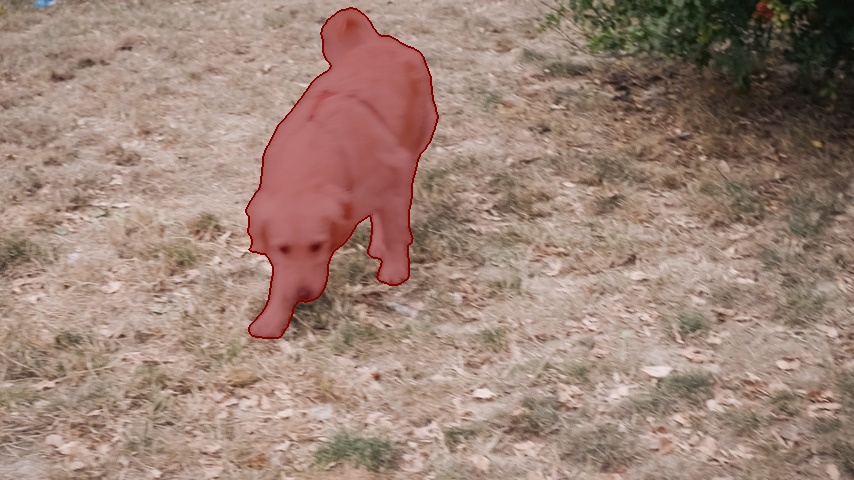}}\vspace{1px} & 
    \raisebox{-0.5\height}{\includegraphics[width=0.3\linewidth]{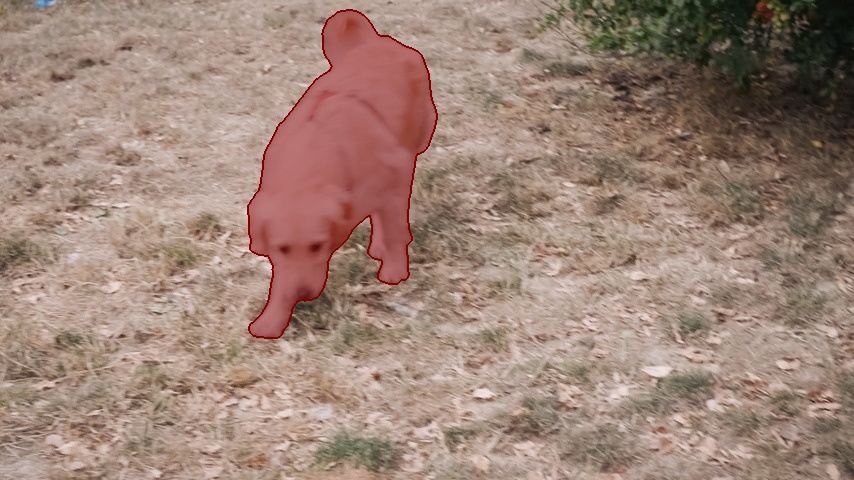}}\vspace{1px}\\
\rotatebox[origin=c]{90}{Cluttered} & 
    \raisebox{-0.5\height}{\includegraphics[width=0.3\linewidth]{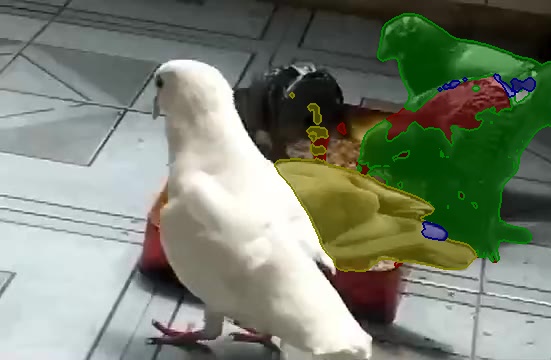}} & 
    \raisebox{-0.5\height}{\includegraphics[width=0.3\linewidth]{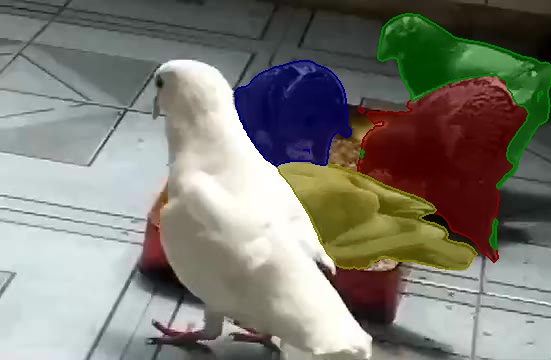}} \\
\end{tabular}

    \caption{Comparisons of Cutie with/without masked attention. While both work well in simple cases, masked attention helps to differentiate similarly-looking objects.}
    \label{fig:app:cmp-masked-attn}
\end{figure}

\begin{figure}[t]
    \centering
    \centering
\begin{tabular}{c@{\hspace{2pt}}c@{\hspace{2pt}}c@{\hspace{2pt}}c}
\includegraphics[width=0.24\linewidth]{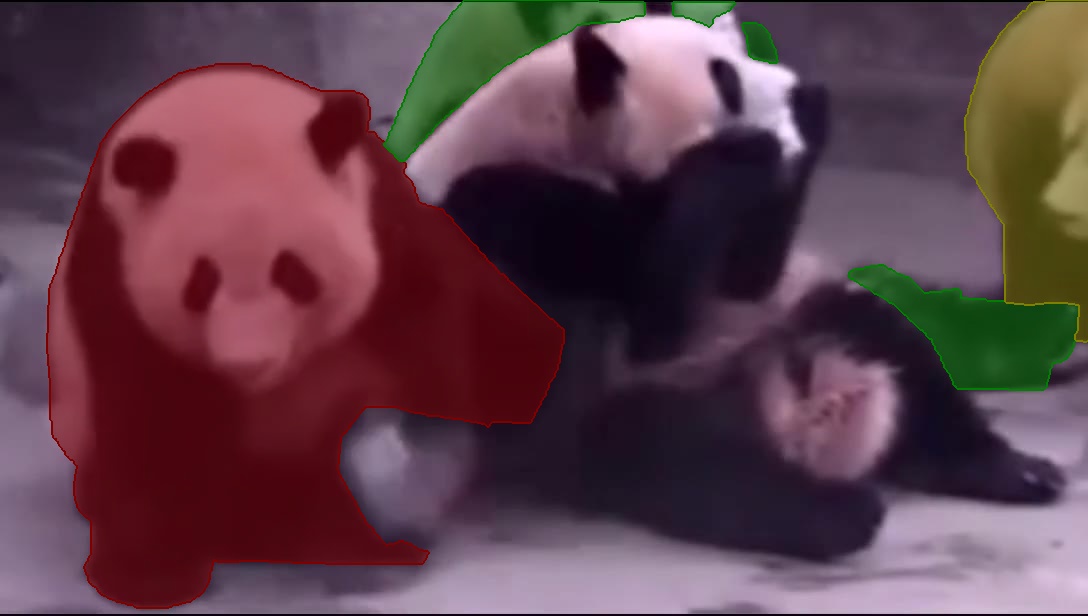} & 
\includegraphics[width=0.24\linewidth]{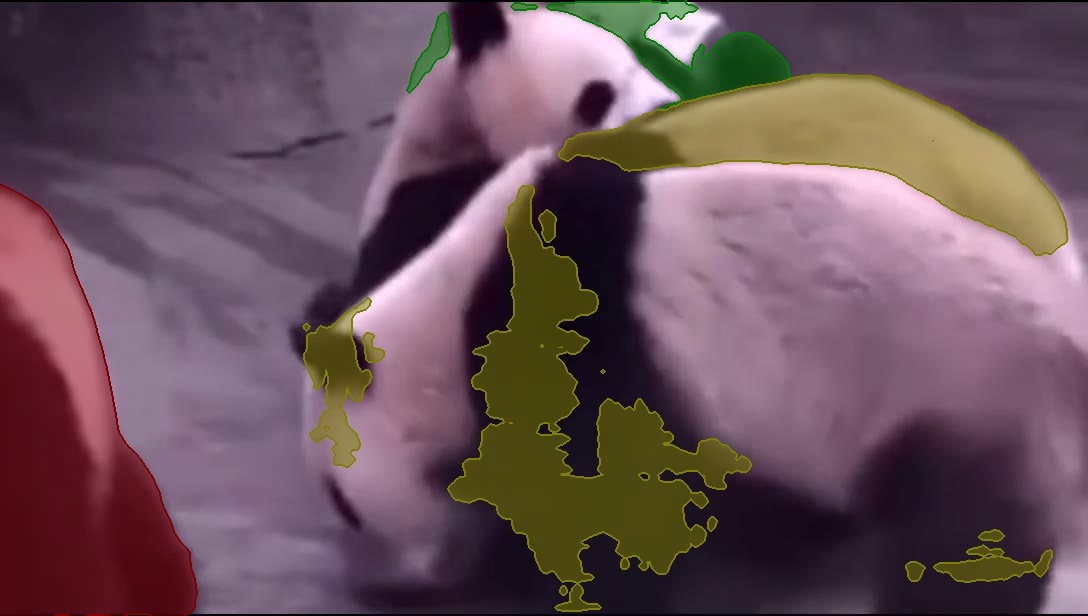} & 
\includegraphics[width=0.24\linewidth]{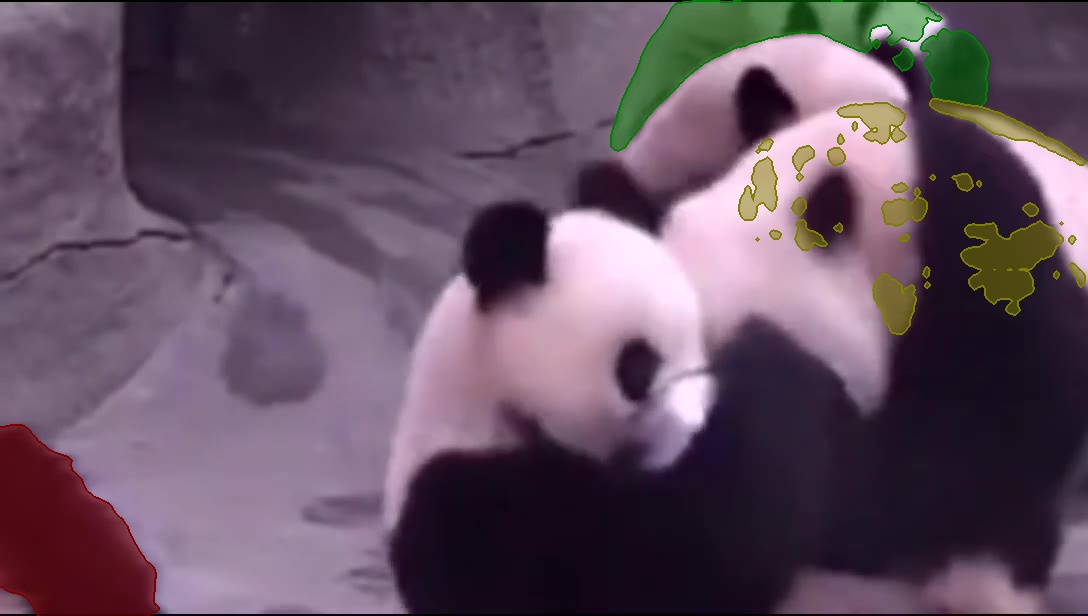} & 
\includegraphics[width=0.24\linewidth]{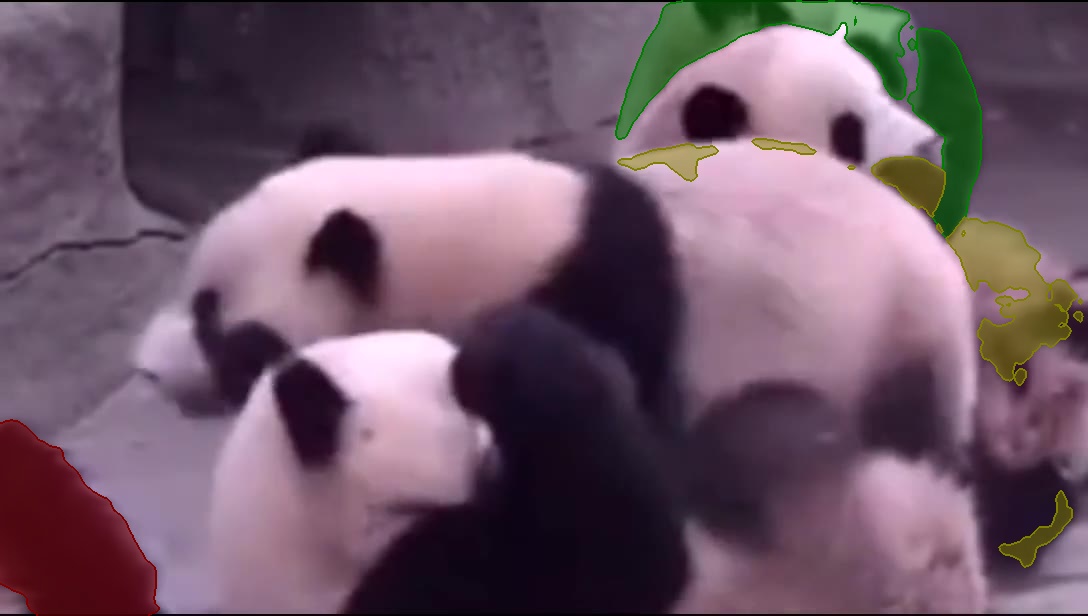} \\
\includegraphics[width=0.24\linewidth]{rebuttal/default/7e52d/00000.jpg} & 
\includegraphics[width=0.24\linewidth]{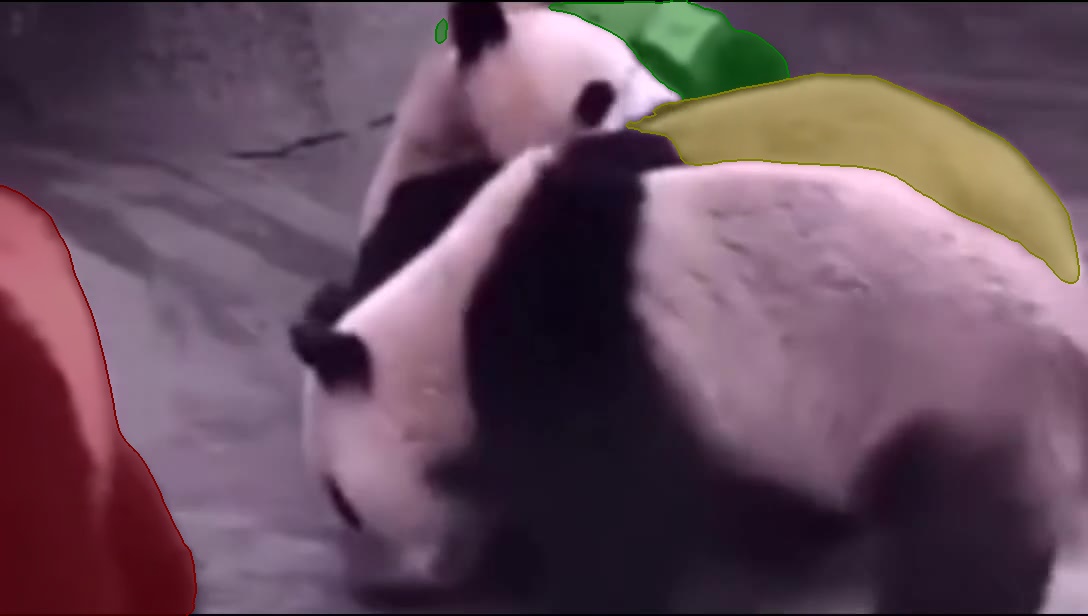} & 
\includegraphics[width=0.24\linewidth]{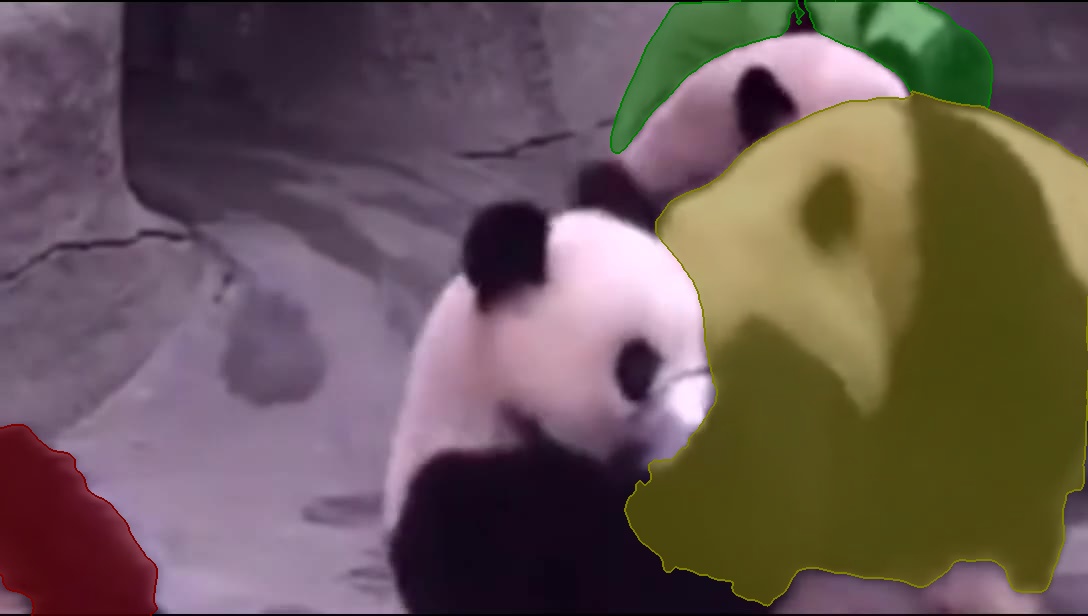} & 
\includegraphics[width=0.24\linewidth]{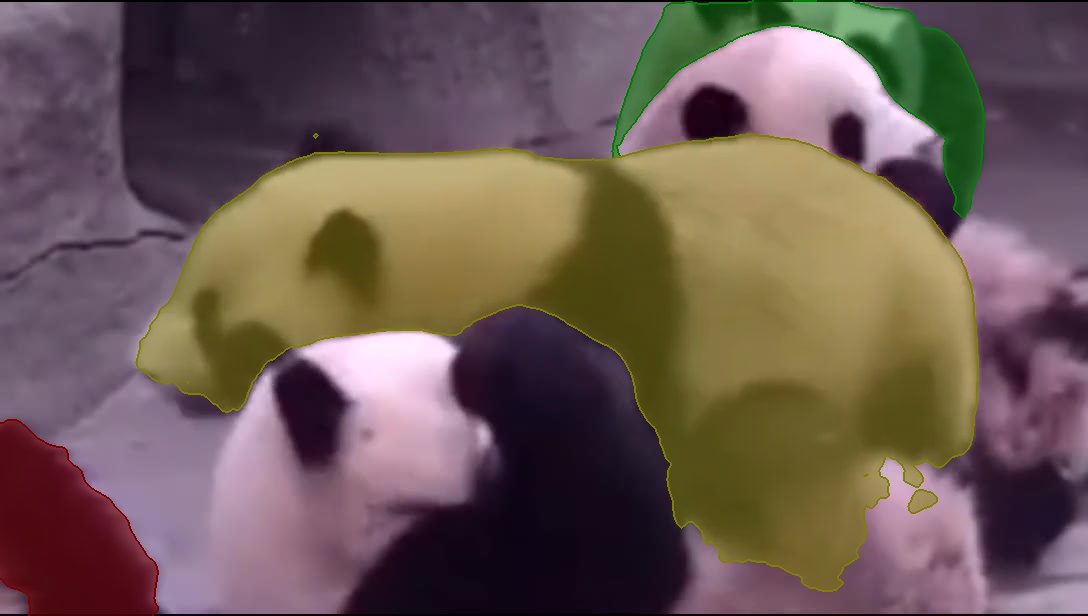} \\
\includegraphics[width=0.24\linewidth]{rebuttal/default/7e52d/00000.jpg} & 
\includegraphics[width=0.24\linewidth]{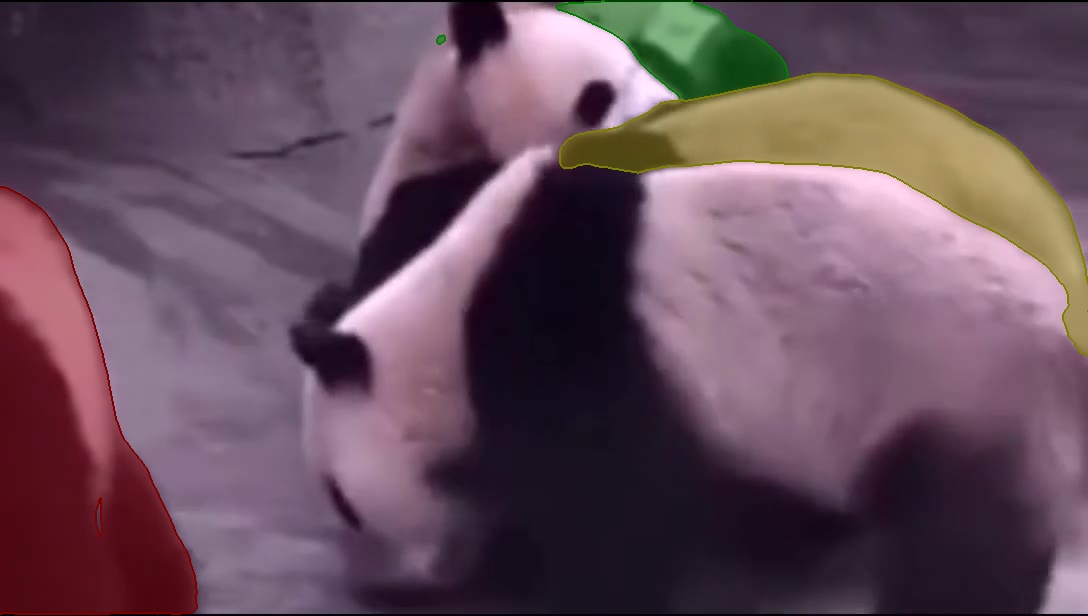} & 
\includegraphics[width=0.24\linewidth]{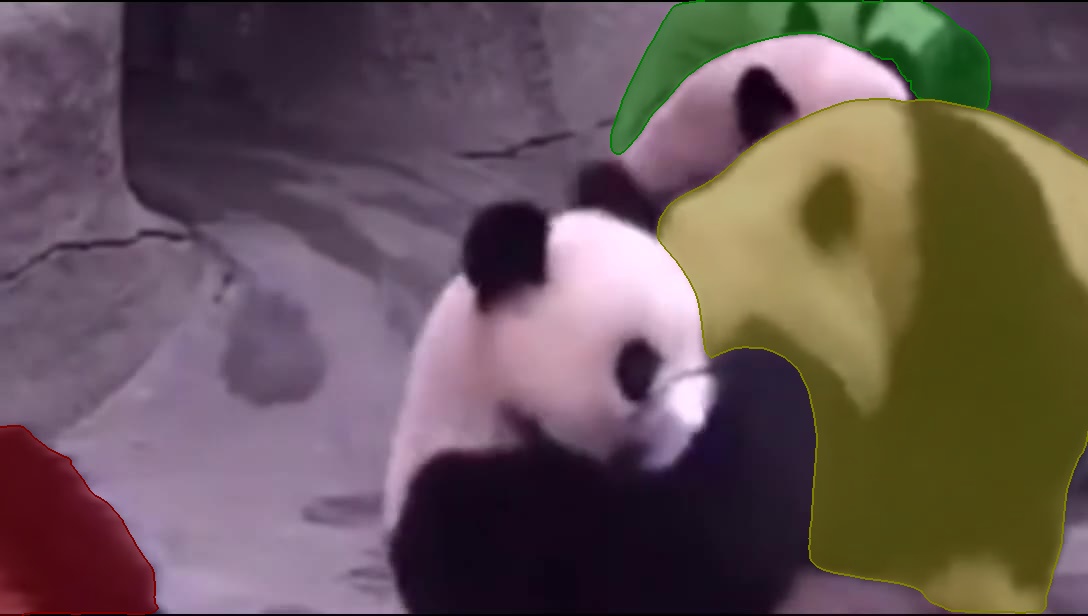} & 
\includegraphics[width=0.24\linewidth]{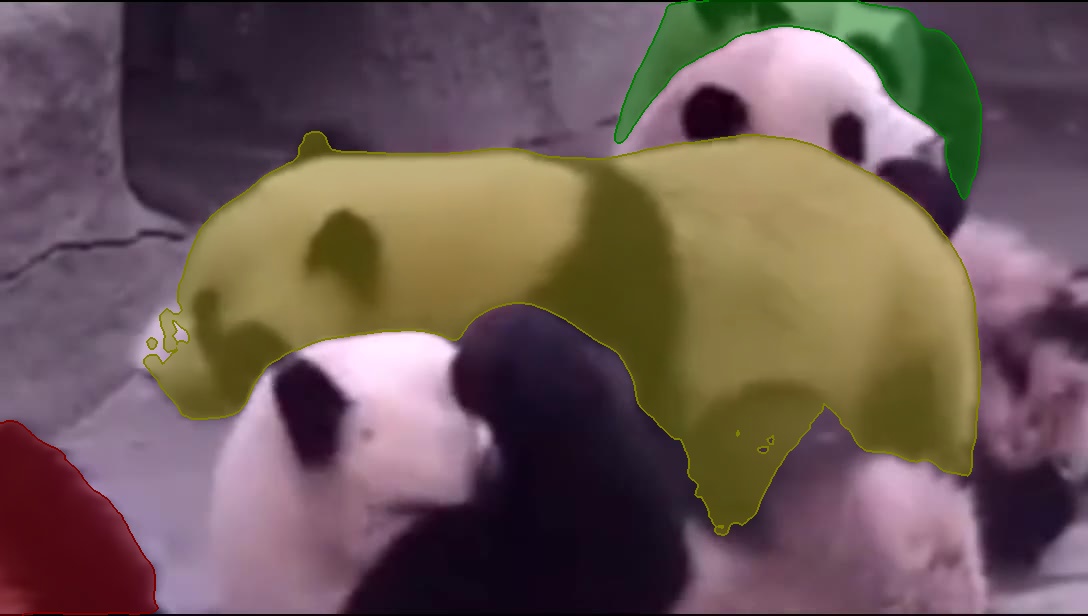} 
\end{tabular}

    \caption{Top-to-bottom: Without object queries, Cutie's default model, and ground-truth.
    The leftmost frame is a reference frame.}
    \label{fig:app:cmp-object-transformer}
\end{figure}

Figure~\ref{fig:app:cmp-masked-attn} qualitatively compares results with/without using masked attention -- while both work well for simpler cases, masked attention helps in challenging cases with similar objects.
Figure~\ref{fig:app:cmp-object-transformer} visualizes the benefits of the object transformer.
Using the object transformer leads to more complete and accurate outputs.

\section{Details on BURST Evaluation}\label{sec:app:burst-details}
In BURST~\cite{athar2023burst}, we update the memory every $10$th frame following~\cite{cheng2022xmem}.
Since BURST contains high-resolution images (e.g., 1920$\times$1200), we downsize the images such that the shorter edge has no more than 600 pixels instead of the default 480 pixels for all methods.
Following~\cite{athar2023burst}, we assess Higher Order Tracking Accuracy (HOTA)~\cite{luiten2021hota} on common and uncommon object classes separately.

For better performance on long videos, we experiment with the long-term memory~\cite{cheng2022xmem} in addition to our default FIFO memory strategy.
The long-term memory is a plug-in addition to our pixel memory -- it routinely compresses the attentional ``working memory'' into a long-term memory storage instead of discarding them as in our first-in-first-out approach. 
The long-term memory can be adopted without any re-training. 
We follow the default long-term memory parameters in XMem~\cite{cheng2022xmem} and present the improvement in the main paper.

\section{Additional Quantitative Results}\label{sec:app:quantitative}

\subsection{Speed-Accuracy Trade-off}
We note that the performance of Cutie can be further improved by changing hyperparameters like memory interval and the size of the memory bank during inference, at the cost of a slower running time.
Here, we present ``Cutie+'', which adjusts the following hyperparameters without re-training:
\begin{enumerate}
    \item Maximum memory frames $T_{\max} = 5 \to T_{\max} = 10$
    \item Memory interval $r=5 \to r=3$
    \item Maximum shorter side resolution during inference $480\to720$ pixels
\end{enumerate}
These settings apply to DAVIS~\cite{perazzi2016benchmark} and MOSE~\cite{ding2023mose}. 
For YouTubeVOS, we keep the memory interval $r=5$ and set the maximum shorter side resolution during inference to $600$ for two reasons: 1) YouTubeVOS is annotated every 5 frames, and aligning the memory interval with annotation avoids adding unannotated objects into the memory as background, and 2) YouTubeVOS has lower video quality and using higher resolution makes artifacts more apparent.
The results of Cutie+ are tabulated in the bottom portion of Table~\ref{tab:app:external-training}. 

\subsection{Comparisons with Methods that Use External Training}

\begin{table}
    \small
\centering
\begin{NiceTabular}
{l@{\hspace{2pt}}l@{\hspace{6pt}}C{2.2em}@{}C{2.2em}@{}C{2.2em}@{\hspace{5pt}}C{2.2em}@{}C{2.2em}@{}C{2.2em}@{\hspace{5pt}}C{2.2em}@{}C{2.2em}@{}C{2.2em}@{\hspace{5pt}}C{2.2em}@{}C{2.2em}@{}C{2.2em}@{}C{2.2em}@{}C{2.2em}@{}R{2em}}[colortbl-like]
\toprule
&& \multicolumn{3}{c}{\small MOSE} & \multicolumn{3}{c}{\small DAVIS-17 val} & \multicolumn{3}{c}{\small DAVIS-17 test} & \multicolumn{6}{c}{\small YouTubeVOS-2019 val} \\
\cmidrule(lr{\dimexpr 4\tabcolsep+8pt}){3-6} \cmidrule(lr{\dimexpr 4\tabcolsep+8pt}){6-9} \cmidrule(lr{\dimexpr 4\tabcolsep+8pt}){9-12} \cmidrule(lr){12-17}
Method && \mjf & \mj & \mf & \mjf & \mj & \mf & \mjf & \mj & \mf & \mg & \mjs & \mfs & \mju & \mfu & FPS \\
\toprule
SimVOS-B~\cite{wu2023scalable} &  & - & - & - & 81.3 & 78.8 & 83.8 & - & - & - & - & - & - & - & - & 3.3 \\
SimVOS-B~\cite{wu2023scalable} & w/ MAE~\cite{he2021masked} & - & - & - & 88.0 & 85.0 & 91.0 & 80.4 & 76.1 & 84.6 & 84.2 & 83.1 & - & 79.1 & - & 3.3 \\
JointFormer~\cite{zhang2023joint} &  & - & - & - & - & - & - & 65.6 & 61.7 & 69.4 & 73.3 & 75.2 & 78.5 & 65.8 & 73.6 & 3.0 \\
JointFormer~\cite{zhang2023joint} & w/ MAE~\cite{he2021masked}  & - & - & - & 89.7 & 86.7 & 92.7 & 87.6 & 84.2 & 91.1 & 87.0 & 86.1 & 90.6 & 82.0 & 89.5 & 3.0 \\
JointFormer~\cite{zhang2023joint} & w/ MAE~\cite{he2021masked} + BL30K~\cite{cheng2021mivos} & - & - & - & 90.1 & 87.0 & \textbf{93.2} & \textbf{88.1} & \textbf{84.7} & \textbf{91.6} & \textbf{87.4} & \textbf{86.5} & \textbf{90.9} & 82.0 & \textbf{90.3} & 3.0 \\
ISVOS~\cite{wang2022look} & & - & - & - & 80.0 & 76.9 & 83.1 & - & - & - & - & - & - & - & - & 5.8$^\ast$  \\
ISVOS~\cite{wang2022look} & w/ COCO~\cite{lin2014microsoft} & - & - & - & 87.1 & 83.7 & 90.5 & 82.8 & 79.3 & 86.2 & 86.1 & 85.2 & 89.7 & 80.7 & 88.9 & 5.8$^\ast$  \\
ISVOS~\cite{wang2022look} & w/ COCO~\cite{lin2014microsoft} + BL30K~\cite{cheng2021mivos} & - & - & - & 88.2 & 84.5 & 91.9 & 84.0 & 80.1 & 87.8 & 86.3 & 85.2 & 89.7 & 81.0 & 89.1 & 5.8$^\ast$  \\
\rowcolor{defaultColor}
Cutie-small & & 62.2 & 58.2 & 66.2 & 87.2 & 84.3 & 90.1 & 84.1 & 80.5 & 87.6 & 86.2 & 85.3 & 89.6 & 80.9 & 89.0 & \textbf{45.5} \\
\rowcolor{defaultColor}
Cutie-base & & 64.0 & 60.0 & 67.9 & 88.8 & 85.4 & 92.3 & 84.2 & 80.6 & 87.7 & 86.1 & 85.5 & 90.0 & 80.6 & 88.3 & 36.4 \\
\midrule
\rowcolor{defaultColor}
Cutie-small & w/ MOSE~\cite{ding2023mose} & 67.4 & 63.1 & 71.7 & 86.5 & 83.5 & 89.5 & 83.8 & 80.2 & 87.5 & 86.3 & 85.2 & 89.7 & 81.1 & 89.2 & \textbf{45.5} \\
\rowcolor{defaultColor}
Cutie-base & w/ MOSE~\cite{ding2023mose} & 68.3 & 64.2 & 72.3 & 88.8 & 85.6 & 91.9 & 85.3 & 81.4 & 89.3 & 86.5 & 85.4 & 90.0 & 81.3 & 89.3 & 36.4 \\
\rowcolor{defaultColor}
Cutie-small & w/ MEGA & 68.6 & 64.3 & 72.9 & 87.0 & 84.0 & 89.9 & 85.3 & 81.4 & 89.2 & 86.8 & 85.2 & 89.6 & 82.1 & \textbf{90.4} & \textbf{45.5} \\
\rowcolor{defaultColor}
Cutie-base & w/ MEGA & 69.9 & 65.8 & 74.1 & 87.9 & 84.6 & 91.1 & 86.1 & 82.4 & 89.9 & 87.0 & 86.0 & 90.5 & 82.0 & 89.6 & 36.4 \\
\rowcolor{defaultColor}
Cutie-small+ & & 64.3 & 60.4 & 68.2 & 88.7 & 86.0 & 91.3 & 85.7 & 82.5 & 88.9 & 86.7 & 85.7 & 89.8 & 81.7 & 89.6 & 20.6 \\
\rowcolor{defaultColor}
Cutie-base+ & & 66.2 & 62.3 & 70.1 & \textbf{90.5} & \textbf{87.5} & \textbf{93.4} & 85.9 & 82.6 & 89.2 & 86.9 & 86.2 & 90.7 & 81.6 & 89.2 & 17.9 \\
\rowcolor{defaultColor}
Cutie-small+ & w/ MOSE~\cite{ding2023mose} & 69.0 & 64.9 & 73.1 & 89.3 & 86.4 & 92.1 & 86.7 & 83.4 & 90.1 & 86.5 & 85.4 & 89.7 & 81.6 & 89.2 & 20.6 \\
\rowcolor{defaultColor}
Cutie-base+ & w/ MOSE~\cite{ding2023mose} & 70.5 & 66.5 & 74.6 & 90.0 & 87.1 & 93.0 & 86.3 & 82.9 & 89.7 & 86.8 & 85.7 & 90.0 & 81.8 & 89.6 & 17.9 \\
\rowcolor{defaultColor}
Cutie-small+ & w/ MEGA & 70.3 & 66.0 & 74.5 & 89.3 & 86.2 & 92.5 & 87.1 & 83.8 & 90.4 & 86.8 & 85.4 & 89.5 & 82.3 & 90.0 & 20.6 \\
\rowcolor{defaultColor}
Cutie-base+ & w/ MEGA & \textbf{71.7} & \textbf{67.6} & \textbf{75.8} & 88.1 & 85.5 & 90.8 & \textbf{88.1} & \textbf{84.7} & \textbf{91.4} & \textbf{87.5} & \textbf{86.3} & 90.6 & \textbf{82.7} & \textbf{90.5} & 17.9 \\
\midrule
\bottomrule
\end{NiceTabular}

    \caption{Quantitative comparison on common video object segmentation benchmarks, including methods that use external training data.
    Recent vision-transformer-based methods~\cite{wang2022look,wu2023scalable,zhang2023joint} depend largely on pretraining, either with MAE~\cite{he2021masked} or pretraining a separate Mask2Former~\cite{cheng2022masked} network on COCO instance segmentation~\cite{lin2014microsoft}.
    Note they do not release code at the time of writing, and thus they cannot be reproduced on datasets that they do not report results on. 
    Cutie performs competitively to those recent (slow) transformer-based methods, especially with added training data.
    MEGA is the aggregated dataset consisting of DAVIS~\cite{perazzi2016benchmark}, YouTubeVOS~\cite{xu2018youtubeVOS}, MOSE~\cite{ding2023mose}, OVIS~\cite{qi2022occluded}, and BURST~\cite{athar2023burst}.
    $^\ast$estimated FPS.}
    \label{tab:app:external-training}
\end{table}

Here, we present comparisons with methods that use external training: SimVOS~\cite{shi2015hierarchicalECSSD}, JointFormer~\cite{zhang2023joint}, and ISVOS~\cite{wang2022look} in Table~\ref{tab:app:external-training}. 
Note, we could not obtain the code for these methods at the time of writing. 
ISVOS~\cite{wang2022look} does not report running time -- we estimate to the best of our ability with the following information: 
1) For the VOS branch, it uses XMem~\cite{cheng2022xmem} as the baseline with a first-in-first-out 16-frame memory bank, 
2) for the instance branch, it uses Mask2Former~\cite{cheng2022masked} with an unspecified backbone. Beneficially for ISVOS, we assume the lightest backbone (ResNet-50), and 
3) the VOS branch and the instance branch share a feature extraction backbone.
Our computation is as follows:
\begin{enumerate}
    \item Time per frame for XMem with a 16-frame first-in-first-out memory bank (from our testing): 75.2 ms
    \item Time per frame for Mask2Former with ResNet-50 backbone (from Mask2Former paper): 103.1 ms
    \item Time per frame of the doubled-counted feature extraction backbone (from our testing): 6.5 ms
\end{enumerate}
Thus, we estimate that ISVOS would take (75.2+103.1-6.5) = 171.8 ms per frame, which translates to 5.8 frames per second.

In an endeavor to reach a better performance with Cutie by adding more training data, we devise a ``MEGA'' training scheme that includes training on BURST~\cite{athar2023burst} and OVIS~\cite{qi2022occluded} in addition to DAVIS~\cite{perazzi2016benchmark}, YouTubeVOS~\cite{xu2018youtubeVOS}, and MOSE~\cite{ding2023mose}. 
We train for an additional 50K iterations in the MEGA setting.
The results are tabulated in the bottom portion of Table~\ref{tab:app:external-training}. 

\subsection{Results on YouTubeVOS-2018 and LVOS}\label{sec:app:lvos}

\begin{table}
    \centering
\begin{NiceTabular}
{l@{\hspace{2pt}}l@{\hspace{6pt}}C{2.2em}@{}C{2.2em}@{}C{2.2em}@{}C{2.2em}@{}C{2.2em}@{\hspace{5pt}}C{2.2em}@{}C{2.2em}@{}C{2.2em}@{\hspace{5pt}}C{2.2em}@{}C{2.2em}@{}C{2.2em}@{\hspace{5pt}}R{2em}}[colortbl-like]
\toprule
&& \multicolumn{5}{c}{\small YouTubeVOS-2018 val} & \multicolumn{3}{c}{\small LVOS val} & \multicolumn{3}{c}{\small LVOS test} & \\
\cmidrule(lr{\dimexpr 4\tabcolsep+8pt}){3-8} \cmidrule(lr{\dimexpr 4\tabcolsep+8pt}){8-11} \cmidrule(lr{\dimexpr 4\tabcolsep+8pt}){11-14} 
Method && \mg & \mjs & \mfs & \mju & \mfu & \mjf & \mj & \mf & \mjf & \mj & \mf & FPS \\
\midrule
DEVA~\cite{cheng2023tracking} &  & 85.9 & 85.5 & 90.1 & 79.7 & 88.2 & 58.3 & 52.8 & 63.8 & 54.0 & 49.0 & 59.0 & 25.3\\
DEVA~\cite{cheng2023tracking} & w/ MOSE~\cite{ding2023mose} & 85.8 & 85.4 & 90.1 & 79.7 & 88.2 & 55.9 & 51.1 & 60.7 & 56.5 & 52.2 & 60.8 & 25.3 \\
DDMemory~\cite{hong2022lvos} & & 84.1 & 83.5 & 88.4 & 78.1 & 86.5 & \textbf{60.7} & 55.0 & \textbf{66.3} & 55.0 & 49.9 & 60.2 & 18.7 \\
\rowcolor{defaultColor}
Cutie-small & & \textbf{86.3} & 85.5 & 90.1 & \textbf{80.6} & \textbf{89.0} & 58.8 & 54.6 & 62.9 & \textbf{57.2} & \textbf{53.7} & \textbf{60.7} & \textbf{45.5}\\
\rowcolor{defaultColor}
Cutie-base & & 86.1 & \textbf{85.8} & \textbf{90.5} & 80.0 & 88.0 & 60.1 & \textbf{55.9} & 64.2 & 56.2 & 51.8 & 60.5 & 36.4 \\
\midrule
\rowcolor{defaultColor}
Cutie-small & w/ MOSE~\cite{ding2023mose} & 86.8 & 85.7 & 90.4 & 81.6 & 89.7 & 60.7 & 55.6 & 65.8 & 56.9 & 53.5 & 60.2  & \textbf{45.5}\\
\rowcolor{defaultColor}
Cutie-base & w/ MOSE~\cite{ding2023mose} & 86.6 & 85.7 & 90.6 & 80.8 & 89.1 & 63.5 & 59.1 & 67.9 & 63.6 & 59.1 & 68.0 & 36.4  \\
\rowcolor{defaultColor}
Cutie-small & w/ MEGA & \textbf{86.9} & 85.5 & 90.1 & \textbf{81.7} & \textbf{90.2} & 62.9 & 58.3 & 67.4 & 66.4 & 61.9 & 70.9  & \textbf{45.5}\\
\rowcolor{defaultColor}
Cutie-base & w/ MEGA & \textbf{87.0} & \textbf{86.4} & \textbf{91.1} & 81.4 & 89.2 & \textbf{66.0} & \textbf{61.3} & \textbf{70.6} & \textbf{66.7} & \textbf{62.4} & \textbf{71.0} & 36.4  \\
\midrule
\bottomrule
\end{NiceTabular}

    \caption{Quantitative comparison on YouTubeVOS-2018~\cite{xu2018youtubeVOS} and LVOS~\cite{hong2022lvos}. 
    DDMemory~\cite{hong2022lvos} is the baseline method presented in LVOS~\cite{hong2022lvos} with no available official code at the time of writing.
    Note, we think LVOS is significantly different than other datasets because it contains a lot more tiny objects. See Section~\ref{sec:app:lvos} for details.
    MEGA is the aggregated dataset consisting of DAVIS~\cite{perazzi2016benchmark}, YouTubeVOS~\cite{xu2018youtubeVOS}, MOSE~\cite{ding2023mose}, OVIS~\cite{qi2022occluded}, and BURST~\cite{athar2023burst}.
    }
    \label{tab:app:lvos-results}
\end{table}

\begin{table}
    \centering
\begin{NiceTabular}{l@{\hspace{6pt}}l@{\hspace{6pt}}l@{\hspace{12pt}}c@{\hspace{12pt}}c@{\hspace{12pt}}c@{\hspace{12pt}}c@{\hspace{12pt}}c@{\hspace{12pt}}c@{\hspace{12pt}}c}[colortbl-like]
    \toprule
     & & & \multicolumn{3}{c}{BURST val} & \multicolumn{3}{c}{BURST test} & \\
     \cmidrule(lr{\dimexpr 4\tabcolsep+16pt}){4-7}
     \cmidrule(lr{\dimexpr 4\tabcolsep+48pt}){7-10}
     Method &&& All & Com. & Unc. & All & Com. & Unc. & Memory usage \\
     \midrule
     DeAOT~\cite{yang2022decoupling} & FIFO & w/ MOSE~\cite{ding2023mose} & 51.3 & 56.3 & 50.0 & 53.2 & 53.5 & 53.2 & 10.8G \\
     DeAOT~\cite{yang2022decoupling} & INF & w/ MOSE~\cite{ding2023mose} & 56.4 & 59.7 & 55.5 & 57.9 & 56.7 & 58.1 & 34.9G \\
     XMem~\cite{cheng2022xmem} & FIFO & w/ MOSE~\cite{ding2023mose} & 52.9 & 56.0 & 52.1 & 55.9 & 57.6 & 55.6 & 3.03G \\
     XMem~\cite{cheng2022xmem} & LT & w/ MOSE~\cite{ding2023mose} & 55.1 & 57.9 & 54.4 & 58.2 & 59.5 & 58.0 & 3.34G \\
     \rowcolor{defaultColor}
     Cutie-small & FIFO & w/ MOSE~\cite{ding2023mose} & 56.8 & 61.1 & 55.8 & 61.1 & 62.4 & 60.8 & \textbf{1.35G} \\
     \rowcolor{defaultColor}
     Cutie-small & LT & w/ MOSE~\cite{ding2023mose} & 58.3 & 61.5 & 57.5 & 61.6 & 63.1 & 61.3 & 2.28G \\
     \rowcolor{defaultColor}
     Cutie-base & LT & w/ MOSE~\cite{ding2023mose} & 58.4 & 61.8 & 57.5 & 62.6 & 63.8 & 62.3 & 2.36G \\
     \midrule
     \rowcolor{defaultColor}
     Cutie-small & LT & w/ MEGA & \textbf{61.6} & \textbf{65.3} & \textbf{60.6} & 64.4 & 63.7 & 64.6 & 2.28G \\
     \rowcolor{defaultColor}
     Cutie-base & LT & w/ MEGA & 61.2 & 65.0 & 60.3 & \textbf{66.0} & \textbf{66.5} & \textbf{65.9} & 2.36G \\
     \midrule
     \bottomrule
\end{NiceTabular}

    \caption{Extended comparisons of performance on long videos on the BURST dataset~\cite{athar2023burst}, including our results when trained in the MEGA setting.
    Com.\ and Unc.\ stand for common and uncommon objects respectively.
    Mem.: maximum GPU memory usage. FIFO: first-in-first-out memory bank; INF: unbounded memory; LT: long-term memory~\cite{cheng2022xmem}. DeAOT~\cite{yang2022decoupling} is not compatible with long-term memory.}
    \label{tab:app:burst-extended-results}
\end{table}

Here, we provide additional results on the YouTubeVOS-2018 validation set and LVOS~\cite{hong2022lvos} validation/test sets in Table~\ref{tab:app:lvos-results}.
FPS is measured on YoutubeVOS-2018/2019 following the main paper.
YouTubeVOS-2018 is the old version of YouTubeVOS-2019 -- we present our main results using YouTubeVOS-2019 and provide results on YouTubeVOS-2018 for reference.
Note that these results are ready at the time of paper submission and are referred to in the main paper. The complete tables are listed here due to space constraints in the main paper.

\begin{figure}
    \centering
    \includegraphics[width=\linewidth]{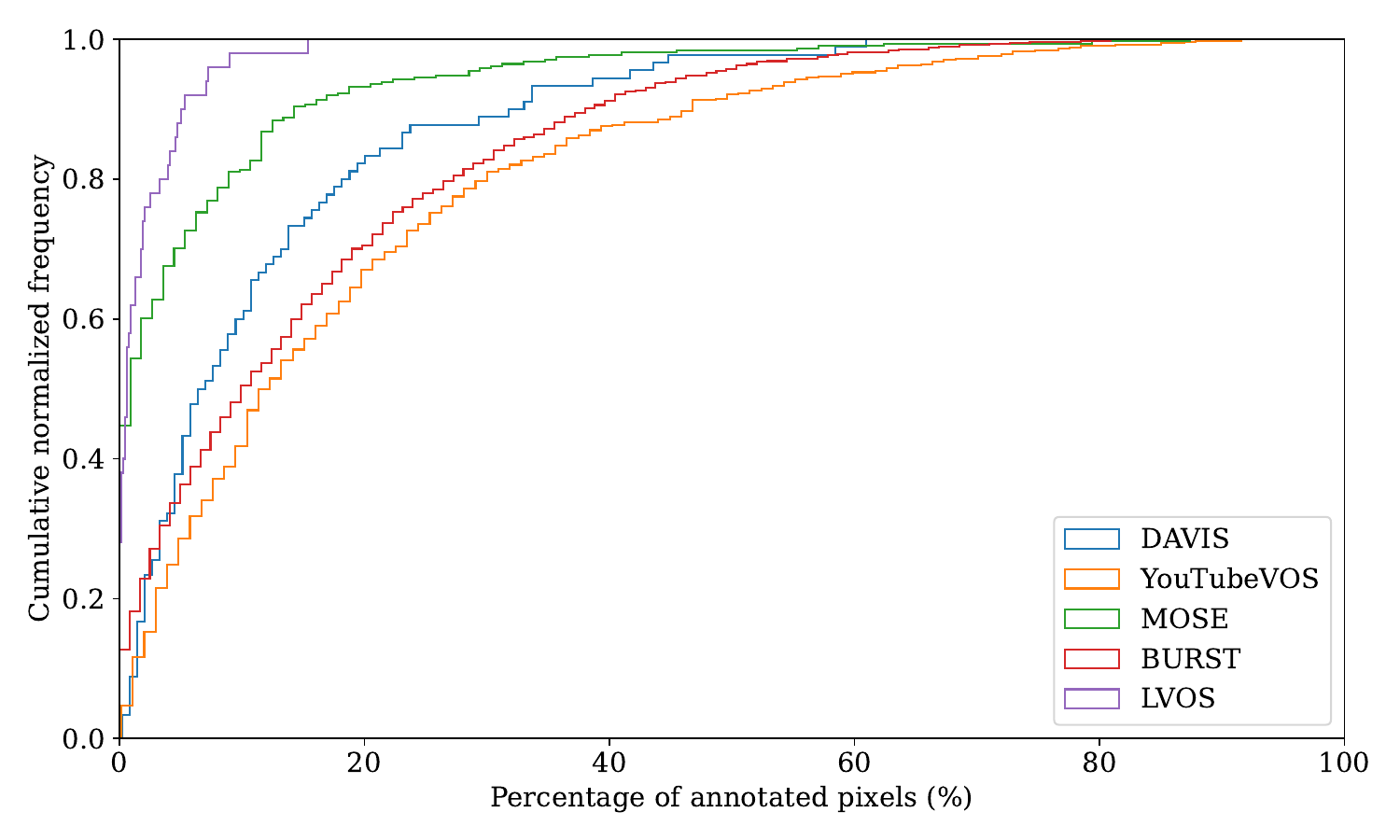}
    \caption{Cumulative frequency graph of annotated pixel areas (as percentages of the total image area) for different datasets.
    Tiny objects are significantly more prevalent on LVOS~\cite{hong2022lvos} than on other datasets.}
    \label{fig:app:lvos-object-sizes}
\end{figure}

LVOS~\cite{hong2022lvos} is a recently proposed long-term video object segmentation benchmark, with 50 videos in its validation set and test set respectively.
Note, we have also presented results in another long-term video object segmentation benchmark, BURST~\cite{athar2023burst} in the main paper, which contains 988 videos in the validation set and 1419 videos in the test set.
We test Cutie on LVOS \emph{after} completing the design of Cutie, adopt long-term memory~\cite{cheng2022xmem}, and perform no tuning.
We note that our method (Cutie-base) performs better than DDMemory, the baseline presented in LVOS~\cite{hong2022lvos}, on the test set and has a comparable performance on the validation set, while running about twice as fast.
Upon manual inspection of the results, we observe that one of the unique challenges in LVOS is the prevalence of tiny objects, which our algorithm has not been specifically designed to handle.
We quantify this observation by analyzing the first frame annotations of all the videos in the validation sets of DAVIS~\cite{perazzi2016benchmark}, YouTubeVOS~\cite{xu2018youtubeVOS}, MOSE~\cite{ding2023mose}, BURST~\cite{athar2023burst}, and LVOS~\cite{hong2022lvos}, as shown in Figure~\ref{fig:app:lvos-object-sizes}.
Tiny objects are significantly more prevalent on LVOS~\cite{hong2022lvos} than on other datasets.
We think this makes LVOS uniquely challenging for methods that are not specifically designed to detect small objects.

\subsection{Performance Variations}
To assess performance variations with respect to different random seeds, we train Cutie-small with five different random seeds (including both pretraining and main training with the MOSE dataset) and report mean$\pm$standard deviation on the MOSE~\cite{ding2023mose} validation set and the YouTubeVOS 2019~\cite{xu2018youtubeVOS} validation set in Table~\ref{tab:app:random-variations}.
Note, the improvement brought by our model (i.e., 8.7~\mjf~on MOSE and 0.9~\mg~on YouTubeVOS over XMem~\cite{cheng2022xmem}) corresponds to $+24.2 \text{~s.d.}$ and $+8.2\text{~s.d.}$ respectively.

\begin{table}[h]
    \centering
    \centering
\begin{tabular}
{C{2.2em}@{\hspace{6pt}}C{2.2em}@{\hspace{6pt}}C{2.2em}@{\hspace{6pt}}C{2.2em}@{\hspace{6pt}}C{2.2em}@{\hspace{6pt}}C{2.2em}@{\hspace{6pt}}C{2.2em}@{\hspace{6pt}}C{2.2em}}
\toprule
\multicolumn{3}{c}{\hspace{-10pt}MOSE val\hspace{10pt}} & \multicolumn{5}{c}{YouTubeVOS-2019 val} \\
\cmidrule(lr){1-3} \cmidrule(lr){4-8}
\mjf & \mj & \mf & \mg & \mjs & \mfs & \mju & \mfu \\
\midrule
67.3$\pm$0.36 & 63.1$\pm$0.36 & 71.6$\pm$0.35 & 86.2$\pm$0.11 & 85.1$\pm$0.20 & 89.6$\pm$0.27 & 81.1$\pm$0.19 & 89.3$\pm$0.13 \\
\midrule
\bottomrule
\end{tabular}

    \caption{Performance variations (median$\pm$standard deviation) across five different random seeds.}
    \label{tab:app:random-variations}
\end{table}

\section{Implementation Details}\label{sec:app:implementation}
Here, we include more implementation details for completeness. 
Our training and testing code will be released for reproducibility.

\subsection{Extension to Multiple Objects}
We extend Cutie to the multi-object setting following~\cite{oh2019videoSTM,cheng2021stcn,cheng2022xmem,cheng2023tracking}. 
Objects are processed independently (in parallel as a batch) except for 
1) the interaction at the first convolutional layer of the mask encoder, which extracts features corresponding to a target object with a 5-channel input concatenated from the image (3-channel), the mask of the target object (1-channel), and the sum of masks of all non-target objects (1-channel); 
2) the interaction at the soft-aggregation layers~\cite{oh2019videoSTM} used to generate segmentation logits -- where the object probability distributions at every pixel are normalized to sum up to one. 
Note these are standard operations from prior works~\cite{oh2019videoSTM,cheng2021stcn,cheng2022xmem,cheng2023tracking}.
Parts of the computation (i.e., feature extraction from the query image and affinity computation) are shared between objects while the rest are not. 
We experimented with object interaction within the object transformer in the early stage of this project but did not obtain positive results. 

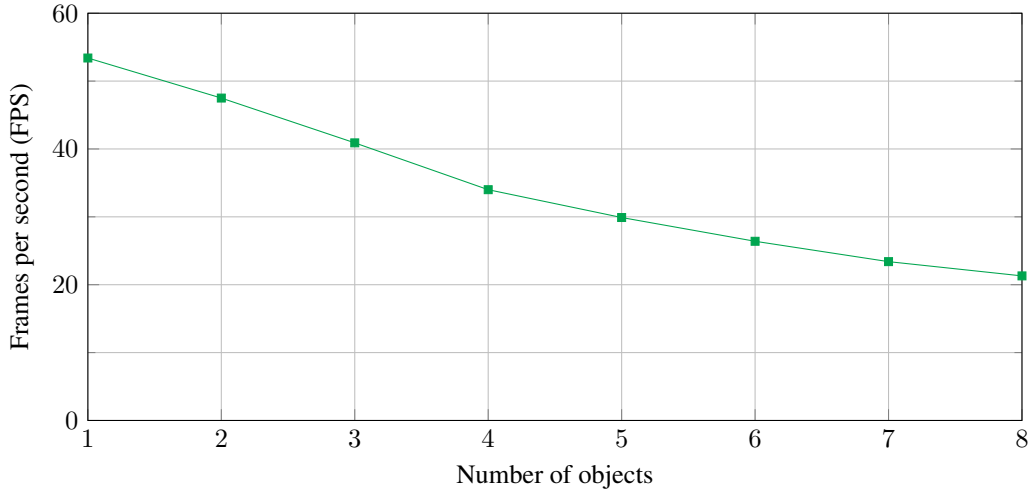
\begin{figure}
    \centering
    \begin{tikzpicture}
	\begin{axis}[
		xlabel={Number of objects},
		ylabel={Frames per second (FPS)},
		xmin=1, xmax=8,
		ymin=0, ymax=60,
        yminorticks=true,
        minor y tick num=1,
		grid=both,
		height=7cm, width=14cm,
        xtick={1, 2, 3, 4, 5, 6, 7, 8},
		]
				\addplot[mark=*,mark=square*,color=Green,mark size=1.5pt]
		coordinates {
(1),(53.4)
(2),(47.5)
(3),(40.9)
(4),(34.0)
(5),(29.9)
(6),(26.4)
(7),(23.4)
(8),(21.3)
		};
  
	\end{axis}
\end{tikzpicture}
    \caption{Cutie-small's processing speed with respect to the number of objects in the video. Common benchmarks (DAVIS~\cite{perazzi2016benchmark}, YouTubeVOS~\cite{xu2018youtubeVOS}, and MOSE~\cite{ding2023mose}) average 2-3 objects per video with longer-term benchmarks like BURST~\cite{athar2023burst} averaging 5.57 objects per video -- our model remains real-time (25+ FPS) in these scenarios.
    For evaluation, we use standard $854\times480$ test videos with 100 frames each.
    }
\label{fig:app:plot-fps_vs_num_objects}
\end{figure}

Figure~\ref{fig:app:plot-fps_vs_num_objects} plots the FPS against the number of objects. 
Our method slows down with more objects but remains real-time when handling a common number of objects in a scene (29.9 FPS with 5 objects). 
For instance, the BURST~\cite{athar2023burst} dataset averages 5.57 object tracks per video and DAVIS-2017~\cite{perazzi2016benchmark} averages just 2.03.

\begin{figure}
    \centering
    \begin{tikzpicture}
	\begin{axis}[
		xlabel={Number of processed frames},
		ylabel={GPU memory usage (MB)},
		xmin=100, xmax=3000,
        ymode=log,
		ymin=100, ymax=25600,
        yminorticks=true,
        xtick={100, 500, 1000, 1500, 2000, 2500, 3000},
		grid=both,
		height=7cm, width=14cm,
		legend style={font=\small,at={(0.55,0.8)},anchor=west},
		legend cell align={left},
		]
				\addplot[mark=*,mark=square*,color=Green,mark size=1.5pt]
		coordinates {
(100),(260.8)
(200),(464.0)
(300),(464.0)
(400),(464.0)
(500),(464.0)
(600),(464.0)
(700),(464.0)
(800),(464.0)
(900),(464.0)
(1000),(464.0)
(1100),(464.0)
(1200),(464.0)
(1300),(464.0)
(1400),(464.0)
(1500),(464.0)
(1600),(464.0)
(1700),(464.0)
(1800),(464.0)
(1900),(464.0)
(2000),(464.0)
(2100),(464.0)
(2200),(464.0)
(2300),(464.0)
(2400),(464.0)
(2500),(464.0)
(2600),(464.0)
(2700),(464.0)
(2800),(464.0)
(2900),(464.0)
(3000),(464.0)
		};
		\addlegendentry{Ours (first-in-first-out)}
		
		\addplot[mark=*,mark=diamond*,color=Cerulean]
		coordinates {
			(100),(260.8)
(200),(689.9)
(300),(707.6)
(400),(722.6)
(500),(743.9)
(600),(756.4)
(700),(775.1)
(800),(788.7)
(900),(807.4)
(1000),(828.2)
(1100),(839.2)
(1200),(860.5)
(1300),(872.5)
(1400),(890.9)
(1500),(906.6)
(1600),(923.9)
(1700),(946.2)
(1800),(957.3)
(1900),(975.2)
(2000),(991.0)
(2100),(1007.5)
(2200),(1007.5)
(2300),(1007.5)
(2400),(1007.5)
(2500),(1007.5)
(2600),(1007.5)
(2700),(1007.5)
(2800),(1007.5)
(2900),(1007.5)
(3000),(1007.5)
		};
		\addlegendentry{Ours (long-term memory)}

		\addplot[mark=*,mark=pentagon*,color=Brown]
		coordinates {
(100),(624.3)
(200),(1035.6)
(300),(1050.5)
(400),(1069.6)
(500),(1089.9)
(600),(1102.2)
(700),(1122.3)
(800),(1135.8)
(900),(1155.0)
(1000),(1171.3)
(1100),(1189.0)
(1200),(1208.2)
(1300),(1220.9)
(1400),(1241.6)
(1500),(1255.8)
(1600),(1274.3)
(1700),(1294.7)
(1800),(1307.3)
(1900),(1328.7)
(2000),(1341.2)
(2100),(1360.7)
(2200),(1360.7)
(2300),(1360.7)
(2400),(1360.7)
(2500),(1360.7)
(2600),(1360.7)
(2700),(1360.7)
(2800),(1360.7)
(2900),(1360.7)
(3000),(1360.7)
		};
		\addlegendentry{XMem (long-term memory)}

  		\addplot[mark=*,mark=triangle*,color=Magenta]
		coordinates {
(100),(2010.7)
(200),(3630.5)
(300),(5252.1)
(400),(6875.4)
(500),(8500.3)
(600),(10120.7)
(700),(11742.9)
(800),(13367.1)
(900),(14981.8)
(1000),(16598.6)
(1100),(18211.0)
(1200),(19829.4)
(1300),(21451.1)
(1400),(23067.7)
		};
		\addlegendentry{DeAOT (unbounded memory)}
  
	\end{axis}
\end{tikzpicture}
    \caption{Running GPU memory usage (log-scale) comparisons with respect to the number of processed frames during inference.
    By default, we use a first-in-first-out (FIFO) memory bank which leads to constant memory usage over time. 
    Optionally, we include the long-term memory from XMem~\cite{cheng2022xmem} in our method for better performance on long videos. 
    Our method (with long-term memory) uses less memory than XMem because of a smaller channel size (256 in our model; 512 in XMem).
    DeAOT~\cite{yang2022decoupling} has an unbounded memory size and is impractical for processing long videos -- our hardware (32GB V100, server-grade GPU) cannot process beyond 1,400 frames.
    }
\label{fig:app:plot-mem_vs_length}
\end{figure}
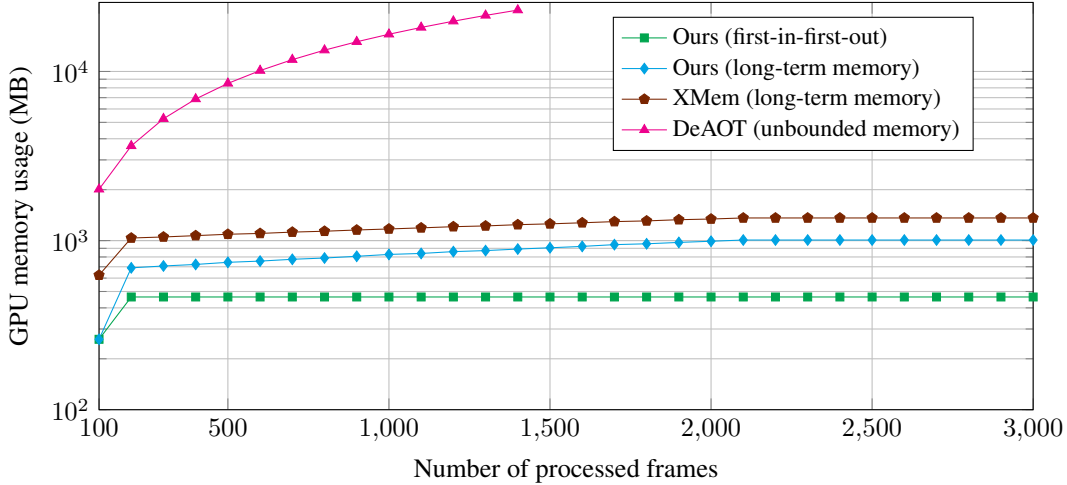

Additionally, we plot the memory usage with respect to the number of processed frames during inference in Figure~\ref{fig:app:plot-mem_vs_length}.

\subsection{Streaming Average Algorithm for the Object Memory}
To recap, we store a compact set of $N$ vectors which make up a high-level summary of the target object in the object memory $S\in\mathbb{R}^{N\times C}$.
At a high level, we compute $S$ by mask-pooling over all encoded object features with $N$ different masks. Concretely, given object features $U\in \mathbb{R}^{THW\times C}$ and $N$ pooling masks $\{W_q\in [0, 1]^{THW}, 0<q\leq N\}$, where $T$ is the number of memory frames, the $q$-th object memory $S_q\in\mathbb{R}^C$ is computed by
\begin{equation}
    S_q = \frac{\sum^{THW}_{i=1}{U(i) W_q(i)}}{\sum^{THW}_{i=1}{W_q(i)}}.
\end{equation}

During inference, we use a classic streaming average algorithm such that this operation takes constant time and memory with respect to the memory length. 
Concretely, for the $q$-th object memory at time step $t$, we keep track of a cumulative memory $\sigma^t_{S_q}\in\mathbb{R}^{C}$ and a cumulative weight $\sigma^t_{W_q}\in\mathbb{R}$. 
We update the accumulators and find $S_q$ via
\begin{equation}
    \sigma^t_{S_q} = \sigma^{t-1}_{S_q} + \sum^{THW}_{i=1}{U(i) W_q(i)}, \quad\quad
    \sigma^t_{W_q} = \sigma^{t-1}_{W_q} + \sum^{THW}_{i=1}{W_q(i)}, \quad\quad\text{and}\quad\quad
    S_q = \frac{\sigma^t_{S_q}}{\sigma^t_{W_q}},
\end{equation}
where $U$ and $W_q$ can be discarded after every time step.

\subsection{Training Details}
As mentioned in the main paper, we train our network in two stages: static image pretraining and video-level main training following prior works~\cite{oh2019videoSTM,yang2021associating,cheng2022xmem}. Backbone weights are initialized from ImageNet~\cite{deng2009imagenet} pretraining, following prior work~\cite{oh2019videoSTM,yang2021associating,cheng2022xmem}.
We implement our network with PyTorch~\cite{PyTorch} and use automatic mixed precision (AMP) for training. 

\subsubsection{Pretraining}
Our pretraining pipeline follows the open-sourced code of~\cite{cheng2021mivos,cheng2021stcn,cheng2022xmem}, and is described here for completeness.
For pretraining, we use a set of image segmentation datasets: ECSSD~\cite{shi2015hierarchicalECSSD}, DUTS~\cite{wang2017DUTS}, FSS-1000~\cite{li2020fss}, HRSOD~\cite{zeng2019towardsHRSOD}, and BIG~\cite{cheng2020cascadepsp}.
We mix these datasets and sample HRSOD~\cite{zeng2019towardsHRSOD} and BIG~\cite{cheng2020cascadepsp} five times more often than the others as they are more accurately annotated.
From a sampled image-segmentation pair, we generate synthetic sequences of length three by deforming the pair with random affine transformation, thin plate spline transformation~\cite{duchon1977splines}, and cropping (with crop size $384\times384$).
With the generated sequence, we use the first frame (with ground-truth segmentation) as the memory frame to segment the second frame. Then, we encode the second frame with our predicted segmentation and concatenate it with the first-frame memory to segment the third frame.
Loss is computed on the second and third frames and back-propagated through time.

\subsubsection{Main Training}
For main training, we use two different settings. 
The ``without MOSE'' setting mixes the training sets of DAVIS-2017~\cite{perazzi2016benchmark} and YouTubeVOS-2019~\cite{xu2018youtubeVOS}. 
The ``with MOSE'' setting mixes the training sets of DAVIS-2017~\cite{perazzi2016benchmark}, YouTubeVOS-2019~\cite{xu2018youtubeVOS}, and MOSE~\cite{ding2023mose}.
In both settings, we sample DAVIS~\cite{perazzi2016benchmark} two times more often as its annotation is more accurate. 
To sample a training sequence, we first randomly select a ``seed'' frame from all the frames and randomly select seven other frames from the same video. 
We re-sample if any two consecutive frames have a temporal frame distance above $D$. We employ a curriculum learning schedule following~\cite{cheng2022xmem} for $D$, which is set to $[5, 10, 15, 5]$ correspondingly after $[0\%, 10\%, 30\%, 80\%]$ of training iterations.

For data augmentation, we apply random horizontal mirroring, random affine transformation, cut-and-paste~\cite{ghiasi2021simple} from another video, and random resized crop (scale $[0.36, 1.00]$, crop size $480\times480$).
We follow stable data augmentation~\cite{cheng2023tracking} to apply the same crop and rotation to all the frames in the same sequence.
We additionally apply random color jittering and random grayscaling to the sampled images  following~\cite{cheng2021stcn,cheng2022xmem}. 

To train on a sampled sequence, we follow the process of pretraining, except that we only use a maximum of three memory frames to segment a query frame following~\cite{cheng2022xmem}. When the number of past frames is smaller or equal to 3, we use all of them, otherwise, we randomly sample three frames to be the memory frames.
We compute the loss at all frames except the first one and back-propagate through time.

\subsubsection{Point Supervision}
As mentioned in the main paper, we adopt point supervision~\cite{cheng2022masked} for training. 
As reported by~\cite{cheng2022masked}, using point supervision for computing the loss has insignificant effects on the final performance while using only one-third of the memory during training.
In Cutie, we note that using point supervision reduces the memory cost during loss computation but has an insignificant impact on the overall memory usage.
We use importance sampling with default parameters following~\cite{cheng2022masked}, i.e., with an oversampling ratio of 3, and sample $75\%$ of all points from uncertain points and the rest from a uniform distribution. 
We use the uncertainty function for semantic segmentation (by treating each object as an object class) from~\cite{kirillov2020pointrend}, which is the logit difference between the top-2 predictions.
Note that using point supervision also focuses the loss in uncertain regions but this is not unique to our framework.
Prior works XMem~\cite{cheng2022xmem} and DeAOT~\cite{yang2022decoupling} use bootstrapped cross-entropy to similarly focus on difficult-to-segment pixels. 
Overall, we do not notice significant segmentation accuracy differences in using point supervision vs.\ the loss in XMem~\cite{cheng2022xmem}.

\subsection{Decoder Architecture}
Our decoder design follows XMem~\cite{cheng2022xmem} with a reduced number of channels. XMem~\cite{cheng2022xmem} uses 256 channels while we use 128 channels.
This reduction in the number of channels improves the running time. We do not observe a performance drop from this reduction which we think is attributed to better input features (which are already refined by the object transformer).

The inputs to the decoder are the object readout feature $R_L$ at stride 16 and skip-connections from the query encoder at strides 8 and 4. 
The skip-connection features are first projected to $C$ dimensions with a $1\times1$ convolution. 
We process the object readout features with two upsampling blocks and incorporate the skip-connections for high-frequency information in each block.
In each block, we first bilinearly upsample the input feature by two times, then add the upsampled features with the skip-connection features. This sum is processed by a residual block~\cite{he2016deepResNet} with two $3\times3$ convolutions as the output of the upsample block.
In the final layer, we use a $3\times3$ convolution to predict single-channel logits for each object. The logits are bilinearly upsampled to the original input resolution.
In the multi-object scenario, we use soft-aggregation~\cite{oh2019videoSTM} to merge the object logits.

\subsection{Details on Pixel Memory}
As discussed in the main paper, we derive our pixel memory design from XMem~\cite{cheng2022xmem} without claiming contributions. Namely, the attentional component is derived from the working memory, and the recurrent component is derived from the sensory memory of XMem~\cite{cheng2022xmem}.
Long-term memory~\cite{cheng2022xmem}, which compresses the working memory during inference, can be adopted without re-training for evaluation on long videos.

\subsubsection{Attentional Component}
For the attentional component, we store memory keys $\mathbf{k}\in\mathbb{R}^{THW\times C^\text{k}}$ and memory value $\mathbf{v}\in\mathbb{R}^{THW\times C}$ and later retrieve features using a query key $\mathbf{q}\in\mathbb{R}^{HW\times C^\text{k}}$. Here, $T$ is the number of memory frames and $H,W$ are image dimensions at stride 16.
As we use the anisotropic L2 similarity function~\cite{cheng2022xmem}, we additionally store a memory shrinkage $\mathbf{s}\in[1, \infty]^{THW}$ term and use a query selection term $\mathbf{e}\in[0, 1]^{HW\times C^\text{k}}$ during retrieval.

The anisotropic L2 similarity function $d(\cdot, \cdot)$ is computed as
\begin{equation}
    d(\mathbf{q}_i, \mathbf{k}_j) = -\mathbf{s}_j \sum_c^{C^\text{k}} \mathbf{e}_{ic} (\mathbf{k}_{ic} - \mathbf{q}_{jc}).
\end{equation}
We compute memory keys $\mathbf{k}$, memory shrinkage terms $\mathbf{s}$, query keys $\mathbf{q}$, and query selection terms $\mathbf{e}$ by projecting features encoded from corresponding RGB images using the query encoder.
Since these terms are only dependent on the image, they, and thus the affinity matrix $A^{\text{pix}}$ can be shared between multiple objects with no additional costs.
The memory value $\mathbf{v}$ is computed by fusing features from the mask encoder (that takes both image and mask as input) and the query encoder. This fusion is done by first projecting the input features to $C$ dimensions with $1\times1$ convolutions, adding them together, and processing the sum with two residual blocks, each with two $3\times3$ convolutions.

\subsubsection{Recurrent Component}
The recurrent component stores a hidden state $\mathbf{h}^{HW\times C}$ which is updated by a Gated Recurrent Unit (GRU)~\cite{cho2014propertiesGRU} every frame. 
This GRU takes multi-scale inputs (from stride 16, 8, and 4) from the decoder to update the hidden state $\mathbf{h}$.
We first area-downsample the input features to stride 16, then project them to $C$ dimensions before adding them together. This summed input feature, together with the last hidden state, is fed into a GRU as defined in XMem~\cite{cheng2022xmem} to generate a new hidden state. 

Every time we insert a new memory frame, i.e., every $r$-th frame, we apply a \textit{deep update} as in XMem~\cite{cheng2022xmem}. Deep update uses a separate GRU that takes the output of the mask encoder as its input feature. This incurs minimal overhead as the mask encoder is invoked during memory frame insertion anyway. Deep updates refresh the hidden state and allow it to receive updates from a deeper network.

\section{Interactive Tool for Video Segmentation}\label{sec:app:interactive-tool}
\begin{figure}
    \centering
    \includegraphics[width=\linewidth]{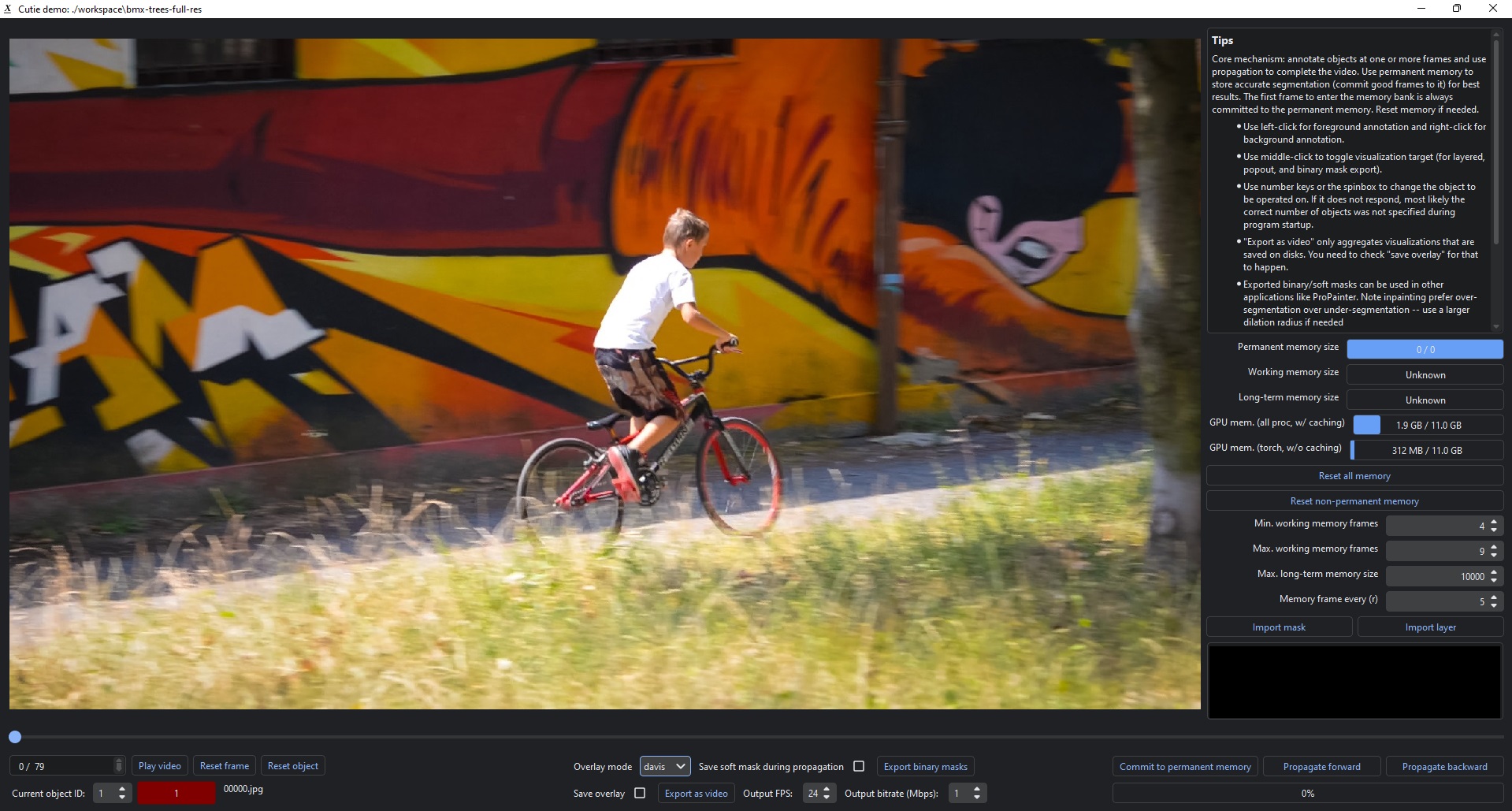}
    \caption{Screenshot of our interactive video segmentation tool. }
    \label{fig:app:tool}
\end{figure}

Based on the video object segmentation capability of Cutie, we build an interactive video segmentation tool to facilitate research and data annotation.
We follow the decoupled paradigm of MiVOS~\cite{cheng2021mivos} -- users annotate one or more frames using an interactive image segmentation tool such as RITM~\cite{sofiiuk2022reviving} and use Cutie for propagating these image segmentations through the video.
Users can also include multiple permanent memory frames (as in XMem++~\cite{bekuzarov2023xmem++}) to increase the segmentation robustness.
Figure~\ref{fig:app:tool} shows a screenshot of this tool. 

This interactive tool is open-source to benefit researchers, data annotators, and video editors.

\end{document}